\documentclass[10pt]{article}
\usepackage[letterpaper,margin=1in]{geometry}
\usepackage[T1]{fontenc}
\usepackage{times}
\usepackage{microtype}
\usepackage{amsmath,amssymb,amsthm}
\usepackage{graphicx}
\usepackage{booktabs}
\usepackage{xcolor}
\usepackage[colorlinks=true,linkcolor=blue!60!black,citecolor=blue!60!black,urlcolor=blue!60!black]{hyperref}
\usepackage{natbib}
\usepackage{caption}
\usepackage{enumitem}
\usepackage{float}
\usepackage{multirow}
\newtheorem{proposition}{Proposition}[section]
\newtheorem{corollary}[proposition]{Corollary}
\newcommand{\suppapp}[1]{Appendix~\ref{#1}}
\newcommand{\supptab}[1]{Table~\ref{#1}}

\hypersetup{pdftitle={The Convention Gap: Beyond Literal Content in Human-AI Cooperation},
  pdfauthor={Makoto Fukushima, Hua-Dong Xiong, Ehsan Moradi Pari}}

\title{The Convention Gap: Beyond Literal Content in Human--AI Cooperation}
\author{Makoto Fukushima\thanks{Corresponding author: \texttt{makoto.fukushima@jp.honda-ri.com}}\\ Honda Research Institute Japan
  \and Hua-Dong Xiong\\ Honda Research Institute USA
  \and Ehsan Moradi Pari\\ Honda Research Institute USA}
\date{}

\providecommand{\Description}[2][]{}
\begin{document}
\maketitle
\begin{abstract}
Cooperative AI agents are often evaluated against other AIs, yet human cooperation is thought to rely on implicit conventions: shared rules for reading meaning beyond the literal message. We propose the \emph{convention gap} as a diagnostic of such implicit communication. In the card game Hanabi, players see each other's cards but not their own, hints are the only messages about one's own hand, and a play fails if the card cannot be added to the stacks on the table. The convention gap is the failure probability predicted from the literal content of the hints and the visible cards, minus the observed failure rate of the cards played. This prediction can be computed exactly. We prove that the population gap (the value the gap approaches over many games) is zero when the partner's choices convey nothing beyond their literal content (given a fair shuffle and a player who uses only what was said and seen). A nonzero population gap thus indicates that partner choices conveyed more; it does not measure how much. We applied it to public human-human and human-AI games and to AI-AI games we simulated among rule-based agents. The gap was +26.2 percentage points (pp) in human pairs, +16.4~pp for human plays with AI partners, and $-$0.6~pp in AI pairs, most of whose plays had zero literal failure probability. In human pairs it was +46~pp on the 5\% of plays of cards that no hint had pointed to and +36~pp on the 65\% pointed to once. On the same turns, the card in hand with the lowest literal failure probability would have failed 15.6\% of the time, against 5.3\% for the card humans played, and the timing of hints predicted which card they played. The human-AI games paired people with three AI agents of a published study: Full and Intentional, which choose hints by reasoning about intent, and Outer, which does not. The literal failure probability of the cards humans played was similar across the three (38--41\%), yet human failure ranged from 14.4\% to 34.4\% and the gap from +24.1 to +6.2~pp, and the contrast with Outer persisted within participants. Across five levels of Off-Belief Learning agents, trained so that the first level reads past actions only for their literal content and each higher level interprets more, the gap rose monotonically from +1.6 to +21.7~pp: higher levels played cards with greater literal failure probability (7.8\% to 24.1\%) while failing less often than the first level. These results argue for reporting the convention gap on human-partnered play, not only AI-AI performance, when evaluating agents intended for human partners.

\end{abstract}

\section{Introduction}

AI agents for cooperative tasks increasingly operate alongside human partners \citep{dafoe2021cooperative}. They are often evaluated against other AIs, in self-play or in cross-play with a different AI. If human cooperation relies on implicit conventions that AIs do not use, these benchmarks may measure an agent's compatibility with other AIs rather than with humans. We propose a diagnostic that compares play outcomes with an exact literal-information reference, and apply it to human-human, AI-AI, and human-AI play.

The card game Hanabi \citep{bard2020hanabi} is a standard testbed for convention-based cooperation. Each player sees the partner's cards but not their own. Hints are the only messages about one's own hand: a hint \emph{positively touches} the cards it points at and negatively constrains the others. Further literal information comes from counting the cards visible in the partner's hand, the discard pile and the \emph{fireworks} (the stacks of played cards). Playing a card that does not fit the fireworks costs a \emph{life token}, and three such failures end the game. Experienced human pairs supplement literal content with conventions: shared protocols for reading meaning beyond what is literally said \citep{grice1975logic,sidji2023hidden}.

We introduce the \emph{convention gap}, a signed calibration diagnostic that compares observed play failures with the failure probability predicted by an exact literal reference model. In Hanabi this reference is computable by enumeration. Its population value (the large-sample limit of the gap pooled over plays) is zero under three conditions (Proposition~\ref{prop:zero}). The shuffle is uniform, the partner's action choices carry no information beyond their literal content, and the \emph{receiver} (the player whose plays are scored) acts only on the game record it has seen. A nonzero population gap therefore shows that at least one of these conditions fails; the observed gap estimates the population gap. The gap is expressed in percentage points of failure and can be negative. Standardized comparisons within our data and Off-Belief Learning (OBL) agents are consistent with a convention-based reading of the departures we observe and do not identify it uniquely. The name reflects this motivating case; the gap is not specific to conventions, since a partner who hints playable cards also produces a positive gap (\S\ref{sec:obl_results}).

We applied the diagnostic to 86,968 plays (human-human, human plays in human-AI, and simulated AI-AI games; \S\ref{sec:gap_exists}--\S\ref{sec:receiver}), to the 13,577 plays of the AIs in the human-AI games (\S\ref{sec:partner_dependence}), and to five levels of OBL agents (\S\ref{sec:obl_results}). The gap was +26.2~pp in human pairs, +16.4~pp for humans with AI partners and $-$0.6~pp in AI-AI play. With AI partners it was larger with the two intent-reasoning agents, Full and Intentional, than with Outer, also within participants. In human pairs, the card played failed less often than the card with the lowest literal failure probability on the same turns. Our contributions are the diagnostic, the proposition that its population value is zero under literal-only partner choices, with an exact null control, and these comparisons.

\section{Related Work}

\textbf{Hanabi agents.} \citet{walton2017evaluating} introduced rule-based agents and evaluated them with one another; the Hanabi Open Agent Dataset (HOAD) \citep{sarmasi2021hoad} includes four of them. \citet{eger2017intentional} ran human-AI experiments with their own agents Outer, Intentional, and Full. Self-play deep RL methods such as BAD \citep{foerster2019bayesian} and SAD \citep{hu2020simplified} achieve strong self-play scores, but their conventions often fail with independently trained partners. Other-Play \citep{hu2020other} was designed to address this, and its agents scored higher than self-play agents when paired with humans.

\textbf{Zero-shot coordination.} Any-Play \citep{lucas2022any} uses intrinsic augmentation for zero-shot coordination. The AH2AC2 challenge \citep{dizdarevic2025ah2ac2} builds human proxy agents from games on hanab.live, a public online Hanabi platform, and \citet{bredell2025augmenting} augment the action space with human-style conventions. \citet{jeon2023behavioral} related behavioral differences between RL agents to ad hoc cooperation failure. These works build agents that transfer. We instead diagnose, without new training, whether play departs from a literal reference.

\textbf{Off-Belief Learning.} OBL \citep{hu2021offbelief} trains a policy that interprets past actions as if a fixed policy $\pi_0$ had taken them, while assuming that future play follows the policy being learned. At level 1, $\pi_0$ is uniformly random over legal actions, so past actions carry almost no information beyond their literal content in the learner's belief (footnote~\ref{fn:legal}). At level $k{+}1$, $\pi_0$ is the level-$k$ policy. The level-1 belief is the trained analogue of our exact literal reference, a finite sum computed by dynamic programming (\S\ref{sec:posterior}). OBL thus constrains at training time the belief that our reference fixes at evaluation time.

\textbf{Human-AI Hanabi studies.} In a study with 246 completed surveys, 41\% of participants paired with an implicature-based agent judged their teammate to be human, against 24\% for an entropy-based baseline \citep{liang2019implicit}. \citet{siu2021evaluation} found that humans preferred rule-based agents over RL agents despite similar scores. \citet{sidji2023hidden} catalogued human conventions qualitatively, and \citet{attig2024perceived} compared perceived cooperativity of a rule-based and an RL agent. Our work adds an automated outcome diagnostic applied to logged games.

\textbf{Pragmatics and communication measures.} \citet{carroll2019utility} showed that self-play agents fail with humans in Overcooked. The Rational Speech Acts (RSA) framework \citep{frank2012predicting,goodman2016pragmatic} formalizes Gricean pragmatics as recursive inference between literal and pragmatic listeners. Our literal reference resembles a literal listener in using only literal hint content, although it is a combinatorial computation over card identities. The gap is a calibration residual against that listener. \citet{lowe2019pitfalls} showed that measures of emergent communication can mislead. They distinguished positive signaling, in which messages depend on the sender's observations or actions, from positive listening, in which messages influence the receiver's behavior. \citet{sivan2025information} subtract an estimated wording baseline from total information; we subtract failure rates.

Prior work thus reports that AI-AI performance need not predict human-AI cooperation and describes implicit conventions in human play. What has been lacking is a quantitative diagnostic with an exact literal reference that applies uniformly across the three settings.

\section{Method}

\textbf{Game setting in brief.} Hanabi's 50-card deck spans 5 colors $\times$ ranks 1--5, with 3/2/2/2/1 copies of ranks 1--5 per color. On each turn a player takes one of three actions. A \emph{play} adds a card to the fireworks, where each color must be built $1{\to}5$; a misplay costs one of 3 shared life tokens, and the third loss ends the game. A \emph{discard} recovers one of 8 information tokens. A \emph{hint} spends an information token to name a color or rank, pointing out all matching cards in the partner's hand. \emph{Partial} scoring counts the cards successfully played (0--25). \emph{Strict} scoring sets the score to 0 when the game ends on the third misplay. All analyses use 2-player games with the standard deck (hanab.live's ``No Variant''; rules in \suppapp{app:rules}); \S\ref{sec:data} lists departures of the recorded games from the standard rules.

\subsection{The Convention Gap}
\label{sec:method_convgap}

We define the convention gap as the \emph{signed calibration residual of a literal reference model}. It is the difference between the mean failure probability under that model (\S\ref{sec:posterior}) and the observed failure rate over $N$ play actions:
\begin{equation}
\text{Convention Gap} = \underbrace{\frac{1}{N}\sum_{i=1}^{N} \hat{p}_i}_{\text{mean }\hat{p}} - \underbrace{\frac{1}{N}\sum_{i=1}^{N} y_i}_{\text{failure rate}}.
\label{eq:convgap}
\end{equation}
Here $\hat{p}_i$ is the literal failure probability of play $i$, the probability under the reference model that it fails and loses a life (Eq.~\ref{eq:posterior}), and $y_i \in \{0,1\}$ is its outcome (1 = failure). $N$ is the number of play actions in the group analyzed (a setting, a player type, or a partner). Every play action is one observation, pooled across games and players; we call this sample quantity the \emph{pooled} (play-weighted) gap. Let $R_g$ be the sum of $\hat{p}_i - y_i$ over the $N_g$ analyzed plays of game $g$. If games are independent draws from one population, the pooled gap converges as games accumulate to the \emph{population gap} $\mathbb{E}[R_g]/\mathbb{E}[N_g]$, which it estimates. A positive gap means that the evaluated plays failed less often than the literal reference predicts.

The next result limits what can produce a nonzero gap. Let $Z$ be the initial shuffled deck, including the future draw order. Let $\mathcal{H}_t$ be the transcript observable to the receiver (the acting player at each scored play) before the play at turn $t$: every action so far, the game's response to it, and the partner cards the receiver sees. The other player is the partner, so on an AI's own plays in human-AI games the human is the partner. Let $\mathcal{H}_t^{+} = (\mathcal{H}_t, a_t)$ be the transcript together with the slot $a_t$ played. Let $D_t$ be the set of shuffles on which replaying the observed actions reproduces $\mathcal{H}_t$ (visible cards and draws, discards, fireworks, and the slots each hint touched). The literal reference law $Q_t$ is uniform over $D_t$ (Eq.~\ref{eq:posterior} is its marginal on the receiver's hand), and $\hat{p}_t = Q_t(y_t = 1)$. $Q_t$ uses the constraints that a transcript imposes, not the likelihood that the partner would have produced that transcript.

Each partner action consists of a \emph{choice} $c_s$ (which attribute to name, or which own card to play or discard) and the game's deterministic \emph{response} $\ell_s = \ell(c_s, Z, \mathcal{H}_{s})$ (the slots touched, or the card revealed and its replacement draw). $P$ denotes the actual, policy-dependent law.

\begin{proposition}[Zero population gap]
\label{prop:zero}
Assume\begin{enumerate}[label=(\roman*),leftmargin=*,nosep]\item the shuffle $Z$ is uniform;\item for every partner action, $P(c_s \mid Z, \mathcal{H}_{s}) = \varphi(c_s, \ell_s, \mathcal{H}_{s})$ for some function $\varphi$, that is, the partner's choice depends on the shuffle only through the response it produces;\footnote{\label{fn:legal}A uniformly random \emph{legal} hinter violates (ii), because under the rule that a hint must touch at least one card the number of legal hints depends on the hand; in companion cells of the null-control simulation it gave gaps of $+0.10$ to $+0.53$~pp (\suppapp{app:nullcontrol}).} and\item the receiver's own actions, including the choice of which card to play at turn $t$, depend on the shuffle only through the transcript (no side information outside it).\end{enumerate}Then $P(Z \mid \mathcal{H}_t^{+}) = Q_t$, so $\hat{p}_t = P(y_t = 1 \mid \mathcal{H}_t^{+})$, and for any selection of plays that is a function of $\mathcal{H}_t^{+}$, $\mathbb{E}[\hat{p}_t - y_t \mid \text{selected}] = 0$. If games are i.i.d.\ draws from one population with $0 < \mathbb{E}[N_g] < \infty$ selected plays per game, the population gap is zero, and the pooled gap of Eq.~\ref{eq:convgap} converges to zero as games accumulate.
\end{proposition}
\begin{proof}
Fix $\mathcal{H}_t^{+}$. By (ii), every $z \in D_t$ yields the same likelihood $\prod_{s < t} \varphi(c_s, \ell_s, \mathcal{H}_{s})$ for the partner's actions. By (iii), the receiver's own actions, including the play decision at $t$, contribute a factor that is constant on $D_t$. Every $z \notin D_t$ has likelihood zero. Under (i) the prior is uniform, so $P(Z = z \mid \mathcal{H}_t^{+}) \propto \mathbf{1}\{z \in D_t\}$, which is $Q_t$. For the indicator $S_t$ of an $\mathcal{H}_t^{+}$-measurable selection, $\mathbb{E}[S_t(\hat{p}_t - y_t)] = \mathbb{E}[S_t\,\mathbb{E}[\hat{p}_t - y_t \mid \mathcal{H}_t^{+}]] = 0$. Summing over turns, the residual sum $R_g$ of game $g$ has $\mathbb{E}[R_g] = 0$. The pooled gap is $\sum_g R_g / \sum_g N_g$ with i.i.d.\ $(R_g, N_g)$ and $0 < \mathbb{E}[N_g] < \infty$, so it converges to $\mathbb{E}[R_g]/\mathbb{E}[N_g] = 0$ by the law of large numbers. The gap of a single game, $R_g/N_g$, need not have mean zero, because $N_g$ depends on earlier outcomes.
\end{proof}

Condition (ii) is strong. It holds for a partner who names the color or rank of a uniformly chosen slot. It fails whenever the partner's choices depend on the receiver's hand beyond the slots a hint touches, for example when the partner hints playable cards or counts the receiver's cards to decide its own plays. It is sufficient, not necessary, for a zero gap.

\begin{corollary}\label{cor:nonzero}
Under (i) and (iii), a nonzero population gap over an $\mathcal{H}_t^{+}$-measurable selection of plays implies that (ii) fails: the partner's choices depend on the hidden cards beyond the literal content they produce. A positive gap means that the selected plays fail less often than the literal reference model predicts, a negative gap that they fail more often. Neither sign by itself establishes that the player used the information in the partner's choices, shared a convention with the partner, or misread a signal. The statement concerns the population and selection scheme that generated the data. Selections that depend on later outcomes (completed games, final-score strata, per-game means) fall outside it (\S\ref{sec:discussion}).
\end{corollary}

\textbf{Scope.} Under (i)--(iii) no receiver policy that acts on the transcript alone can move the population gap, because $\hat{p}_t$ already conditions on the full transcript. The receiver's play policy determines which discrepancy is sampled, and the gap does not locate the inference (\S\ref{sec:receiver}). The gap can be negative, and it can vanish when strata of opposite sign cancel (\S\ref{sec:obl_results}). A zero gap therefore does not show that the partner was uninformative. An agent that plays only cards with $\hat{p} = 0$ has a gap of exactly zero whatever its partner does; we call such a zero \emph{pinned}. A larger gap does not by itself mean better cooperation. Three results bear on this scope. The AI-AI agents, which meet (iii), give pinned, near-zero or negative gaps (\S\ref{sec:gap_exists}). The human-human gap comes from zero- and one-touch plays (cards that no hint or one hint had positively touched), where hints leave $\hat{p}$ high (\S\ref{sec:hint_localization}). If Full acts on the transcript alone, as (iii) requires (not verified), a positive gap on Full's own plays is evidence that its human partners' choices carried information beyond literal content (\S\ref{sec:partner_dependence}).

\subsection{The Literal Failure Probability}
\label{sec:posterior}

Let the acting player hold $h$ cards. Let $C_j$ be the set of identities $v = (c,r)$ that every hint given while card $j$ was in the hand leaves possible (hints that touched it and hints that did not). Let $m_v$ be the number of copies of $v$ not visible to the player (total copies minus those in the partner's hand, the discard pile and the fireworks). For an assignment $x = (x_1,\dots,x_h)$ with $x_j \in C_j$, let $n_v(x)$ be the number of cards assigned identity $v$. The literal failure probability of playing card $k$ is
\begin{equation}
\hat{p}_t = Q_t(y_t = 1) = \frac{\sum_{x} \mathbf{1}\{x_k \notin \text{Playable}\}\, W(x)}{\sum_{x} W(x)},
\label{eq:posterior}
\end{equation}
where $W(x) = \prod_{v} (m_v)_{n_v(x)}$, the sums run over $x \in \prod_j C_j$, and $(m)_n = m(m-1)\cdots(m-n+1)$ is the falling factorial. Past hints constrain the unknown shuffle only through the cards still held, and the undrawn deck is unconstrained. $W(x)$ is therefore proportional to the number of shuffles in $D_t$ whose hand is $x$, and Eq.~\ref{eq:posterior} is the hand marginal of $Q_t$ in Proposition~\ref{prop:zero}. It is an exact finite-sum posterior, computed by dynamic programming and checked against brute-force enumeration on 936 plays (\suppapp{app:joint}). It excludes any inference about why the partner chose a hint. A per-card approximation ignores the constraints on the other cards. Under it, the human-human gap is $+28.2$~pp, and the other setting-level gaps differ by at most 0.5~pp (\suppapp{app:joint}).

\emph{Worked example.} Hints have narrowed card $k$ to colors $\{$Red, Blue$\}$ and ranks $\{2,3,4,5\}$. The other cards in the hand were all drawn after those hints and carry no constraints. The fireworks show Red at 1 and Blue at 3. The partner's hand and the discard pile expose one copy each of R2, R3, R4, and B3. Then $m_v = 1$ for R2, R3, R4, R5, B2, and B5, $m_v = 2$ for B4, and $m_v = 0$ for B3. The total weight is 8, and the playable identities R2 and B4 carry weight $1 + 2 = 3$. Eq.~\ref{eq:posterior} gives $\hat{p} = 5/8$. The literal reference thus predicts success at only 37.5\%. If such plays succeed more often than that, the gap is positive.

\subsection{Data Sources}
\label{sec:data}\label{sec:datasources}

We replayed each game action by action through our engine and computed $\hat{p}$ before each play was applied (\suppapp{app:pipeline}). All games are two-player games with the standard deck. Two corpora are public recordings and one we simulated.

\textbf{Human-human} (hanab.live): 425 completed games, 9{,}017 plays by 155 players. We fetched the 1{,}500 most recent two-player games with the standard deck listed by the public API on 21 February 2026; the fetch did not select by player or outcome. We dropped 25 games that our engine could not replay and 420 that did not reach a terminal state (a selection on outcome; \S\ref{sec:discussion}), leaving 1{,}055 games. Removing every game in which a verified bot made a play left 598. We also removed every play by an account flagged by a bot screen (\suppapp{app:bots}), which left 425 games with at least one human play. In 11 retained games the partner is a verified bot (1) or a flagged account (10); without them the gap is +26.0~pp. Screening at the game level only (removing verified-bot games but keeping the plays of flagged accounts) gives 598 games and a gap of +27.5~pp. The platform's community publishes a convention system, so we expect this corpus to lie toward the upper end of convention use; it is not a sample of casual play. Thirteen games use a rule-changing option (gap without them +26.3~pp; \suppapp{app:sensitivity}).

\textbf{Human-AI}: HanabiData \citep{eger2017intentional}, 2{,}040 games of 228 participants with one of three AIs, with 15{,}472 human plays in 2{,}035 games and 13{,}577 AI plays. Outer plays known-safe cards and gives informative hints, Intentional chooses hints by the action it intends the human to take, and Full also interprets the hints it receives as play or discard signals (\supptab{tab:hd_agents}). Our main estimates are pooled over plays (play-weighted). Two participants supplied 53\% of the human plays, and the median participant played 2 games, so we also report equal-participant values (each participant's gap weighted equally); participant-cluster bootstrap intervals are asymmetric for the same reason. The AI type was drawn at random for every game \citep{eger2017intentional}; 122 participants met one type, 58 two and 48 all three (\suppapp{app:sensitivity}). Games recorded with one of five preset decks (seeds 1 to 5, the decks the source study used for participants' first games; 240 \emph{preset-deck games}) violate (i). The source interface departed from standard rules: the AI gave itself 15 hints in four games, and the interface allowed discards at eight information tokens. \suppapp{app:sensitivity} gives the estimates without the preset-deck and self-hint games.

\textbf{AI-AI} (simulated): all AI-AI games are our own simulations. They use our Python reimplementations of six rule-based agents of \citet{walton2017evaluating} (IGGI, Internal, Outer, Piers, Van den Bergh, Flawed), written from the Java code distributed with HOAD \citep{sarmasi2021hoad}, plus a local baseline (Simple). We ran 100 games for each of 49 ordered pairings with fixed seeds (62{,}479 plays). The simulator enforces the standard rules (every hint must touch a card). Each agent receives only its own observation, so the agents meet (iii) by construction; a hidden-state resampling test supports this (\suppapp{app:agents}). For the five agents with a published two-player self-play score (IGGI, Internal, Outer, Piers, Van den Bergh), our 100-game means lie within two standard errors of it. In 1{,}000 further self-play games per agent (not part of the corpus), Van den Bergh's mean is 0.3 points (about three standard errors) below its published score (\suppapp{app:agents}). The name Outer denotes two different implementations: ours, after \citet{walton2017evaluating}, and the HanabiData agent of \citet{eger2017intentional}. We therefore compare partners only within each corpus.

\textbf{OBL}: the released checkpoints of \citet{hu2021offbelief}, 1{,}000 games per condition (\S\ref{sec:obl_results}).

\subsection{Hint-Quality Metrics}
\label{sec:failure_metrics}

\textbf{Hint playability} is the pooled fraction of hint-touched cards that are playable, so each hint is weighted by the cards it touches. The \textbf{candidate-elimination fraction} is the per-hint mean of $(C_\text{before} - C_\text{after})/C_\text{before}$, where $C$ sums the candidate identities over all cards of the target's hand. The per-hint mean playability and a touched-cards-only variant are in \suppapp{app:hintquality}.

\section{Results}

Gaps are pooled over plays unless stated otherwise.

\subsection{Convention Gap Across Cooperation Settings}
\label{sec:gap_exists}

\begin{figure*}[t]
\centering
\includegraphics[width=0.66\textwidth]{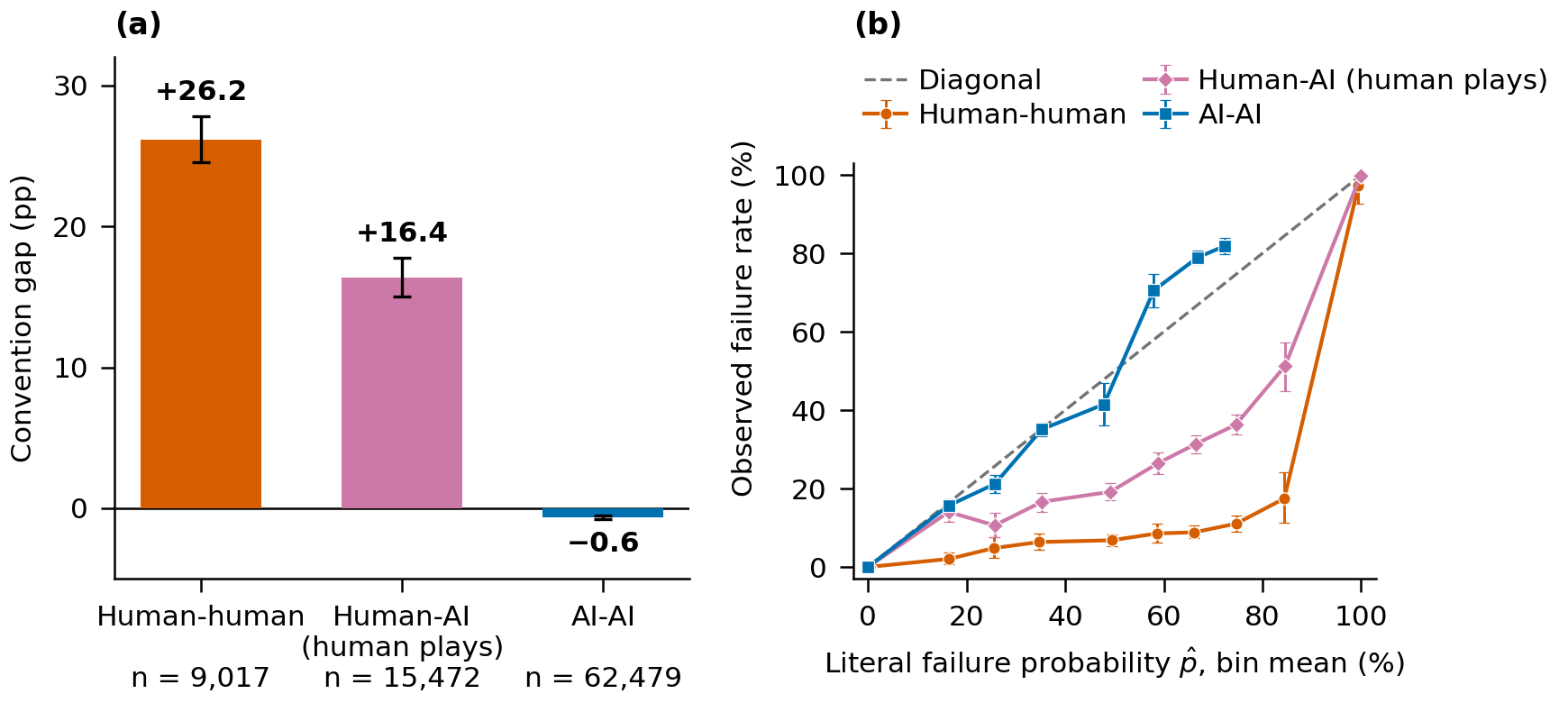}
\caption{Convention gap and calibration by setting. Human-AI values use human plays only. \textbf{(a)} Convention gap by setting. Error bars are 95\% cluster-bootstrap CIs with acting players (human-human), participants (human-AI) or games (AI-AI) resampled; the AI-AI interval over the seven acting agents is in Table~\ref{tab:headline}. \textbf{(b)} Observed failure rate against mean $\hat{p}$ in ten bins of $\hat{p}$ ($[0, 0.1]$, then $(0.1k, 0.1(k{+}1)]$); bins with fewer than 10 plays omitted; error bars are 95\% game-cluster bootstrap CIs (2{,}000 resamples). On the dashed diagonal the observed rate equals $\hat{p}$.}
\Description{Panel a: bar chart of the convention gap for human-human, human-AI and AI-AI play. Panel b: observed failure rate against mean literal failure probability in ten bins, with the diagonal.}\label{fig:calibration}
\end{figure*}

Human pairs failed at plays about 26~pp less often than the literal reference predicted (Table~\ref{tab:headline}, Figure~\ref{fig:calibration}a). The human-AI gap on human plays was +16.4~pp play-weighted and +14.6~pp with equal participant weights (\suppapp{app:sensitivity}). The positive gaps remained after clustering, with wider intervals. Players linked by shared games form 65 connected groups. With each group as one unit, a heuristic $t$ interval with effective degrees of freedom was $[+23.1, +29.2]$~pp (\suppapp{app:stats}).

The AI-AI gap was near zero overall ($-$0.64~pp). Its interval excluded zero with plays, games or the 49 pairings resampled. It included zero with the seven acting agents as clusters ($[-2.18, +0.16]$~pp), which are too few for a reliable interval. Most AI-AI plays were pinned (\S\ref{sec:method_convgap}): plays with $\hat{p} = 0$ were 81.7\% of AI-AI plays, against 42.1\% and 35.1\% in the two human settings (Table~\ref{tab:headline}; \supptab{tab:pinned_zero}).

Simple, IGGI, Internal, and our Outer played only such cards. On plays with $\hat{p} > 0$ the gaps were +45.2 (human-human), +25.2 (human-AI) and $-$3.5~pp (AI-AI). Among the agents that played cards with $\hat{p} > 0$, Piers and Van den Bergh had gaps near zero on such plays (+0.04~pp $[-1.5, +1.6]$ and +1.4~pp $[-0.3, +3.1]$; game-cluster bootstrap CIs). Their partners hint playable cards (rule sets in \suppapp{app:agents}), so (ii) fails in this corpus as well; the near-zero values describe the plays these two agents select (\suppapp{app:convgap_by_subject}). The exact null control (\S\ref{sec:obl_results}) tests the estimator directly. Flawed's was $-$7.4~pp $[-8.6, -6.2]$. Over all its plays Flawed failed at 47.1\% against a mean $\hat{p}$ of 41.8\%, the sign opposite to the human gap (\suppapp{app:convgap_by_subject}).

Figure~\ref{fig:calibration}b shows calibration in ten equal-width bins of $\hat{p}$ (values in \supptab{tab:calibration_bins}). For $\hat{p} \le 0.5$, AI-AI points lay between 6.4~pp below the diagonal and on it. These bins held 93\% of AI-AI plays (gap +0.2~pp), most of them at $\hat{p} = 0$. The three bins between 0.5 and 0.8 lay 10 to 13~pp above the diagonal, and every play in them was Flawed's. Human-human points lay below the diagonal in every bin above the first. At $\hat{p} \in (0.7, 0.8]$ the human-human failure rate was 11.0\%, a 64~pp departure. Skill strata and a logistic recalibration are in \suppapp{app:skill} and \suppapp{app:logistic}.

\begin{table*}[t]
\centering
\caption{Convention gap by setting: mean literal failure probability $\hat{p}$ of the played cards minus their observed failure rate, pooled over plays. Intervals are 95\% percentile bootstrap CIs (10{,}000 resamples) with games or clusters resampled. Clusters are acting players (human-human), participants (human-AI), and acting agents (AI-AI). Last two columns: share of plays with $\hat{p} = 0$ (gap pinned at zero) and gap on the other plays (cluster CI; game CI for AI-AI).}
\label{tab:headline}
\small
\setlength{\tabcolsep}{4pt}
\resizebox{\textwidth}{!}{\begin{tabular}{lrrrcccllcl}
\toprule
Setting & $N$ plays & Games & Clusters & Mean $\hat{p}$ & Failure & Gap (pp) & Game CI & Cluster CI & $\hat{p}{=}0$ & Gap, $\hat{p}{>}0$ \\
\midrule
Human-human & 9{,}017 & 425 & 155 & 31.4\% & 5.3\% & +26.16 & [+25.32, +27.01] & [+24.54, +27.80] & 42.1\% & +45.2 [+43.2, +47.1] \\
Human-AI (human plays) & 15{,}472 & 2{,}035 & 228 & 39.1\% & 22.7\% & +16.38 & [+15.72, +17.06] & [+15.04, +17.76] & 35.1\% & +25.2 [+21.8, +28.7] \\
AI-AI & 62{,}479 & 4{,}900 & 7 & 8.0\% & 8.6\% & $-$0.64 & [$-$0.79, $-$0.49] & [$-$2.18, +0.16] & 81.7\% & $-$3.5 [$-$4.3, $-$2.7] \\
\bottomrule
\end{tabular}}
\end{table*}

\subsection{The Gap by Number of Hints Touching the Card}
\label{sec:hint_localization}

\begin{figure}[t]
\centering
\includegraphics[width=\columnwidth]{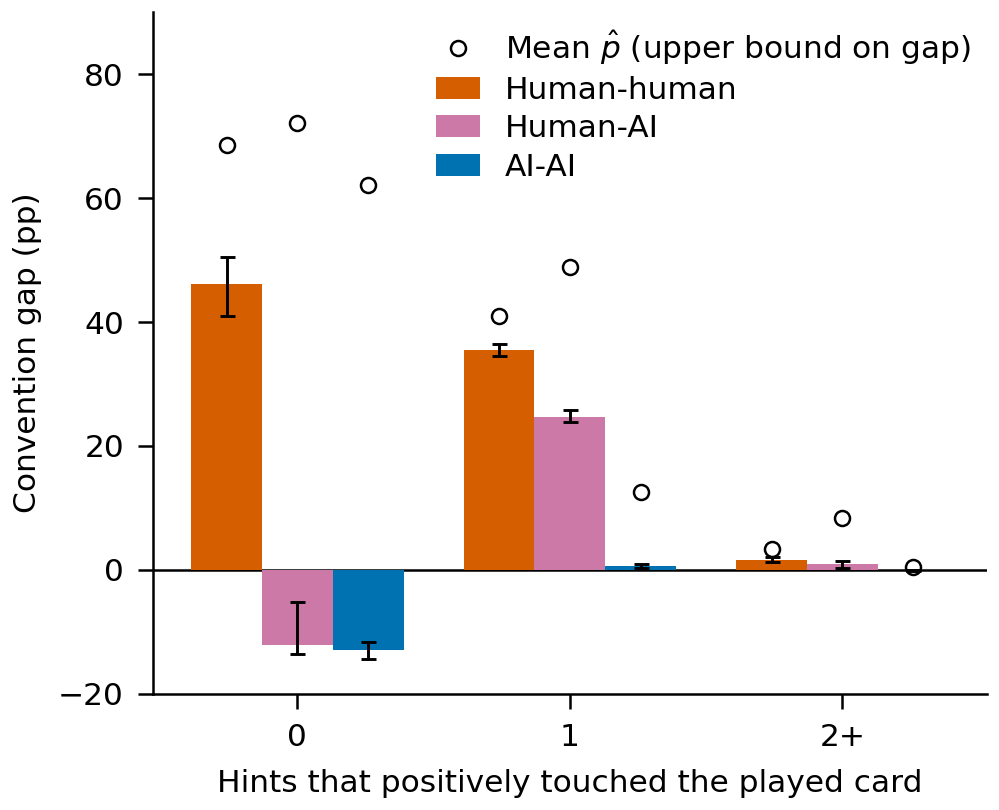}
\caption{Convention gap by the number of hints that positively touched the played card (human-AI: human plays only). Error bars are 95\% cluster-bootstrap CIs (10{,}000 resamples; games for human-human and AI-AI, participants for human-AI). Open circles give the mean $\hat{p}$ of each stratum, the largest value the gap can take. Values and $N$: \supptab{tab:hintcount}.}
\Description{Grouped bar chart of the convention gap by the number of hints that touched the played card (zero, one, two or more) for the three settings, with open circles for the mean literal failure probability.}\label{fig:hintcount_bars}
\end{figure}

We stratified plays by the number of hint actions that had positively touched the played card: zero-touch, one-touch and two-or-more-touch (2+) plays (Figure~\ref{fig:hintcount_bars}; one color and one rank hint count two). The gap of a stratum cannot exceed its mean $\hat{p}$, so we report mean $\hat{p}$ and failure with each gap. On zero-touch plays, human pairs had a mean $\hat{p}$ of 68.5\%, a failure rate of 22.3\%, and a gap of +46.2~pp (484 plays, 5\% of their plays). Human plays with an AI partner had 72.2\%, 84.3\%, and $-$12.2~pp (778 plays; participant-cluster bootstrap CI $[-13.6, -5.1]$), so with AI partners zero-touch cards failed more often than $\hat{p}$ predicts. AI-AI plays had 62.2\%, 75.2\%, and $-$13.0~pp (3{,}748 plays). On one-touch plays the values were 41.1\%, 5.5\%, and +35.6~pp for human-human (65\% of plays). They were 48.9\%, 24.2\%, and +24.7~pp for human-AI, and 12.5\%, 11.9\%, and +0.6~pp for AI-AI. On 2+ plays, mean $\hat{p}$ was 3.3\%, 8.3\%, and 0.5\%, so the ceiling was low; the gaps were +1.6, +0.9, and $-$0.1~pp. One-touch plays carried 23.2~pp of the 26.2~pp pooled human-human gap, and zero-touch plays 2.5~pp.

Zero-touch cards were not free of literal constraints. A negative constraint from a hint to another card applied to 78\% of human-human, 45\% of human-AI, and 53\% of AI-AI zero-touch plays. The strata also differ in risk, so the raw comparison across strata is confounded by $\hat{p}$. Among plays with $\hat{p} > 0.4$, standardized to a common distribution over bins of $\hat{p}$ (\supptab{tab:hint_standardized}), the human-human gap was +43.8~pp at zero touches and +54.9~pp at one (difference $-$11.1~pp, CI $[-17.2, -6.1]$). The human-AI gaps were $-$12.1 and +35.5~pp (difference $-$47.6~pp, CI $[-52.2, -40.1]$). At equal risk the one-touch gap was the larger in both. Over the $\hat{p}$ bins that one-touch and 2+ plays share, the human-human gaps were +25.9 and +22.2~pp (the latter on 137 plays with $\hat{p} > 0$). In human pairs the small raw gap at 2+ is therefore mostly the ceiling. In human-AI play the 2+ gap stayed lower at equal risk (+5.4 vs.\ +18.5~pp).

\subsection{Per-AI Convention Gap Within Human-AI Play}
\label{sec:partner_dependence}

Within human-AI play, the human gap varied by AI partner (Table~\ref{tab:partner_main}; intervals in \supptab{tab:convgap_agent}). Mean $\hat{p}$ on the plays humans selected was similar across partners (37.9--40.6\%). The AI type was drawn at random for each game (\S\ref{sec:data}), but the plays compared are those the humans selected with each partner. The play-weighted gap was 17.9~pp higher with Full than with Outer (participant-cluster bootstrap 95\% CI [+16.4, +19.1]). It was 14.0~pp higher with Intentional than with Outer ([+12.5, +16.1]). Without the two participants with the most plays (\S\ref{sec:data}) the differences were 17.6 and 14.9~pp. After each partner's plays were reweighted to the pooled distribution over cells of $\hat{p}$ bin, hint stratum (zero, one, 2+ touches; \S\ref{sec:hint_localization}) and deck stage (terciles of deck size), they were 17.6 and 17.1~pp (\supptab{tab:partner_standardized}). The equal-participant gaps showed the same pattern (Full and Intentional 14.5 and 14.8~pp above Outer). The sign of Full minus Intentional depended on the weighting ($+$3.9~pp play-weighted, $+$0.5~pp standardized, $-$0.3~pp equal-participant), so we claim no ordering of these two.

The partner difference persisted within participants (Table~\ref{tab:within_main}). Both contrasts against Outer were positive at both thresholds ($+$16.0 to $+$19.7~pp). They remained so without the preset-deck games ($+$21.1 and $+$18.1~pp at $\geq$1 game; \supptab{tab:within}). Pairing removes stable differences between participants, but not learning or order effects (\suppapp{app:within}).

\begin{table}[t]
\centering
\caption{Human plays by AI partner: $N$ plays, mean $\hat{p}$, failure rate, and gap (pp), play-weighted (PW) and equal-participant (EP). Last column: each AI's PW gap on its own plays ($n$). $^{\ast}$Pinned: only 4 and 2 plays had $\hat{p} > 0$.}
\label{tab:partner_main}
\footnotesize
\setlength{\tabcolsep}{3pt}
\begin{tabular}{lrccccc}
\toprule
Partner & $N$ & $\hat{p}$ & Fail & PW & EP & AI own ($n$) \\
\midrule
Full & 4{,}769 & 38.5\% & 14.4\% & +24.1 & +19.4 & +16.9 (5{,}965) \\
Intentional & 5{,}122 & 37.9\% & 17.6\% & +20.2 & +19.7 & 0.0$^{\ast}$ (4{,}979) \\
Outer & 5{,}581 & 40.6\% & 34.4\% & +6.2 & +4.9 & 0.0$^{\ast}$ (2{,}633) \\
\bottomrule
\end{tabular}
\end{table}

\begin{table}[t]
\centering
\caption{Equal-participant paired contrasts of the human gap, for participants with at least 1 or 3 games with both partners. CI: 95\% $t$ interval. Sign: participants with positive\,/\,negative contrast; $p$: two-sided sign test.}
\label{tab:within_main}
\footnotesize
\setlength{\tabcolsep}{3pt}
\begin{tabular}{llrlcc}
\toprule
Games & Contrast & $n$ & $\Delta$gap (pp) [CI] & Sign & $p$ \\
\midrule
$\geq$1 & Full $-$ Outer & 69 & +16.0 [+12.2, +19.8] & 57/12 & $3.7\times10^{-8}$ \\
$\geq$1 & Intentional $-$ Outer & 64 & +18.8 [+14.7, +22.9] & 58/6 & $9.0\times10^{-12}$ \\
$\geq$1 & Full $-$ Intentional & 69 & $-$3.4 [$-$7.8, +0.9] & 32/37 & 0.63 \\
\midrule
$\geq$3 & Full $-$ Outer & 22 & +19.7 [+16.4, +23.1] & 22/0 & $4.8\times10^{-7}$ \\
$\geq$3 & Intentional $-$ Outer & 21 & +18.4 [+14.0, +22.7] & 20/1 & $2.1\times10^{-5}$ \\
$\geq$3 & Full $-$ Intentional & 18 & +1.8 [$-$2.4, +6.0] & 13/5 & 0.096 \\
\bottomrule
\end{tabular}
\end{table}

The gap on each AI's own plays also varied. Full's was $+$16.9~pp play-weighted ([+13.5, +18.5]) and $+$10.1~pp equal-participant ([+7.1, +13.0]), at a mean $\hat{p}$ of 32.3\% ($+$17.4 and $+$12.5~pp without the preset-deck games). Intentional and Outer had gaps near zero because they almost never played a card with $\hat{p}>0$ (4 of 4{,}979 and 2 of 2{,}633 plays; \supptab{tab:full_mechanism}). On its plays with $\hat{p}>0$, Full's gap was $+$27.5~pp ([+21.2, +30.2]; \supptab{tab:pinned_zero}).

Hint metrics over 20{,}088 AI hints separated Outer from the other two (participant-cluster bootstrap CIs; \suppapp{app:hintquality}). The cards touched were playable for 43.5\% [41.2, 44.6] with Outer, 57.4\% [56.2, 60.0] with Full and 59.3\% [57.2, 60.7] with Intentional. When the human's next action was a play, it failed 30.8\% [26.7, 35.5] of the time after an Outer hint, 7.6\% [6.3, 9.0] after Full and 8.3\% [5.8, 11.2] after Intentional. Outer's hints touched more cards (1.78 per hint against 1.52 for each of the others). They eliminated the largest share of candidate identities over the whole hand (39.0\% against 36.5\% and 36.2\%) but the smallest on the touched cards (68.4\% against 74.1\% for each of the others).

\subsection{One-Step Card-Policy Contrasts and Hint Timing}
\label{sec:receiver}

On each play turn we computed $\hat{p}$ for every card in hand. We compared the card played with the card another one-step rule would have picked. The rules were the lowest-$\hat{p}$ card and the most recently hinted card (the newest card if none was hinted). Both break ties by the newest card. Ties are common: the last hint touched more than one card in hand on 33--50\% of turns. Each substitution keeps the player's earlier actions and the decision to play, so the contrasts describe the observed turns and are not effects of deploying another policy. Table~\ref{tab:receiver} gives mean $\hat{p}$, failure and gap for each rule (further rules and the oldest-card and other tie rules: \suppapp{app:receiver}).

\begin{table}[t]
\centering
\caption{One-step card-policy contrasts on observed play turns (pooled over play turns; H-AI: human plays in human-AI games, by AI partner). Mean $\hat{p}$ (\%), failure (\%) and gap (pp) for the card played, the lowest-$\hat{p}$ card and the most recently hinted card (newest-card tie-breaks). Gap intervals and intervals of the contrasts are in \suppapp{app:receiver}.}
\label{tab:receiver}
\footnotesize
\setlength{\tabcolsep}{2pt}
\begin{tabular}{lrrrrrrrrr}
\toprule
 & \multicolumn{3}{c}{Played} & \multicolumn{3}{c}{Lowest $\hat{p}$} & \multicolumn{3}{c}{Last hinted} \\
Setting & $\hat{p}$ & Fail & Gap & $\hat{p}$ & Fail & Gap & $\hat{p}$ & Fail & Gap \\
\midrule
Human-human & 31.4 & 5.3 & +26.2 & 27.1 & 15.6 & +11.5 & 47.9 & 26.7 & +21.3 \\
H-AI: Full & 38.5 & 14.4 & +24.1 & 32.0 & 22.5 & +9.5 & 43.7 & 17.2 & +26.5 \\
H-AI: Intent. & 37.9 & 17.6 & +20.2 & 30.9 & 23.0 & +7.9 & 43.0 & 19.8 & +23.1 \\
H-AI: Outer & 40.6 & 34.4 & +6.2 & 33.4 & 35.8 & $-$2.4 & 50.1 & 39.8 & +10.3 \\
AI-AI & 8.0 & 8.6 & $-$0.6 & 8.0 & 8.4 & $-$0.4 & 28.0 & 27.0 & +1.0 \\
\bottomrule
\end{tabular}
\end{table}

The gap of the played card minus the gap of the lowest-$\hat{p}$ card equals their difference in mean $\hat{p}$ minus their difference in failure. In human-human play (9{,}017 turns) this gap difference was +14.7~pp [+13.7, +15.7] (game-cluster bootstrap 95\% CI; [+13.0, +16.5] with acting-player clusters). The played cards had higher $\hat{p}$ (+4.4~pp [+4.1, +4.7]) and failed less often ($-$10.3~pp [$-$11.2, $-$9.4]). Across tie rules the difference ranged from +14.7 to +19.1~pp. With AI partners it was +14.6 (Full), +12.3 (Intentional) and +8.6~pp (Outer). With Outer it came mostly from higher $\hat{p}$ (+7.2~pp), and the failure difference was $-$1.4~pp [$-$3.1, +0.1] (participant-cluster bootstrap CI), an interval that includes zero. The most recently hinted card failed more often than the played card in every human setting under every tie rule (Full: 17.2\% vs.\ 14.4\%; human-human: 26.7\% vs.\ 5.3\%). With every AI partner the last-hinted card's gap exceeded the played card's under newest-card ties (Full: +26.5 vs.\ +24.1~pp) and fell below it under oldest-card ties (Full: +22.6~pp). In human-human play the last-hinted card's gap was below the played card's (+21.3 vs.\ +26.2~pp). Across card rules, a larger gap therefore does not mean fewer failures. The AI-AI agents' played card coincided with the lowest-$\hat{p}$ card on 91.9\% of 62{,}479 turns (gap difference $-$0.2~pp).

We modeled which card was played with a conditional logit over the cards in hand, with each play turn as one stratum (\suppapp{app:receiver}). Each card had two groups of covariates. Cards of equal $\hat{p}$ differ in rank and future value, so $\hat{p}$ is not a sufficient statistic of the literal information about a card. Each card's $\hat{p}$ therefore entered flexibly (bins, a linear term, and indicators for $\hat{p}=0$ and for the hand minimum). The timing cues were the turns since the card was last hinted and whether it was hinted on the previous turn. Relative to the model with the $\hat{p}$ terms only, the timing cues raised the held-out log-likelihood per turn by 0.21~nats across games and 0.20 across players in human-human play (from $-$0.69; uniform choice gives $-$1.61). In human-AI play the gains were 0.39 across games and 0.38 across participants (from $-$0.97 and $-$0.98). With hint count, newest card and hand position also in the model, the gains were 0.08 (human-human) and 0.16 to 0.19 (human-AI). In that model the coefficient per turn since the hint was $-$0.26 [$-$0.32, $-$0.21] in human-human and $-$0.38 [$-$0.47, $-$0.36] in human-AI play (acting-player or participant cluster percentile bootstrap, 95\%). The intervals for ``hinted last turn'' and, in human-AI play, ``newest card'' included zero.

\subsection{Game Score and the Gap Answer Different Questions}
\label{sec:score_missed}

Under strict scoring, mean scores were 18.5 in human-human (SD 9.6, $n = 425$ games), 10.7 in AI-AI (SD 6.8, $n = 4{,}900$) and 7.2 in human-AI games (SD 8.5, $n = 2{,}040$; \suppapp{app:gamescore}). Mean score was lower in human-AI than in AI-AI games, whereas the gap was larger (+16.4 vs.\ $-$0.6~pp). A third misplay ended 57.4\% of human-AI and 26.0\% of AI-AI games (all of the latter in games with Flawed). Under partial scoring the means were 20.3, 11.7 and 12.1. The human-AI minus AI-AI difference was $-$3.5 points under strict scoring (95\% CI [$-$5.5, $-$1.7], participants and games resampled) and +0.4 [$-$1.0, +1.7] under partial scoring, so the human-AI deficit in score appeared only under strict scoring. The gap does not use the end-of-game score and is the same under both conventions. Both pooled quantities average over the three AI partners: by partner, the play-weighted human-AI gap ranged from +6 (Outer) to +24~pp (Full; \S\ref{sec:partner_dependence}) and the strict score from 2.9 (Outer) to 12.1 (Intentional). Score answers a task-performance question; the gap answers a reference-calibration question. They are complementary.

\subsection{Convention Gap Across the OBL Reasoning Hierarchy}
\label{sec:obl_results}

OBL \citep{hu2021offbelief} trains a hierarchy of agents. OBL1 forms its beliefs as if past actions had come from a uniformly random policy, so it is trained to read hints approximately by their literal content (a uniformly random legal policy still carries the hand information of footnote~\ref{fn:legal}). Each level $k{+}1$ is trained with beliefs induced by level $k$. OBL's theory neither predicts that the gap grows with level nor guarantees (ii) for OBL1, so we use the hierarchy as convergent evidence. We ran the released checkpoints in three conditions. Self-play used one checkpoint per level (1{,}000 games each). OBL1 cross-play used the ten pairings among five independently trained OBL1 models (100 games each). Inter-level play paired each two of the five levels (ten pairings) on 1{,}000 deals, played in both seatings. Every game was replayed through the same engine (\suppapp{app:obl}).

\begin{table}[t]
\centering
\caption{Convention gap (mean $\hat{p}$ minus failure rate) on OBL plays; 1{,}000 games per row, one checkpoint per level in self-play. 95\% CIs: game-level bootstrap, conditional on the checkpoints. $^{\dagger}$Ten pairings among five OBL1 models, pooled.}
\label{tab:obl}
\footnotesize
\setlength{\tabcolsep}{3pt}
\begin{tabular}{lrccl}
\toprule
Condition & Plays & Mean $\hat{p}$ & Failure & Gap (pp) [95\% CI] \\
\midrule
OBL1 self-play & 22{,}791 & 7.8\% & 6.3\% & +1.57 [+1.33, +1.81] \\
OBL2 self-play & 24{,}204 & 15.4\% & 3.0\% & +12.40 [+12.10, +12.71] \\
OBL3 self-play & 24{,}538 & 19.1\% & 2.2\% & +16.91 [+16.60, +17.21] \\
OBL4 self-play & 24{,}731 & 22.3\% & 2.5\% & +19.88 [+19.54, +20.21] \\
OBL5 self-play & 24{,}871 & 24.1\% & 2.4\% & +21.68 [+21.34, +22.02] \\
OBL1 cross-play$^{\dagger}$ & 22{,}616 & 7.7\% & 6.5\% & +1.22 [+0.98, +1.45] \\
\bottomrule
\end{tabular}
\end{table}

\begin{figure}[t]
\centering
\includegraphics[width=0.78\columnwidth]{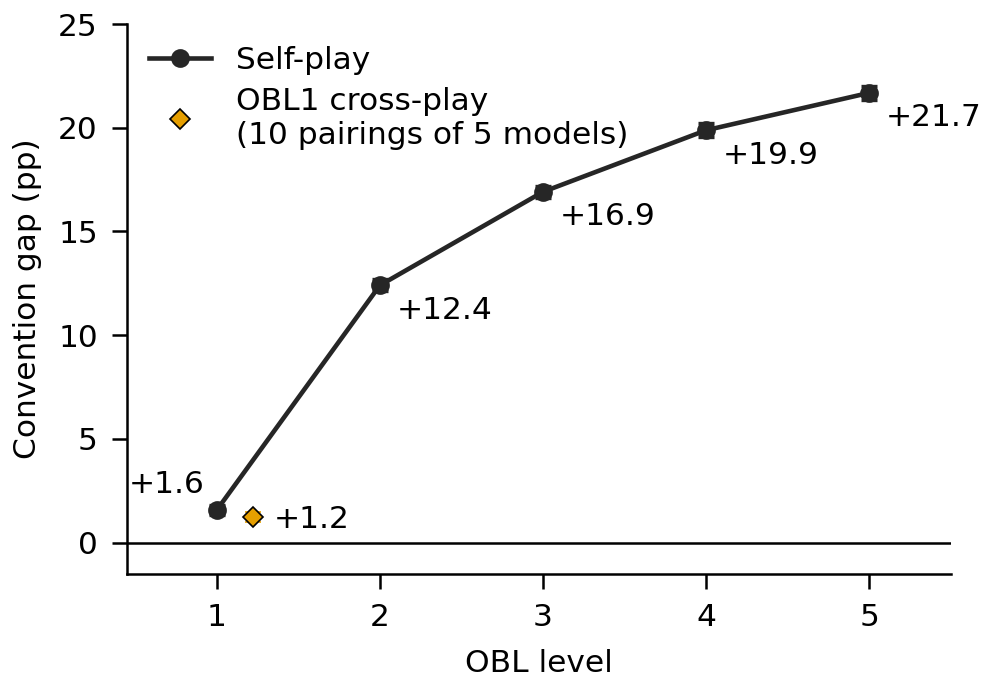}
\caption{Gap by OBL level in self-play (circles) and in OBL1 cross-play (diamond, offset to the right of level 1). The 95\% game-bootstrap CIs are narrower than the markers. OBL5 is a released checkpoint that \citet{hu2021offbelief} do not report; levels 1--4 alone show the same order.}
\Description{Convention gap for OBL levels 1 to 5 in self-play, rising from about 1.6 to 21.7 percentage points, and one point for OBL1 cross-play near level 1.}\label{fig:obl_main}
\end{figure}

The gap rose monotonically, from +1.57~pp at OBL1 (interval excluding zero; +1.22~pp in cross-play) to +12.40 to +21.68~pp at OBL2 to OBL5 (Table~\ref{tab:obl}, Figure~\ref{fig:obl_main}). Higher levels chose plays with greater $\hat{p}$ (7.8\% to 24.1\%) while observed failure stayed at 2--3\% from OBL2 on, and the order held at common hint-stratum weights (\suppapp{app:obl}). An aggregate gap near zero can hide offsetting discrepancies. With OBL1 in the first seat, its own plays with an OBL5 partner had +0.1~pp [$-$0.2, +0.5] overall. By hint count the gap was $-$19.5, +2.5 and +0.3~pp on zero-, one- and two-or-more-touch plays (4\%, 31\% and 65\% of plays), and each interval excluded zero. With OBL1 in the second seat the overall gap was also near zero ($-$0.1~pp) and each stratum interval again excluded zero (\suppapp{app:obl}).

As an exact null control for Proposition~\ref{prop:zero}, we simulated partners that satisfy (ii) by construction. Across 12 combinations of partner and receiver policy (cells; 2{,}059{,}997 plays) the pooled gap was $-$0.01~pp (SE 0.02), and every cell's 95\% interval included zero (largest $|z| = 1.84$). Of 74 strata (hint count and $\hat{p}$ bin, $\geq$200 plays), 7 had $|z| > 1.96$ (3.7 expected by chance). One of them, the zero-touch stratum of one cell, was positive in six further runs with independent seeds (+0.42~pp [+0.18, +0.67]; +0.36~pp, SE 0.11, with a run of a separately written game loop) but not in a prespecified replication of 200{,}000 games in which $\hat{p}$ was computed from the action log by an independent implementation ($-$0.09~pp [$-$0.33, +0.16]; \suppapp{app:nullcontrol}). The excess in the earlier runs remains unexplained. With partners that hint playable cards and so violate (ii), two of the same receivers gave gaps of +1.15 to +15.50~pp (four cells). The control tests the estimator under the designed null only; the replay of logged games is checked against the recorded final scores (\suppapp{app:pipeline}).

\section{Discussion}
\label{sec:discussion}

\textbf{What the gap does and does not show.} The gap is a signed calibration residual against the literal reference, in percentage points of failure. Under a uniform shuffle~(i) and a receiver acting on the transcript alone~(iii), a nonzero population gap is evidence that partner choices were informative beyond their literal content (Proposition~\ref{prop:zero}). In the exact null control, which meets all three conditions by construction, the pooled gap was $-$0.01~pp over 2.06 million plays (\S\ref{sec:obl_results}). The hanab.live logs cannot exclude side information outside the transcript (chat, timing, a previously seen shuffle), which would violate~(iii). The gap is not an amount of information, and it can vanish by cancellation (\S\ref{sec:obl_results}). A larger gap is not better play: Full's own plays had a gap of +16.9~pp and failed 15.4\% of the time, whereas Intentional and Outer almost never played a card with $\hat{p} > 0$ and almost never failed. In the terms of \citet{lowe2019pitfalls}, a nonzero gap under (i) and (iii) is evidence of positive signaling: the partner's choices depended on hidden cards beyond their literal content. It does not establish positive listening.

\textbf{What the receiver side shows.} The one-step contrasts and the card-choice model of \S\ref{sec:receiver} are descriptive evidence about the receiver's side. The gap does not distinguish a receiver who reads hints as instructions, a fixed play rule the partner has adapted to, and unilateral statistical adaptation to a partner. Separating these requires intervention, such as partner-swap designs.

\textbf{Where the human departure from the literal reference was observed.} Most of the human-human gap came from one-touch plays. Zero-touch plays, though they failed far less often than $\hat{p}$ predicted, were 5\% of plays, and 78\% of them carried negative constraints (\S\ref{sec:hint_localization}). Candidate conventions include playing the newest card, the good-touch principle (hint only cards that will be played) and hint-timing inference, which overlap with the ``hidden rules'' of \citet{sidji2023hidden}. Two selection effects bear on these values. The completed-game filter conditions on how a game ended; the games that did not reach a terminal state had a gap of +28.5~pp (the 164 of these 420 games that keep plays after the same bot exclusion; \suppapp{app:robustness}). Skill strata are defined from final scores, so within-stratum gaps are descriptive (\suppapp{app:skill}).

\textbf{Partner dependence and intent-based play.} The human gap was higher with Full and with Intentional than with Outer under every weighting, and both contrasts persisted within participants (\S\ref{sec:partner_dependence}, Table~\ref{tab:within_main}). We claim no ordering of Full and Intentional. Full's positive own-play gap is consistent with its design, which interprets received hints as play signals \citep{eger2017intentional}. Intentional uses intent reasoning only to choose hints. The gap alone does not identify the mechanism. Whether the gap disappears without the intent-inference module requires an ablation.

\textbf{Implications.} On plays with $\hat{p} > 0$, the rule-based AI-AI agents, pooled, showed no positive gap ($-$3.5~pp), whereas human pairs showed a large one (+45.2~pp). A positive gap is not specific to human play: OBL agents reached +21.7~pp in self-play (\S\ref{sec:obl_results}), and a rule-based receiver that plays the hinted card reached +15.5~pp with a partner that hints playable cards (\suppapp{app:nullcontrol}). The gap depends on both players: with human receivers it differed by AI partner within the same participants. We therefore suggest that evaluations of agents intended for human partners report calibration against a literal reference on human-partnered play. These statements are scoped to the corpora analyzed here.

\textbf{Porting the diagnostic.} The recipe (decision points with observable success, a specified literal state at each, observed against predicted failure) transfers to other tasks, as with RSA's literal listeners \citep{frank2012predicting,goodman2016pragmatic} and corpus-specific wording baselines \citep{sivan2025information}. The exactness does not; Hanabi can serve as a calibration anchor for approximate baselines.

\textbf{Limitations.} The study is observational, not causal. The human-AI data cover three AI types from a single study. Two of its 228 participants supplied 53\% of the human plays; the gap was +14.6~pp [+13.0, +16.1] with equal participant weights and +16.8~pp [+15.2, +18.2] without those two (\suppapp{app:sensitivity}). Human-human plays are dependent within connected groups of players (\S\ref{sec:gap_exists}). The AI-AI games were simulated with our reimplementations of the agents, which were checked against published mean scores (\S\ref{sec:data}), not action by action. The OBL evidence rests on one checkpoint per level. The three settings differ in platform and population, so differences between settings are confounded with provenance.

\section{Conclusion}

We introduced the convention gap, a signed calibration diagnostic against an exact literal reference whose population gap is zero under conditions (i)--(iii) of Proposition~\ref{prop:zero}. It was large for human pairs and for humans partnered with the two intent-reasoning AIs, near zero in AI-AI play, dependent on the AI partner within the same participants, and increasing across the OBL hierarchy. Because the gap depended on both players, we suggest that evaluations of agents intended for human partners report this calibration on human-partnered play.

\paragraph{Reproducibility.} Code, the simulator with its seeds and generation manifest, and the processed play records are in the public repository\\ \url{https://github.com/dockmfgit/hanabi-convention-gap}\\ (archived on Zenodo, doi:10.5281/zenodo.21975883; use the release that matches this version). For OBL we provide scripts and per-game statistics, from which every OBL estimate and interval can be recomputed (\suppapp{app:obl}). The use of AI-assisted tools is described in \suppapp{app:ai_use}. Further descriptive analyses are in \suppapp{app:failure} (failure categories), \suppapp{app:crossmatrix} (cross-play matrices), \suppapp{app:learning} (learning), \suppapp{app:llm_modern} (a port of Full with LLM partners) and \suppapp{app:further}.

\bibliographystyle{plainnat}
\bibliography{refs}

\clearpage
\appendix
\renewcommand{\thetable}{A\arabic{table}}
\renewcommand{\thefigure}{A\arabic{figure}}
\setcounter{table}{0}
\setcounter{figure}{0}
\renewcommand{\theequation}{A\arabic{equation}}
\setcounter{equation}{0}
\makeatletter
\providecommand{\FloatBarrier}{\par\begingroup\let\@elt\relax\edef\@tempa{\@botlist\@deferlist\@dbldeferlist}\ifx\@tempa\@empty\else\clearpage\fi\endgroup}
\makeatother

\section{Simulated AI-AI Agents: Provenance, Rules and Fidelity}
\label{app:agents}

\subsection{Provenance}

The AI-AI corpus consists of games that we simulated. Our pure-Python engine played our own Python reimplementations of seven rule-based agents. Six agents (IGGI, Internal, Outer, Piers, Van den Bergh, Flawed) are production-rule agents of \citet{walton2017evaluating}. We reimplemented them from the Java code that is distributed with HOAD \citep{sarmasi2021hoad}. The seventh, Simple, is a baseline written for this study that plays only certainly playable cards; it has no upstream counterpart. Only four of the seven (IGGI, Outer, Piers, Van den Bergh) belong to the published HOAD dataset. Internal and Flawed are shipped in the HOAD repository but are absent from that dataset. Each agent follows a fixed priority list of play, hint and discard rules with no learned parameters. The first rule that yields an action is taken (Table~\ref{tab:agents}).

For two agents the Java preset differs from the rule listing printed by \citet{walton2017evaluating}, and we followed the code. The Van den Bergh preset (\texttt{vdb-paper}) omits the rule that tells about a useless card. The IGGI preset ends with a random hint that the printed listing does not have.

\begin{table}[ht]
\centering
\caption{The seven agents of the AI-AI corpus. Rules are listed in priority order; ``safe'' means certainly playable given the agent's own hints and the visible cards. Java sources are factory methods of the \texttt{fossgalaxy/hanabi} code distributed with HOAD. ``Osawa discard'' takes the first own card that is known to be safe to discard from its hinted attributes, as in the Java rule.}
\label{tab:agents}
\small
\begin{tabular}{p{1.9cm}p{3.5cm}p{9.1cm}}
\toprule
Agent & Source & Rule list in brief \\
\midrule
Simple & \raggedright Local baseline (no upstream source) & Play a safe card (lowest possible rank first); give the hint that makes most partner cards newly certainly playable; discard the oldest card. \\
IGGI & \raggedright \texttt{buildIGGIPlayer} \citep{walton2017evaluating}; in HOAD dataset & Play a fully known playable card; play a safe card; hint a playable card's missing rank, else color; Osawa discard; discard oldest; random hint. \\
Internal & \raggedright \texttt{buildInternalState}, a rendering of Osawa's internal-state player \citep{osawa2015solving} & Play a safe card; Osawa discard; hint the partner's first playable card (color or rank at random, may repeat information); random hint; random discard. \\
Outer & \raggedright \texttt{buildOuterState}, a rendering of Osawa's outer-state player \citep{osawa2015solving}; in HOAD dataset & Play a safe card; Osawa discard; hint the first playable card whose holder lacks information (rank first); hint any card with missing information (color first); random discard. \\
Piers & \raggedright \texttt{buildPiersPlayer} \citep{walton2017evaluating}; in HOAD dataset & Play a safe card; with more than one life, play the card most likely playable if that probability is at least 0.6; hint a playable card; with fewer than 4 information tokens, hint a dispensable card; Osawa discard; discard oldest; random hint; random discard. \\
Van den Bergh & \raggedright \texttt{vdb-paper} preset \citep{vandenbergh2017aspects,walton2017evaluating}; in HOAD dataset & With more than one life, play at playability at least 0.6, otherwise play a safe card; discard a certainly useless card; hint a playable card; give the hint with the most new information; discard the card most likely useless. \\
Flawed & \raggedright \texttt{buildFlawedPlayer} \citep{walton2017evaluating} & Play a safe card; play at playability at least 0.25; random hint; Osawa discard; discard oldest; random discard. \\
\bottomrule
\end{tabular}
\end{table}

Piers' first Java rule (play the most likely card when the deck is empty) is ported as written. Its guard, evaluated on the agent's own copy of the state, never holds while the agent has a card, so the rule never acts.

\subsection{What the Simulator Enforces}
\label{app:agents_simulator}

\paragraph{Rules.} The engine runs standard two-player No Variant Hanabi in strict mode: it accepts only the legal actions listed in Appendix~\ref{app:rules} and raises an error on any other action. Where a rule list yields no legal action, a deterministic fallback acts (discard the oldest card if legal, otherwise a color hint on the partner's oldest card). The Java code has no such fallback. In the corpus only Simple used it, by design: at 8 information tokens, when no hint makes a partner card newly certainly playable, Simple gives the color hint on the partner's oldest card.

\paragraph{Observation-only policies.} Each policy receives an immutable observation: the hint knowledge of every hand, the partner's cards, the fireworks (the stacks of played cards), the discard pile, tokens, deck size and turn. The observation contains no own-card identity and no deck order. Because each policy receives only this observation, the AI-AI agents meet condition (iii) of Proposition~\ref{prop:zero} by construction: their decisions depend on the hidden cards only through the observation. A resampling test checks this. It resamples the hidden state (own cards and deck) while holding the observation fixed. No agent changed its action in any of the 1{,}719 sampled states (111{,}642 resampled hidden states in all), and the test flagged each of four deliberately leaky control agents.

\paragraph{Hand scans, seeds and manifest.} Rules that scan a hand do so in fixed table-position order, as in the Java code, where a drawn card takes the slot of the removed card. ``Oldest'' refers to true card age. Each seat owns a random generator whose seed is derived from a base seed, the ordered pairing, the game index and the seat. Deck seeds are derived in the same way, so every ordered pairing plays its own independent deals. The package includes a generation manifest with the seeds and the digests of the game files. The corpus has 49 ordered pairings $\times$ 100 games $=$ 4{,}900 games, 62{,}479 plays and 127{,}063 hints.

\subsection{Fidelity to Published Scores}
\label{app:agents_fidelity}

Published two-player self-play scores exist for five of the seven agents (Table~\ref{tab:agents_fidelity}). The Java framework scores a game by its fireworks sum, and the HOAD table does not state its convention. None of the 500 self-play games of these five agents ended on a strike-out, so the strict and partial scores (Appendix~\ref{app:rules}) coincide for this comparison. The 100-game means of the corpus were within two standard errors of the published values. The largest standardized difference was Piers ($z = 1.8$). We also ran 1{,}000 further self-play games per agent; they are not part of the corpus or the package. Their means were 16.99 (IGGI), 17.26 (Piers), 14.54 (Outer), 13.69 (Van den Bergh) and 11.15 (Internal), with standard errors of about 0.11. Van den Bergh was thus 0.3 points, about three standard errors, below its published 14.0. Flawed and Simple have no published value. Agreement of mean scores is necessary for behavioral equivalence, not sufficient.

\begin{table}[ht]
\centering
\caption{Self-play fidelity check. Estimand: mean fireworks sum over the 100 self-play games of each agent in the AI-AI corpus, with its standard error (SE). Published values are two-player self-play means.}
\label{tab:agents_fidelity}
\small
\begin{tabular}{lccl}
\toprule
Agent & Ours, mean (SE) & Published & Source of published value \\
\midrule
IGGI & 17.04 (0.19) & 17.0 & \citet{sarmasi2021hoad}, Table 1 \\
Piers & 17.64 (0.19) & 17.3 & \citet{sarmasi2021hoad}, Table 1 \\
Outer & 14.41 (0.20) & 14.5; 14.53 & \citet{sarmasi2021hoad}, Table 1; \citet{osawa2015solving} \\
Van den Bergh & 13.99 (0.34) & 14.0 & \citet{sarmasi2021hoad}, Table 1 \\
Internal & 10.99 (0.23) & 10.97 & \citet{osawa2015solving}, his own implementation \\
Flawed & 1.38 (0.18) & none & \\
Simple & 11.41 (0.21) & none & \\
\bottomrule
\end{tabular}
\end{table}

\paragraph{Discard rules.} The four discard rules used by the six reimplemented agents (safe card, useless card, probably useless card and their shared base rule; Table~\ref{tab:agents}) were checked against the Java source text and against an independent emulation of the Java rules, included in the public repository as a test: on every acting state of seeded games and on hand-built edge states, the Python rules select the slot the Java rules select.

\paragraph{What was not verified.} No Java agent was executed. The reimplementation rests on a rule-by-rule reading of the Java sources (\texttt{fossgalaxy/hanabi}, commit \texttt{a99d205}, and the HOAD repository), not on an action-by-action comparison. The last-slot tie-break of two probabilistic rules assumes standard Java iteration order. The Van den Bergh agent was checked against the Java preset only, not against \citet{vandenbergh2017aspects}.

\subsection{The HanabiData Agents and Agent Identity Across Corpora}
\label{app:agents_identity}

The human-AI corpus (HanabiData) was recorded with three agents of Eger et al.'s \texttt{pyhanabi} framework \citep{eger2017intentional}: Full, Intentional and Outer. These are Eger et al.'s implementations; Table~\ref{tab:hd_agents} summarizes their rules from the public code that the paper links. We did not run them, and we analyzed only the recorded games. Outer gives hints that add information the partner lacks. Intentional chooses each hint by the action it intends the human to take. Full also interprets each hint it receives as a play or discard signal. The LLM experiments in Appendix~\ref{app:llm_modern} use a port of Full.

\begin{table}[ht]
\centering
\caption{The three agents of the human-AI corpus (HanabiData). Rules are listed in priority order. ``Known'' means certain from the agent's own hint knowledge alone; these agents do not use the visible cards for it. ``May be'' means that at least one identity still possible under its hints qualifies. Rule lists follow \texttt{hanabi.py} of the public \texttt{pyhanabi} repository (commit \texttt{556170a}), whose study server maps the names Outer, Intentional and Full to the classes in the second column; we did not verify that the study ran this exact version.}
\label{tab:hd_agents}
\small
\begin{tabular}{p{1.9cm}p{3.5cm}p{9.1cm}}
\toprule
Agent & Source & Rule list in brief \\
\midrule
Outer & \raggedright \texttt{OuterStatePlayer}, after Osawa's outer-state player \citep{osawa2015solving} & Play a card known to be playable; discard a random card known to be useless; hint a playable partner card (highest rank first), naming its color or rank at random among the attributes this agent has not yet named for that card; hint a random partner card with an attribute not yet named; discard a random card. \\
Intentional & \raggedright \texttt{IntentionalPlayer} \citep{eger2017intentional} & Play a card known to be playable; with fewer than 8 information tokens, discard a random card known to be useless; give the hint that best serves its intentions for the partner's cards (play playable cards; discard useless cards, or non-critical ones: rank below 5 with no copy discarded; keep the rest), scoring the touched cards it would move to the intended action and rejecting any hint predicted to cause an unintended play or discard; discard the card with the lowest expected loss given its hints, the discard pile and the fireworks. \\
Full & \raggedright \texttt{SelfIntentional}\-\texttt{Player} \citep{eger2017intentional} & After receiving a hint, read it as the partner's intention: play a touched card that may be playable, else discard a touched card that may be useless; otherwise as Intentional. \\
\bottomrule
\end{tabular}
\end{table}

No agent appears in both corpora. The only shared name is Outer. Our Outer is Walton-Rivers et al.'s rendering of Osawa's outer-state player, and HanabiData's Outer is Eger et al.'s, both after \citet{osawa2015solving}. They are separate implementations, and their behavioral equivalence has not been tested. We therefore make no same-agent comparison between the AI-AI corpus and the human-AI corpus, and we compare partners only within a corpus (in human-AI play: Full, Intentional and Outer).
\FloatBarrier
\section{Hanabi Rules}
\label{app:rules}

Hanabi is a cooperative card game for 2--5 players \citep{bard2020hanabi}. The deck contains 50 cards across 5 colors (Red, Yellow, Green, Blue, Purple), each with ranks 1--5. Rank~1 has 3 copies, ranks 2--4 have 2 copies each, and rank~5 has 1 copy. In a 2-player game each player holds 5 cards, facing outward: every player can see all other hands but not their own.

The team shares 3 life tokens and 8 information tokens. On each turn a player must take exactly one action. (1)~\textit{Play} a card from their hand onto the fireworks, where each color must be built in ascending order 1$\to$5. A play of a card that does not extend its color stack is a \textit{misplay}: the card is discarded and the team loses one life token (a \textit{life loss}). The third life loss ends the game immediately (a \textit{strike-out}). (2)~\textit{Discard} a card to recover one information token. (3)~\textit{Give a hint} (costs one information token): name a color or a rank and point to all cards in one teammate's hand that match. After playing or discarding, the player draws a replacement from the deck. A successfully played rank-5 card also returns one information token. When the deck is exhausted each player takes one final turn. The game also ends when all 25 cards have been played.

\paragraph{Two scoring conventions.} The \emph{strict} score is the standard-rules score: the number of cards successfully played (0--25), and zero after a strike-out. The \emph{partial} score is the fireworks sum, that is, the number of cards successfully played, also after a strike-out. The two agree in every game that does not end in a strike-out. Each AI-AI game records both.

\paragraph{Legal actions enforced in simulation.} In the AI-AI corpus the engine accepts only the following actions and raises an error otherwise. A play of any own card is always legal. A discard of any own card is legal when fewer than 8 information tokens are available. A hint is legal when at least one token is available, the target is the other player, and the named color or rank is present in the target's hand, so that every hint touches at least one card. No action is accepted after the game has ended.

\paragraph{Departures of the HanabiData interface from these rules.} The human-AI games were recorded with the interface of \citet{eger2017intentional}, and we replay its logs as recorded. All 2{,}040 HanabiData games replay without error, and every replayed score equals the recorded score. Two kinds of recorded action fall outside the legal set above. First, 15 hints in 4 games are addressed to the acting player itself (one game each with Full and Outer, two with Intentional); all 15 are actions of the AI seat and none of a human. Second, the interface allows a discard at 8 information tokens, which gains no token. There are 1{,}375 such discards in 460 of the 2{,}040 games (22.5\%): 549 by humans, 403 by Full, 422 by Intentional and 1 by Outer. Every other recorded action is legal: each of the 43{,}222 hints touches at least one card, no hint is given at zero tokens, and no action follows the end of a game. $\hat{p}$ is computed from the replayed state and is exact under both departures. The discards at eight information tokens therefore need no exclusion; Appendix~\ref{app:sensitivity} gives the gap without the four self-hint games.

All analyses in this paper use two-player games without a variant deck. The AI-AI games follow the standard rules, which the engine enforces; the human-AI games follow them up to the two departures just listed. The hanab.live corpus was selected as two-player ``No Variant'' completed games and was not filtered on optional rule settings; Appendix~\ref{app:sensitivity} lists these settings and gives the gap without the games that carry them.
\FloatBarrier
\section{Bot Screen and Human-Human Corpus Definition}
\label{app:bots}

The hanab.live public game corpus contains accounts belonging to community-developed bots. Bot play uses shared H-Group-style conventions (the hanab.live community's published convention system) and would mix programmed convention play into the human-human estimate if included. We define the human-human corpus by (i) a completed-games filter and (ii) a two-tier bot exclusion rule.

\paragraph{Sample.} The 1{,}500 fetched games are the 1{,}500 most recent two-player ``No Variant'' games that the public hanab.live API listed on 21 February 2026. They span fifteen listing pages of 100 games and game identifiers 1{,}757{,}340 to 1{,}766{,}016, and they were played between 13 and 21 February 2026. They were not selected by player or by outcome. Table~\ref{tab:hh_flow} gives the number of games, plays and acting players at each step of the corpus definition.

\paragraph{Unreplayable games.} Of these 1{,}500 games, 25 could not be replayed by our engine and were dropped before any filter. In 17 games the replay stopped at an action on the bottom card of the deck, which hanab.live's deck-plays option (a player may play the last card of the deck directly from the deck) allows. Three games used the one-extra-card option (six-card hands), 1 used the detrimental-characters option (each player receives a handicap rule), and 4 contained an unsupported action type. The engine supports none of these. Every retained game with the deck-plays option enabled (48 of the 425 corpus games, Appendix~\ref{app:sensitivity}) therefore contains no bottom-deck play. One retained game carries the detrimental-characters option and replayed without error; it is among the 13 rule-changing games of Appendix~\ref{app:sensitivity}.

\paragraph{Completed games only.} We keep only games that reached a terminal state (natural end after the deck ran out, perfect score, or strike-out). We drop \emph{truncated} games, whose action log ended without a terminal state. The replay engine classifies each game (Appendix~\ref{app:pipeline}). A game is truncated if its log carries an explicit hanab.live abnormal-end marker (timeout, termination, or idle timeout), or if the log stops with cards still in the deck and no terminal state. The platform ends the final round early once no further points are possible, so the log of a completed natural game can stop one or two turns short of a full final round. Reaching an empty deck is therefore sufficient to classify a game as natural. Of the 1{,}475 replayable games, 424 were perfect, 339 natural, 292 strike-out and 420 truncated. Of the 420 truncated games, 44 were ended before any card was played (1 to 12 actions) and have no play record. The processed play table therefore holds 1{,}431 games, 376 of them truncated. The human-AI corpus and the AI-AI corpus contain no truncated game.

\begin{table}[ht]
\centering
\caption{Definition of the human-human corpus. Plays are play records; acting players are the accounts that made at least one play in the remaining records. Of the 1{,}475 replayable games, 44 truncated games ended before any card was played and have no play record, so the plays of that row come from 1{,}431 games.}
\label{tab:hh_flow}
\small
\begin{tabular}{lrrr}
\toprule
Step & Games & Plays & Acting players \\
\midrule
Most recent games listed by the API & 1{,}500 & -- & -- \\
Replayable by the engine & 1{,}475 & 26{,}229 & 204 \\
Completed games & 1{,}055 & 22{,}029 & 196 \\
Tier 1: games of verified bot families removed & 598 & 13{,}089 & 178 \\
Tier 2: plays of flagged accounts removed & 425 & 9{,}017 & 155 \\
\bottomrule
\end{tabular}
\end{table}

\begin{sloppypar}
\paragraph{Tier 1 (game-level): verified bot families.} We match account names by pattern: \texttt{will-bot$\backslash$d+}, \texttt{jabot$\backslash$d+}, \texttt{clanker$\backslash$d+}, \texttt{Inybot$\backslash$d*}, \texttt{rand-bot$\backslash$d+}, and \texttt{mac-bot.*}. The matching accounts in the corpus are \texttt{will-bot1}, \texttt{will-bot2}, \texttt{will-bot4}, \texttt{jabot1}, \texttt{jabot4}, \texttt{clanker1}, \texttt{clanker2}, \texttt{clanker4}, \texttt{clanker5}, \texttt{clanker6}, \texttt{Inybot}, \texttt{rand-bot1}, and \texttt{mac-bot-test}. The families were identified by cross-referencing the will-hanabi-bot README and each account's public hanab.live history. Every game in which any matching account made a play is removed from the corpus in its entirety. One game (1762152) contains a verified bot account that only gave hints and discarded, so it is retained; it contributes 6 human plays. Excluding every game in which a verified bot account appears at all gives 424 games and 9{,}011 plays and gives a gap of +26.18~pp (human-human corpus: +26.16~pp).
\end{sloppypar}

\paragraph{Tier 2 (account-level): 29 flagged accounts.} We query every remaining account in the corpus through hanab.live's public \texttt{/api/v1/history/\{user\}} endpoint (at most one request per second, responses cached). Per account we compute the total number of games played on hanab.live, the median and minimum time in seconds between the completions of consecutive games, and the fraction of consecutive games completed within 60 or 120~seconds. An account is flagged if it shows one of three evidence patterns:
\begin{itemize}[nosep]
\item \textit{Bot partner}: the account co-plays at least 90\% of its corpus games with a verified bot (11 accounts).
\item \textit{Identical-timing pair}: two accounts have identical inter-game-gap statistics (median, minimum, and the fractions within 60 and 120~seconds) over their retrieved histories, which suggests a coordinated two-agent process seated at the same table (16 accounts in 8 pairs).
\item \textit{Notes bot signature}: in its own per-game notes the account writes the version tag of will-hanabi-bot, ``\texttt{[INFO: v<version>, <framework>]}'', a string produced by the bot software (one account and its exclusive partner, 2 accounts).
\end{itemize}
We drop only these accounts' own play records. In games in which they appear alongside a human, the human's plays are retained. To avoid publicly labeling accounts, some of which may be human, the repository code identifies the 29 Tier-2 accounts only by 16-hex hashes.

Timing-only or volume-only signals without one of these three patterns (for example, an account with more than 5{,}000 total games but a median inter-game gap of several minutes) are \emph{not} sufficient. The tier is permissive toward high-volume enthusiast accounts and rejects only accounts with positive bot evidence.

\paragraph{Resulting corpus.} After the completed-games filter and the two-tier bot exclusion, the corpus retains 425 games and 9{,}017 human play records from 155 acting players (Table~\ref{tab:hh_flow}). As a robustness check, we also computed the human-human gap under a game-level-only bot policy (Tier 1 alone, completed games; 598 games, 13{,}089 plays). The convention gap was +26.2~pp under the two-tier rule and +27.5~pp (95\% game-cluster bootstrap CI [+26.8, +28.3]) under the game-level-only variant. The Tier-2 accounts' own plays in all completed games, including their games with verified bots, had a gap of +29.8~pp (8{,}647 plays). The 4{,}072 of these plays that remain after Tier~1 had +30.5~pp.

\paragraph{Corpus purity: partners of the retained players.} The two-tier rule removes the plays of verified and flagged accounts, not the games in which a retained player sat with such an account. The 425-game corpus is therefore not purely human-human. No acting player in it is a verified or flagged account. In 1 game (6 plays) the partner was a verified bot, and in 10 games (121 plays, 3 acting players) the partner was one of 3 accounts flagged by the Tier-2 screen. The flag is a screening heuristic and not a bot label, so these are plays whose partner was not established as human. Restricting the corpus to games in which neither player was verified or flagged left 414 games, 8{,}890 plays and 153 acting players. Mean $\hat{p}$ was 31.22\% and the failure rate 5.22\%. The gap was $+26.00$~pp (95\% bootstrap CI over acting players [$+24.38$, $+27.65$]), against $+26.16$~pp [$+24.54$, $+27.80$] in the 425-game corpus. The change was $-0.16$~pp (95\% CI of the change [$-0.54$, $+0.03$]). Dropping only the 10 games with a flagged partner gives 415 games and $+25.98$~pp (Appendix~\ref{app:sensitivity}). The 121 plays with a flagged partner had a gap of $+39.4$~pp (mean $\hat{p}$ 46.9\%, failure 7.4\%); with 3 acting players we report no interval and draw no conclusion from them. The 11 removed games (127 plays) are the games in which only one player's rows survive the filter. Appendix~\ref{app:skill} gives their effect on the skill pair types.

The screening code and the exclusion list are in the public repository (see its README). The cached hanab.live history responses contain plaintext usernames and are not shipped; they can be refetched from the public hanab.live API.

\FloatBarrier
\section{Replay Pipeline and Game Engine}
\label{app:pipeline}

We use a pure-Python game engine that exposes each card's hint knowledge at every decision point, which the literal failure probability $\hat{p}$ (\S\ref{sec:posterior}) requires.

\paragraph{Raw data format.}
Each data source provides game records containing two key fields. The two recorded corpora (hanab.live and HanabiData) supply them from their logs, and the AI-AI corpus supplies them from our own generator (Appendix~\ref{app:agents}). The fields are (1)~the \textit{deck order} and (2)~the ordered \textit{action sequence}. The deck order is a complete permutation of the 50-card Hanabi deck (5 colors $\times$ ranks 1--5, with 3/2/2/2/1 copies). Each action specifies a type (play, discard, color hint, or rank hint) and a target (card index or player). The deck order gives the ground-truth card identities. The action sequence determines what each player could have known at each point in the game.

\paragraph{Game state representation.}
The engine maintains two core data structures. The \textit{game state} tracks the global board: players' hands, fireworks, discard pile, and tokens. The \textit{card knowledge} tracks what each player knows about each card in their own hand from accumulated hints. Specifically, the game state records:

\begin{itemize}[nosep]
\item \textbf{Hands}: each player's current cards as (color, rank) tuples, dealt sequentially from the deck.
\item \textbf{Fireworks}: the highest successfully played rank per color (0 if empty); a card $(c, r)$ is playable iff $\text{fireworks}[c] = r - 1$.
\item \textbf{Discard pile}: all discarded and misplayed cards.
\item \textbf{Tokens}: life tokens (initial 3; reaching 0 ends the game) and information tokens (initial 8; spent on hints, recovered on discards and rank-5 plays).
\item \textbf{Per-card knowledge}: for each card in each player's hand, a knowledge state $\kappa$ recording which colors and ranks remain possible, initialized to all 5 colors and ranks 1--5, and progressively narrowed by hints. A player sees the other players' cards and the knowledge states of all cards, but not the identities of their own cards.
\end{itemize}

\paragraph{Action processing.}
\textit{Hint actions}: matching cards receive a positive constraint (possible set narrowed to the hinted value); non-matching cards receive a negative constraint (hinted value removed). Constraints accumulate across all hints \citep{bard2020hanabi}. \textit{Play actions}: if playable ($\text{fireworks}[c] = r - 1$), fireworks advance; otherwise the card is discarded and a life token is lost. A replacement is drawn with fresh empty knowledge. \textit{Discard actions}: the card is discarded, an information token is recovered, and a replacement is drawn. When the deck empties, each player receives one final turn. Recorded logs are replayed as recorded, so the two source-interface departures of HanabiData (Appendix~\ref{app:rules}) are reproduced and not rejected. The AI-AI games are generated and replayed in strict mode, which raises an error on any action outside the legal set of Appendix~\ref{app:rules}. After replaying every recorded action, the engine labels the game with one of four \emph{game-end reasons}: strike-out (three life losses; score 0), perfect (score 25), natural (deck exhausted with any other score) or truncated. A game is truncated when its log ended without a terminal state, either at an explicit hanab.live abnormal-end marker or with cards still in the deck. This label drives the completed-games filter (Appendix~\ref{app:bots}). The AI-AI corpus has 3{,}627 natural and 1{,}273 strike-out games and no perfect game.

\paragraph{Computation of $\hat{p}$.}
At each play action, \emph{before} applying it, we snapshot the state and compute the literal failure probability $\hat{p}$ from the acting player's perspective. $\hat{p}$ is Eq.~\ref{eq:posterior}, evaluated over the full hand and computed exactly by dynamic programming (Appendix~\ref{app:joint}). Its inputs are the knowledge states of all cards in the acting player's hand. The box below gives the computation, and \S\ref{sec:posterior} gives a worked numeric example.

\begin{center}
\fbox{\begin{minipage}{0.93\textwidth}\small
\textbf{Literal failure probability $\hat{p}$ for a play of card $k$} (computed before the play is applied).
\begin{enumerate}[nosep]
\item \textbf{Candidate sets}: for every card $j$ in the acting player's hand, $C_j$ is the set of identities $v = (c, r)$ that every positive and negative hint received on card $j$ leaves possible.
\item \textbf{Unseen copies}: $m_v = \text{total\_copies}(r) - \text{visible\_copies}(c,r)$, with 3/2/2/2/1 copies for ranks 1--5 and visible copies counted across the partner's hand, the discard pile and the fireworks.
\item \textbf{Weighted sum over assignments}: every assignment $x$ of identities to the hand with $x_j \in C_j$ receives the weight $W(x) = \prod_v (m_v)_{n_v(x)}$, the number of deals consistent with it; the sum is evaluated by dynamic programming.
\item \textbf{Share unplayable}: $\hat{p}$ is the $W$-weighted share of assignments in which $x_k$ is unplayable, where $(c, r)$ is playable iff $\text{fireworks}[c] = r - 1$.
\end{enumerate}
\end{minipage}}
\end{center}
Only cards that are completely unconstrained (no positive and no negative hint) are guaranteed to drop out of the sum. Appendix~\ref{app:joint} compares $\hat{p}$ with a per-card approximation that ignores the constraints on the other cards. After computing $\hat{p}$, we apply the action to observe the outcome, producing one record per play that pairs $\hat{p}_i$ with $y_i \in \{0,1\}$.

\paragraph{Validation.}
The pipeline is validated at four levels. (1)~\emph{External ground truth}: every replayed final score of the 2{,}040 HanabiData games equals the recorded score. Of the 1{,}055 completed hanab.live games, 420 have a score recorded in the hanab.live account histories retrieved for the bot screen (Appendix~\ref{app:bots}); the replayed score equals it in 416. The other 4 are All-or-Nothing games, a rule-changing option under which play can continue past the standard end of the game, where our replay stops (Appendix~\ref{app:sensitivity}). (2)~\emph{Logical extremes of $\hat{p}$} (\S\ref{sec:posterior}): across all 100{,}545 analyzed plays, no play with $\hat{p} = 0$ failed (70{,}182 such plays, 51{,}058 of them AI-AI). No play with $\hat{p} = 1$ succeeded (1{,}329 plays, none of them AI-AI). (3)~\emph{Independent implementation}: the dynamic program matched brute-force enumeration on all 936 cross-checked plays, with maximum absolute difference 0 (Appendix~\ref{app:joint}). (4)~\emph{Unit tests} cover the engine, the agents' legality and observation invariance, the computation of $\hat{p}$ and the statistics (details in the package README).

\paragraph{Data and code availability.} All source code, the processed datasets and an executed notebook are provided in the public repository.\footnote{Code and data are available at \url{https://github.com/dockmfgit/hanabi-convention-gap}.} The package README lists the entry point of each analysis. In the processed hanab.live data, usernames are replaced by 16-hex hashes; because the public game id is kept, the hashes are pseudonyms and not an anonymization of the players. The AI-AI games (4{,}900 games, with the generation manifest of Appendix~\ref{app:agents_simulator}) and the HanabiData games (2{,}040 games, in the replay JSON format) are bundled. The raw hanab.live JSON exports are not shipped because they contain plaintext usernames. They can be refetched by game id from the public hanab.live API, and every replay is deterministic from the JSON alone.

\paragraph{Play counts.} The analyzed total is 100{,}545 plays: 9{,}017 human-human, 29{,}049 human-AI by either seat and 62{,}479 AI-AI. The 1{,}475 replayable hanab.live games hold 26{,}229 plays, of which the completed-games filter and the bot screen remove 17{,}212 (Appendix~\ref{app:bots}). Every HanabiData game and every AI-AI game is analyzed.

\paragraph{Compute requirements.}
The replay pipeline and all analyses run on a single CPU, and no model is trained. On a 2024-era laptop the reproduction notebook executes end-to-end (loading the 100{,}545 play records of the three corpora and reproducing the setting-level figures and tables) in tens of minutes. The replay of the three corpora covered 118{,}139 play actions, including 382 plays of the 25 unreplayable hanab.live games, in about 70~s of wall-clock time on one laptop core. This time includes the replay itself; computing $\hat{p}$ took about 270 to 280~$\mu$s per play and the per-card approximation about 4~$\mu$s. The LLM partner data used in Appendix~\ref{app:llm_modern} required GPU memory for local inference of the open-weight Qwen3.6-A3B model (approximately 16~GB) and OpenAI API access for the GPT-5.4 family; this generation step was performed offline. The repository holds only the post-replay convention-gap records ($n=19{,}093$ plays); the notebook reads these records and runs no LLM. Generating the OBL games of \S\ref{sec:obl_results} required a GPU (Appendix~\ref{app:obl}); the package ships their per-game statistics.

\FloatBarrier
\section{Exact Computation of \texorpdfstring{$\hat{p}$}{p-hat}, Verification and the Per-Card Approximation}
\label{app:joint}

\paragraph{Definition.}
The literal failure probability $\hat{p}$ is Eq.~\ref{eq:posterior} of \S\ref{sec:posterior}. We use the notation of the main text. The acting player holds $h$ cards. $C_j$ is the set of identities $v=(c,r)$ that every positive and negative hint received while card $j$ was in the hand leaves possible. $m_v$ is the number of copies of $v$ not visible to the player. For an assignment $x=(x_1,\dots,x_h)$ with $x_j\in C_j$, $n_v(x)$ is the number of cards assigned identity $v$, and the weight is $W(x)=\prod_v (m_v)_{n_v(x)}$ with $(m)_n=m(m-1)\cdots(m-n+1)$. $W(x)=0$ when $x$ uses more copies of an identity than remain unseen. Under the literal reference law $Q_t$, $\hat{p}_t=Q_t(y_t=1)$ is the $W$-weighted share of assignments in which the played card $x_k$ is not playable. For example, if another card in the hand is identified by hints as the last unseen Red~1, every assignment with positive weight gives that copy to that card, and Red~1 leaves the candidate set of the played card.

\paragraph{Why counts and surviving constraints suffice.}
The visible counts $m_v$ and the per-card sets $C_j$ carry the whole literal history. Every card that has left the player's hand was revealed, and every partner card is visible. A past hint therefore constrains the unknown shuffle only through the cards still held, and these constraints travel with the card as $C_j$. The undrawn deck is unconstrained. Hence $D_t$ is the set of arrangements of the $M=\sum_v m_v$ unseen cards over the $h$ hand slots and the deck with $x_j\in C_j$. A hand assignment $x$ has $(M-h)!/\prod_v (m_v-n_v(x))!$ deck completions, which is proportional to $W(x)$. The uniform law $Q_t$ on $D_t$ thus has the hand marginal of Eq.~\ref{eq:posterior}.

\paragraph{Cancellation and the dynamic program.}
A card is guaranteed to cancel from Eq.~\ref{eq:posterior} only if it is completely unconstrained, that is, if $C_j$ contains every identity with $m_v>0$. Such a card contributes the same factor to every term of the numerator and the denominator. A card that was never positively touched but was excluded from a color or rank by a hint to other cards is constrained, and it is not guaranteed to cancel. It can still cancel in a particular state, for example when its candidate set shares no identity with that of the played card. Negative information is common. On 98.8\% of the 100{,}545 analyzed plays (Appendix~\ref{app:pipeline}) at least one other card in the hand carried a hint constraint. On the remaining 1.2\% (human-human 0.9\%, human-AI 0.2\%, AI-AI 1.7\%), Eq.~\ref{eq:posterior} and Eq.~\ref{eq:percard} agree exactly. The sum is evaluated exactly by a dynamic program over the constrained cards with integer weights; the only floating-point operation is the final division. A state with no feasible assignment cannot arise from a legal transcript. The code raises an error on such a state instead of returning a number, and no play in any corpus triggered it.

\paragraph{Verification.}
The dynamic program was compared with brute-force enumeration of all assignments $x\in\prod_j C_j$. The comparison covered 936 plays on which another card in the hand was constrained. They are 591 plays from a systematic sample of the 118{,}139 replayed play actions of the three corpora (Appendix~\ref{app:pipeline}), all 249 such plays in the first 14 replayable human-human games (in replay order), and 96 plays from the LLM corpus of Appendix~\ref{app:llm_modern}. The systematic sample took every 197th replayed play action, including plays later removed by the corpus definition. It kept a play when another card in the hand was constrained and the enumeration had at most two million assignments. The maximum absolute difference was $0$ on all 936 plays. Unit tests check the exact reduction to Eq.~\ref{eq:percard} when no other card is constrained, hand-computed cross-card cases in both directions, and that inconsistent states raise an error.

\begin{table}[!ht]
\centering
\caption{Play-weighted (pooled) convention gap under $\hat{p}$ (Eq.~\ref{eq:posterior}) and under the per-card approximation (Eq.~\ref{eq:percard}). Human-AI rows use human plays only. Difference = per-card approximation minus $\hat{p}$; the AI-AI row is given to two decimals because its values are small. Intervals for the gaps under $\hat{p}$ are in main-text Table~\ref{tab:headline} and Appendix~\ref{app:stats}.}
\label{tab:joint}
\small
\begin{tabular}{lrccc}
\toprule
Setting & $N$ plays & Gap ($\hat{p}$) & Gap (per-card approx.) & Difference \\
\midrule
Human-human & 9{,}017 & +26.2~pp & +28.2~pp & +2.0 \\
Human-AI (human plays) & 15{,}472 & +16.4~pp & +16.9~pp & +0.5 \\
AI-AI & 62{,}479 & $-$0.64~pp & $-$0.47~pp & +0.17 \\
\midrule
Human-AI, partner Full & 4{,}769 & +24.1~pp & +24.6~pp & +0.5 \\
Human-AI, partner Intentional & 5{,}122 & +20.2~pp & +20.8~pp & +0.6 \\
Human-AI, partner Outer & 5{,}581 & +6.2~pp & +6.7~pp & +0.5 \\
\bottomrule
\end{tabular}
\end{table}

\paragraph{The per-card approximation.}
The per-card approximation ignores the constraints on the other $h-1$ cards:
\begin{equation}
\hat{p}^{\,\mathrm{card}}_t = \frac{\sum_{v \in C_k,\, v \notin \text{Playable}} m_v}{\sum_{v \in C_k} m_v}.
\label{eq:percard}
\end{equation}
It equals Eq.~\ref{eq:posterior} when every other card in the hand is completely unconstrained. In the worked example of \S\ref{sec:posterior} the other cards carry no constraints, so both formulas give $\hat{p} = 5/8$. Table~\ref{tab:joint} compares the gap under Eq.~\ref{eq:posterior} with the gap under the per-card approximation of Eq.~\ref{eq:percard}. Under the per-card approximation the human-human gap would be +28.2~pp, 2.0~pp above its value under $\hat{p}$, and every other row differs by at most 0.6~pp. The sign of every row was the same under both. Per-play differences (per-card minus joint) ran in both directions in the human corpora. In human-human play 17.3\% of plays differed (joint value higher on 196, lower on 1{,}367; mean absolute difference 2.1~pp over all plays and 12.3~pp over the plays that differed). In human-AI play 11.0\% differed (538 higher, 1{,}170 lower). In the AI-AI corpus 4.0\% differed (2{,}511 plays), and the joint value was lower on every one. Only Eq.~\ref{eq:posterior} is the reference law of Proposition~\ref{prop:zero}. The joint value also had a slightly lower Brier score and a higher AUC (in-sample; no interval computed). The Brier score was 0.175 against 0.190 for the per-card value in human-human play and 0.159 against 0.162 in human-AI play. AUC was 0.815 against 0.811 and 0.871 against 0.868. In the AI-AI corpus the two were indistinguishable at the printed precision (Brier 0.0335 for both; AUC 0.9784 against 0.9782).

The pattern across hint strata (Table~\ref{tab:hintcount}, Appendix~\ref{app:convgap_by_subject}) was the same under both formulas. The human-human gaps on zero-touch, one-touch and two-or-more-touch (2+) plays were +46.2, +35.6 and +1.6~pp under Eq.~\ref{eq:posterior} and +47.8, +38.5 and +1.7~pp under Eq.~\ref{eq:percard}. The per-card value is about 70 times faster to compute (Appendix~\ref{app:pipeline}).
\FloatBarrier
\section{One-Step Card-Policy Contrasts and the Card-Choice Model: Full Tables}
\label{app:receiver}

\begin{figure}[!ht]
\centering
\includegraphics[width=\textwidth]{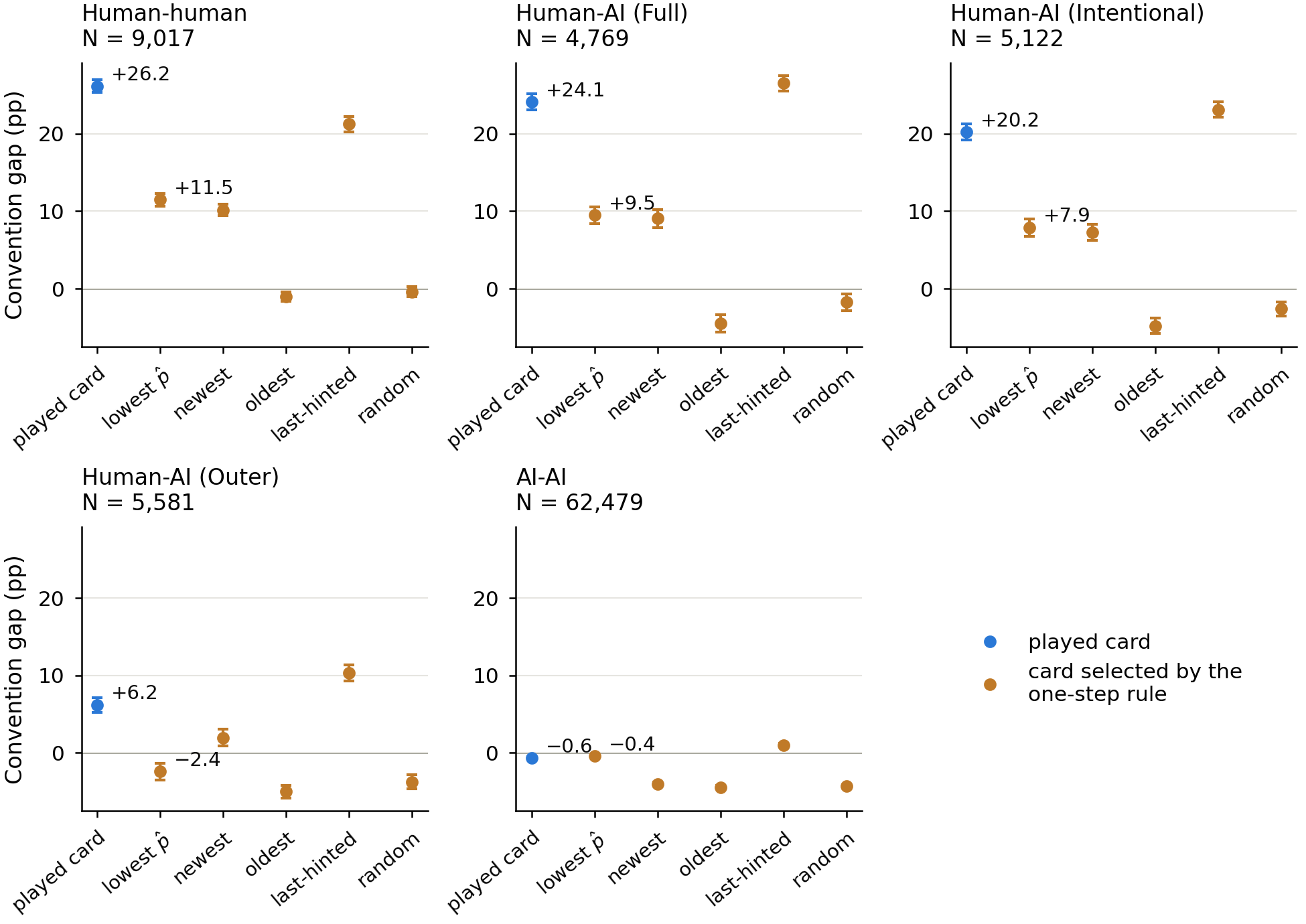}
\caption{One-step card-policy contrasts on observed play turns. Each panel shows the gap (mean $\hat{p}$ minus failure rate, pp, play-weighted) of the card actually played (blue; ``played card''). Orange points (``card selected by the one-step rule'') give the gap of the card that each card rule would have selected from the same hand on the same turns. The rule labeled ``lowest $\hat{p}$'' and the last-hinted rule break ties by the newest card. Error bars are 95\% game-cluster bootstrap intervals (10{,}000 draws). In the AI-AI panel ($N = 62{,}479$ play turns) the gap is $-$0.6~pp for the played card and $-$0.4~pp for the lowest-$\hat{p}$ card. The orange points are one-step substitutions, not deployments of the rule (see text). A gap is not a failure rate: mean $\hat{p}$ and failure for every point are in Table~\ref{tab:receiver_policies}.}
\label{fig:receiver}
\end{figure}

\paragraph{Method: one-step substitutions.} The unit is a play turn, a turn on which the acting player actually played a card. On every play turn $\hat{p}$ (\S\ref{sec:posterior}) was computed for every card in the acting player's hand in the game state before the play. Each card's true playability was read from the deck. A card rule (one-step card policy) selects one card of that hand per turn. For a rule we report the mean $\hat{p}$ of the selected cards, the fraction of them that would have failed, and the gap, which is the first minus the second. All three are pooled over play turns with weight one per turn. One card is played per turn, so this is the play weighting of the main text, and we call it play-weighted. The six rules are:
\begin{itemize}[nosep]
\item the card played;
\item the lowest-$\hat{p}$ card, with ties broken by the newest card;
\item the newest card;
\item the oldest card;
\item the most recently hinted card (``last-hinted''): when the most recent hint touched several cards still in hand, the newest of them, and the newest card if no card in hand has been hinted;
\item a uniformly random card.
\end{itemize}
Figure~\ref{fig:receiver} shows the gaps and Table~\ref{tab:receiver_policies} lists every value. Tables~\ref{tab:receiver_contrast}, \ref{tab:receiver_lasthinted} and~\ref{tab:receiver_lasthinted_ties} compare the played card with the lowest-$\hat{p}$ card and with the last-hinted card, including the sensitivity to the tie rule.

\paragraph{What a substitution shows.} A one-step substitution replaces only the card on a turn the acting player reached and chose to play on. It inherits the acting player's earlier actions, the partner's responses to them and the decision to play. It is not a deployment of the rule, and a contrast between two rules is not a causal attribution of the gap to either player. For two rules $a$ and $b$ the identity $G_a - G_b = (\bar{p}_a - \bar{p}_b) - (F_a - F_b)$ holds, where $\bar{p}$ is mean $\hat{p}$ and $F$ the failure rate; we report both terms.

\paragraph{Intervals and weighting.} Intervals are 95\% percentile cluster bootstraps (10{,}000 draws) in which whole clusters are resampled and all rules share the same draws, so contrasts are paired. The cluster is the game; for human-AI settings the participant-cluster interval (228 participants, resampled jointly across partners) is given beside it. Human-AI play turns are concentrated in few participants (Appendix~\ref{app:sensitivity}). The corpora are 9{,}017 human-human play turns (425 games), 15{,}472 human play turns in human-AI games (2{,}035 games) and 62{,}479 play turns in the AI-AI games (4{,}900 games; Appendix~\ref{app:agents}). For human-human play the intervals in this appendix are game-cluster intervals. Person-level clustering widens them (acting player, dyad and component; Appendix~\ref{app:stats}). For the gap of the played card the acting-player interval is [+24.5, +27.8], against the game-cluster interval [+25.3, +27.0]. For the contrast of the played card with the lowest-$\hat{p}$ card it is [+13.0, +16.5], against [+13.7, +15.7].

\begin{table}[!ht]
\centering
\caption{Card rules on observed play turns: mean $\hat{p}$ (\%), failure rate (\%), gap (pp) with 95\% game-cluster and, for human-AI, participant-cluster bootstrap intervals, and the percentage of turns on which the rule selects the card actually played. All values are play-weighted. The lowest-$\hat{p}$ rule and the last-hinted rule (the most recently hinted card) break ties by the newest card. The played-card rows repeat the intervals of Tables~\ref{tab:clusterci} and~\ref{tab:convgap_agent}. The participant-cluster interval of the Full random-card gap has upper bound +0.002~pp.}
\label{tab:receiver_policies}
\footnotesize
\setlength{\tabcolsep}{4pt}
\begin{tabular}{llrrrrllr}
\toprule
Setting & Rule & $N$ & Mean $\hat{p}$ & Failure & Gap & Game CI & Participant CI & Coincides \\
\midrule
Human-human & played & 9,017 & 31.4 & 5.3 & +26.16 & [+25.32, +27.01] & -- & 100.0 \\
Human-human & lowest-$\hat{p}$ & 9,017 & 27.1 & 15.6 & +11.50 & [+10.64, +12.34] & -- & 73.7 \\
Human-human & newest & 9,017 & 66.1 & 55.9 & +10.15 & [+9.40, +10.90] & -- & 33.7 \\
Human-human & oldest & 9,017 & 71.6 & 72.7 & $-$1.02 & [$-$1.67, $-$0.38] & -- & 20.2 \\
Human-human & last-hinted & 9,017 & 47.9 & 26.7 & +21.26 & [+20.28, +22.24] & -- & 67.1 \\
Human-human & random & 9,017 & 71.3 & 71.7 & $-$0.40 & [$-$1.06, +0.25] & -- & 19.9 \\
\addlinespace
Human-AI (all partners) & played & 15,472 & 39.1 & 22.7 & +16.38 & [+15.72, +17.06] & [+15.04, +17.76] & 100.0 \\
Human-AI (all partners) & lowest-$\hat{p}$ & 15,472 & 32.2 & 27.5 & +4.68 & [+4.02, +5.34] & [+3.37, +6.00] & 61.8 \\
Human-AI (all partners) & newest & 15,472 & 65.0 & 59.1 & +5.90 & [+5.25, +6.54] & [+5.21, +6.67] & 28.9 \\
Human-AI (all partners) & oldest & 15,472 & 72.5 & 77.2 & $-$4.78 & [$-$5.35, $-$4.21] & [$-$5.58, $-$3.75] & 17.1 \\
Human-AI (all partners) & last-hinted & 15,472 & 45.8 & 26.2 & +19.56 & [+18.90, +20.22] & [+18.64, +20.65] & 72.0 \\
Human-AI (all partners) & random & 15,472 & 70.0 & 72.5 & $-$2.48 & [$-$3.03, $-$1.94] & [$-$3.31, $-$1.59] & 20.1 \\
\addlinespace
Human-AI: Full & played & 4,769 & 38.5 & 14.4 & +24.14 & [+23.12, +25.18] & [+22.92, +24.88] & 100.0 \\
Human-AI: Full & lowest-$\hat{p}$ & 4,769 & 32.0 & 22.5 & +9.49 & [+8.41, +10.58] & [+8.21, +10.83] & 62.3 \\
Human-AI: Full & newest & 4,769 & 63.7 & 54.6 & +9.05 & [+7.91, +10.15] & [+7.56, +10.40] & 31.8 \\
Human-AI: Full & oldest & 4,769 & 69.2 & 73.7 & $-$4.49 & [$-$5.59, $-$3.37] & [$-$5.70, $-$2.92] & 17.5 \\
Human-AI: Full & last-hinted & 4,769 & 43.7 & 17.2 & +26.52 & [+25.54, +27.51] & [+25.21, +27.15] & 76.3 \\
Human-AI: Full & random & 4,769 & 68.0 & 69.7 & $-$1.74 & [$-$2.82, $-$0.64] & [$-$2.89, +0.00] & 20.7 \\
\addlinespace
Human-AI: Intentional & played & 5,122 & 37.9 & 17.6 & +20.25 & [+19.21, +21.27] & [+18.01, +22.45] & 100.0 \\
Human-AI: Intentional & lowest-$\hat{p}$ & 5,122 & 30.9 & 23.0 & +7.92 & [+6.79, +9.03] & [+6.09, +9.79] & 62.5 \\
Human-AI: Intentional & newest & 5,122 & 63.6 & 56.3 & +7.24 & [+6.18, +8.28] & [+6.00, +8.44] & 31.9 \\
Human-AI: Intentional & oldest & 5,122 & 70.8 & 75.6 & $-$4.80 & [$-$5.78, $-$3.80] & [$-$6.12, $-$3.39] & 16.8 \\
Human-AI: Intentional & last-hinted & 5,122 & 43.0 & 19.8 & +23.14 & [+22.14, +24.11] & [+21.70, +24.88] & 74.6 \\
Human-AI: Intentional & random & 5,122 & 68.1 & 70.7 & $-$2.62 & [$-$3.53, $-$1.73] & [$-$3.55, $-$1.14] & 19.9 \\
\addlinespace
Human-AI: Outer & played & 5,581 & 40.6 & 34.4 & +6.20 & [+5.19, +7.19] & [+4.85, +7.52] & 100.0 \\
Human-AI: Outer & lowest-$\hat{p}$ & 5,581 & 33.4 & 35.8 & $-$2.39 & [$-$3.50, $-$1.34] & [$-$3.72, $-$0.94] & 60.6 \\
Human-AI: Outer & newest & 5,581 & 67.5 & 65.5 & +1.98 & [+0.88, +3.03] & [+0.79, +3.17] & 23.6 \\
Human-AI: Outer & oldest & 5,581 & 76.8 & 81.8 & $-$5.02 & [$-$5.87, $-$4.20] & [$-$5.54, $-$3.95] & 17.1 \\
Human-AI: Outer & last-hinted & 5,581 & 50.1 & 39.8 & +10.33 & [+9.24, +11.39] & [+9.12, +11.52] & 66.0 \\
Human-AI: Outer & random & 5,581 & 72.3 & 76.0 & $-$3.73 & [$-$4.66, $-$2.82] & [$-$4.58, $-$2.63] & 19.9 \\
\addlinespace
AI-AI & played & 62,479 & 8.0 & 8.6 & $-$0.64 & [$-$0.79, $-$0.49] & -- & 100.0 \\
AI-AI & lowest-$\hat{p}$ & 62,479 & 8.0 & 8.4 & $-$0.42 & [$-$0.57, $-$0.27] & -- & 91.9 \\
AI-AI & newest & 62,479 & 66.3 & 70.3 & $-$4.01 & [$-$4.28, $-$3.73] & -- & 15.3 \\
AI-AI & oldest & 62,479 & 72.3 & 76.7 & $-$4.43 & [$-$4.65, $-$4.22] & -- & 17.4 \\
AI-AI & last-hinted & 62,479 & 28.0 & 27.0 & +1.02 & [+0.82, +1.22] & -- & 63.9 \\
AI-AI & random & 62,479 & 65.5 & 69.7 & $-$4.29 & [$-$4.52, $-$4.07] & -- & 20.3 \\
\bottomrule
\end{tabular}
\end{table}

\paragraph{Played card versus lowest-$\hat{p}$ card: both terms.} Table~\ref{tab:receiver_contrast} gives the contrast between the played card and the lowest-$\hat{p}$ card under three tie rules. The mean-$\hat{p}$ term is non-negative by construction and does not depend on the tie rule. In human-human play the played cards had 4.38~pp higher mean $\hat{p}$ and failed 10.29~pp less often than the lowest-$\hat{p}$ cards, so the gap contrast was +14.67~pp. With Full the two terms were +6.53 and $-$8.11~pp, and with Intentional +6.96 and $-$5.37~pp. With Outer the contrast of +8.59~pp was mostly the mean-$\hat{p}$ term (+7.20~pp); the failure term was $-$1.40~pp, and its participant-cluster interval [$-$3.14, +0.05] included zero. In the AI-AI games the played card was the lowest-$\hat{p}$ card on 91.9\% of turns and was in the set of lowest-$\hat{p}$ cards on 99.8\% of turns. The AI-AI contrast was $-$0.22~pp [$-$0.28, $-$0.15] under newest-card ties and changed sign across tie rules ($-$0.22, +0.08 and +0.29~pp), so it is not distinguishable from zero in a rule-independent way. Table~\ref{tab:receiver_hint} splits the two gaps by the number of hints that positively touched the played card. In human-human play the difference came from one-touch plays (+35.57 against +17.18~pp) and zero-touch plays (+46.22 against +1.96~pp).

\begin{table}[!ht]
\centering
\caption{Contrast between the card played and the lowest-$\hat{p}$ card on the same play turns, under three tie rules for the lowest-$\hat{p}$ card (pp, play-weighted). $\Delta$ gap $=$ $\Delta$ mean $\hat{p}$ $-$ $\Delta$ failure. Brackets are 95\% game-cluster bootstrap intervals; the last column is the participant-cluster interval of $\Delta$ gap (human-AI only). For Outer under newest-card ties the participant-cluster interval of $\Delta$ failure is [$-$3.14, +0.05].}
\label{tab:receiver_contrast}
\footnotesize
\setlength{\tabcolsep}{3pt}
\resizebox{\textwidth}{!}{\begin{tabular}{llrllll}
\toprule
Setting & Tie rule & Gap of lowest-$\hat{p}$ & $\Delta$ mean $\hat{p}$ & $\Delta$ failure & $\Delta$ gap & Participant CI \\
\midrule
Human-human & newest & +11.50 & +4.38 [+4.07, +4.69] & $-$10.29 [$-$11.19, $-$9.40] & +14.67 [+13.71, +15.65] & -- \\
Human-human & random & +9.14 & +4.38 [+4.07, +4.69] & $-$12.64 [$-$13.62, $-$11.69] & +17.02 [+15.99, +18.06] & -- \\
Human-human & oldest & +7.10 & +4.38 [+4.07, +4.69] & $-$14.68 [$-$15.73, $-$13.64] & +19.06 [+17.97, +20.16] & -- \\
\addlinespace
Human-AI (all partners) & newest & +4.68 & +6.91 [+6.62, +7.22] & $-$4.78 [$-$5.49, $-$4.10] & +11.70 [+11.00, +12.39] & [+10.74, +12.37] \\
Human-AI (all partners) & random & +3.63 & +6.91 [+6.62, +7.22] & $-$5.84 [$-$6.55, $-$5.13] & +12.75 [+12.05, +13.46] & [+11.91, +13.48] \\
Human-AI (all partners) & oldest & +3.08 & +6.91 [+6.62, +7.22] & $-$6.39 [$-$7.13, $-$5.66] & +13.30 [+12.58, +14.03] & [+12.36, +14.06] \\
\addlinespace
Human-AI: Full & newest & +9.49 & +6.53 [+6.04, +7.03] & $-$8.11 [$-$9.35, $-$6.88] & +14.64 [+13.39, +15.88] & [+12.91, +15.85] \\
Human-AI: Full & random & +8.68 & +6.53 [+6.04, +7.03] & $-$8.93 [$-$10.19, $-$7.63] & +15.46 [+14.19, +16.72] & [+13.78, +16.70] \\
Human-AI: Full & oldest & +7.88 & +6.53 [+6.04, +7.03] & $-$9.73 [$-$11.05, $-$8.39] & +16.26 [+14.94, +17.57] & [+14.08, +17.59] \\
\addlinespace
Human-AI: Intentional & newest & +7.92 & +6.96 [+6.44, +7.51] & $-$5.37 [$-$6.64, $-$4.16] & +12.33 [+11.13, +13.60] & [+11.16, +13.50] \\
Human-AI: Intentional & random & +6.94 & +6.96 [+6.44, +7.51] & $-$6.35 [$-$7.61, $-$5.14] & +13.31 [+12.08, +14.58] & [+12.40, +14.71] \\
Human-AI: Intentional & oldest & +6.51 & +6.96 [+6.44, +7.51] & $-$6.77 [$-$8.06, $-$5.49] & +13.74 [+12.47, +15.05] & [+12.62, +15.19] \\
\addlinespace
Human-AI: Outer & newest & $-$2.39 & +7.20 [+6.70, +7.73] & $-$1.40 [$-$2.50, $-$0.31] & +8.59 [+7.53, +9.68] & [+7.16, +9.54] \\
Human-AI: Outer & random & $-$3.38 & +7.20 [+6.70, +7.73] & $-$2.38 [$-$3.52, $-$1.28] & +9.58 [+8.48, +10.69] & [+8.38, +10.79] \\
Human-AI: Outer & oldest & $-$4.17 & +7.20 [+6.70, +7.73] & $-$3.17 [$-$4.32, $-$2.07] & +10.37 [+9.26, +11.50] & [+9.07, +11.59] \\
\addlinespace
AI-AI & newest & $-$0.42 & +0.01 [+0.00, +0.01] & +0.22 [+0.16, +0.29] & $-$0.22 [$-$0.28, $-$0.15] & -- \\
AI-AI & random & $-$0.72 & +0.01 [+0.00, +0.01] & $-$0.08 [$-$0.18, +0.02] & +0.08 [$-$0.01, +0.18] & -- \\
AI-AI & oldest & $-$0.93 & +0.01 [+0.00, +0.01] & $-$0.28 [$-$0.40, $-$0.16] & +0.29 [+0.17, +0.41] & -- \\
\bottomrule
\end{tabular}}
\end{table}

\paragraph{Tie-break sensitivity.} At least two cards shared the lowest $\hat{p}$ on 22.4\% of human-human turns, 22.9--28.1\% of human-AI turns and 16.3\% of AI-AI turns. A card with $\hat{p} = 0$ was in hand on 45.6\% of human-human turns, 38.3\% of human-AI turns (all partners) and 81.7\% of AI-AI turns; the AI-AI agents played such a card on all but 8 of those AI-AI turns. Breaking ties by the newest card, at random or by the oldest card changes only the failure term. The gap of the lowest-$\hat{p}$ card was +11.50 / +9.14 / +7.10~pp in human-human play and $-$0.42 / $-$0.72 / $-$0.93~pp in AI-AI play. With the AI partners it was +9.49 / +8.68 / +7.88 (Full), +7.92 / +6.94 / +6.51 (Intentional) and $-$2.39 / $-$3.38 / $-$4.17~pp (Outer). The human-human contrast therefore lay between +14.67 and +19.06~pp depending on the tie rule. No single value is identified without choosing a tie rule.

\paragraph{Played card versus most recently hinted card.} Table~\ref{tab:receiver_lasthinted} gives the same two terms for the last-hinted rule with newest-card tie-breaking. Under that tie rule the last-hinted card in human-AI play had a larger gap than the played card and also failed more often (Full 17.2\% against 14.4\%; Intentional 19.8\% against 17.6\%; Outer 39.8\% against 34.4\%). In human-human play it had a smaller gap (+21.26~pp) and failed more often (26.7\% against 5.3\%). The gap therefore does not order rules by immediate outcome.

\paragraph{Tie-break sensitivity of the last-hinted rule.} A hint usually touches several cards, so the most recently hinted card is often not unique. The most recent hint had touched at least two cards still in hand on 41.6\% of human-human turns and 50.1\% of AI-AI turns. In human-AI play the share was 39.1\% (Full 33.5\%, Intentional 33.4\%, Outer 49.0\%). Table~\ref{tab:receiver_lasthinted_ties} repeats the rule with the tie broken by the newest card, at random or by the oldest card. The failure ordering did not depend on the tie rule: the last-hinted card failed more often than the played card in every setting under every tie rule. The gap ordering did depend on it. With every AI partner the gap of the last-hinted card was larger than that of the played card under newest-card ties and smaller under oldest-card ties. Under oldest-card ties it was +22.63 against +24.14~pp (Full), +19.60 against +20.25~pp (Intentional) and +5.59 against +6.20~pp (Outer). Under oldest-card ties the participant-cluster interval of the difference included zero for Intentional and Outer. With random ties the last-hinted gap was still the larger with every AI partner, but for Full the participant-cluster interval of the difference included zero ($-$0.43~pp [$-$1.07, +0.25]). With the AI partners, which of the two cards has the larger gap therefore depends on the tie rule. In human-human play the played card had the larger gap under every tie rule ($\Delta$ gap +4.90 to +14.16~pp), and in AI-AI play the last-hinted card had the larger gap under every tie rule ($-$1.66 to $-$1.02~pp).

\begin{table}[!ht]
\centering
\caption{Contrast between the card played and the most recently hinted card on the same play turns (played minus last-hinted, newest-card ties; pp, play-weighted). Brackets are 95\% game-cluster bootstrap intervals; the last column is the participant-cluster interval of $\Delta$ gap.}
\label{tab:receiver_lasthinted}
\footnotesize
\begin{tabular}{lllll}
\toprule
Setting & $\Delta$ mean $\hat{p}$ & $\Delta$ failure & $\Delta$ gap & Participant CI \\
\midrule
Human-human & $-$16.48 [$-$17.31, $-$15.62] & $-$21.38 [$-$22.79, $-$20.03] & +4.90 [+3.94, +5.91] & -- \\
Human-AI (all partners) & $-$6.70 [$-$7.17, $-$6.23] & $-$3.52 [$-$4.09, $-$2.95] & $-$3.18 [$-$3.55, $-$2.82] & [$-$3.85, $-$2.56] \\
Human-AI: Full & $-$5.17 [$-$5.93, $-$4.41] & $-$2.79 [$-$3.66, $-$1.90] & $-$2.39 [$-$2.88, $-$1.87] & [$-$2.84, $-$1.79] \\
Human-AI: Intentional & $-$5.10 [$-$5.90, $-$4.30] & $-$2.21 [$-$3.11, $-$1.29] & $-$2.89 [$-$3.42, $-$2.38] & [$-$3.85, $-$1.99] \\
Human-AI: Outer & $-$9.47 [$-$10.29, $-$8.64] & $-$5.34 [$-$6.43, $-$4.24] & $-$4.13 [$-$4.88, $-$3.36] & [$-$5.17, $-$3.22] \\
AI-AI & $-$20.06 [$-$20.36, $-$19.76] & $-$18.40 [$-$18.76, $-$18.04] & $-$1.66 [$-$1.86, $-$1.47] & -- \\
\bottomrule
\end{tabular}
\end{table}

\begin{table}[!ht]
\centering
\caption{The last-hinted rule under three tie rules for turns on which the most recent hint touched several cards still in hand (``Tied'': percentage of play turns). Mean $\hat{p}$ and failure in \%, gap in pp, play-weighted. $\Delta$ gap is played minus last-hinted with its 95\% game-cluster bootstrap interval; the last column (``Person CI'') is the person-cluster interval of $\Delta$ gap, by participant (human-AI) or acting player (human-human) (10{,}000 draws). The played and newest-card rows are those of Tables~\ref{tab:receiver_policies} and~\ref{tab:receiver_lasthinted}. The random tie-break is a single draw.}
\label{tab:receiver_lasthinted_ties}
\footnotesize
\setlength{\tabcolsep}{4pt}
\begin{tabular}{llrrrrll}
\toprule
Setting & Card / tie rule & Tied & Mean $\hat{p}$ & Failure & Gap & $\Delta$ gap [game CI] & Person CI \\
\midrule
Human-human & played & 41.6 & 31.4 & 5.3 & +26.16 & -- & -- \\
Human-human & last-hinted, newest & & 47.9 & 26.7 & +21.26 & +4.90 [+3.94, +5.91] & [+3.33, +6.92] \\
Human-human & last-hinted, random & & 50.2 & 34.1 & +16.08 & +10.09 [+9.06, +11.17] & [+8.25, +12.22] \\
Human-human & last-hinted, oldest & & 51.8 & 39.8 & +12.00 & +14.16 [+13.09, +15.26] & [+12.04, +16.51] \\
\addlinespace
Human-AI (all partners) & played & 39.1 & 39.1 & 22.7 & +16.38 & -- & -- \\
Human-AI (all partners) & last-hinted, newest & & 45.8 & 26.2 & +19.56 & $-$3.18 [$-$3.55, $-$2.82] & [$-$3.85, $-$2.56] \\
Human-AI (all partners) & last-hinted, random & & 47.0 & 29.6 & +17.46 & $-$1.08 [$-$1.47, $-$0.69] & [$-$1.83, $-$0.31] \\
Human-AI (all partners) & last-hinted, oldest & & 48.3 & 32.9 & +15.48 & +0.90 [+0.50, +1.31] & [+0.32, +1.65] \\
\addlinespace
Human-AI: Full & played & 33.5 & 38.5 & 14.4 & +24.14 & -- & -- \\
Human-AI: Full & last-hinted, newest & & 43.7 & 17.2 & +26.52 & $-$2.39 [$-$2.88, $-$1.87] & [$-$2.84, $-$1.79] \\
Human-AI: Full & last-hinted, random & & 45.1 & 20.5 & +24.57 & $-$0.43 [$-$0.97, +0.11] & [$-$1.07, +0.25] \\
Human-AI: Full & last-hinted, oldest & & 46.7 & 24.1 & +22.63 & +1.50 [+0.98, +2.04] & [+1.10, +2.22] \\
\addlinespace
Human-AI: Intentional & played & 33.4 & 37.9 & 17.6 & +20.25 & -- & -- \\
Human-AI: Intentional & last-hinted, newest & & 43.0 & 19.8 & +23.14 & $-$2.89 [$-$3.42, $-$2.38] & [$-$3.85, $-$1.99] \\
Human-AI: Intentional & last-hinted, random & & 45.1 & 23.6 & +21.47 & $-$1.22 [$-$1.76, $-$0.68] & [$-$2.23, $-$0.26] \\
Human-AI: Intentional & last-hinted, oldest & & 46.8 & 27.2 & +19.60 & +0.65 [+0.06, +1.24] & [$-$0.28, +1.55] \\
\addlinespace
Human-AI: Outer & played & 49.0 & 40.6 & 34.4 & +6.20 & -- & -- \\
Human-AI: Outer & last-hinted, newest & & 50.1 & 39.8 & +10.33 & $-$4.13 [$-$4.88, $-$3.36] & [$-$5.17, $-$3.22] \\
Human-AI: Outer & last-hinted, random & & 50.5 & 42.8 & +7.70 & $-$1.50 [$-$2.34, $-$0.65] & [$-$2.34, $-$0.22] \\
Human-AI: Outer & last-hinted, oldest & & 51.2 & 45.6 & +5.59 & +0.61 [$-$0.26, +1.49] & [$-$0.12, +1.93] \\
\addlinespace
AI-AI & played & 50.1 & 8.0 & 8.6 & $-$0.64 & -- & -- \\
AI-AI & last-hinted, newest & & 28.0 & 27.0 & +1.02 & $-$1.66 [$-$1.86, $-$1.47] & -- \\
AI-AI & last-hinted, random & & 29.1 & 28.4 & +0.75 & $-$1.39 [$-$1.56, $-$1.21] & -- \\
AI-AI & last-hinted, oldest & & 31.4 & 31.0 & +0.38 & $-$1.02 [$-$1.18, $-$0.85] & -- \\
\bottomrule
\end{tabular}
\end{table}

\begin{table}[!ht]
\centering
\caption{Played card versus lowest-$\hat{p}$ card (newest-card ties), by the number of hints that positively touched the card actually played. Mean $\hat{p}$ (\%) and failure (\%) are those of the played card; gaps are in pp, play-weighted. A stratum's gap is bounded by its mean $\hat{p}$ (Appendix~\ref{app:robustness}). The strata are defined by the played card, so the lowest-$\hat{p}$ column refers to the same turns, not to cards with the same number of touches. No intervals were computed for the lowest-$\hat{p}$ column.}
\label{tab:receiver_hint}
\footnotesize
\begin{tabular}{llrrrrr}
\toprule
Setting & Positive touches & $N$ & Mean $\hat{p}$ & Failure & Gap, played & Gap, lowest-$\hat{p}$ \\
\midrule
Human-human & 0 & 484 & 68.5 & 22.3 & +46.22 & +1.96 \\
Human-human & 1 & 5,884 & 41.1 & 5.5 & +35.57 & +17.18 \\
Human-human & 2+ & 2,649 & 3.3 & 1.7 & +1.62 & +0.62 \\
\addlinespace
Human-AI (all partners) & 0 & 778 & 72.2 & 84.3 & $-$12.15 & $-$11.33 \\
Human-AI (all partners) & 1 & 10,500 & 48.9 & 24.2 & +24.67 & +7.81 \\
Human-AI (all partners) & 2+ & 4,194 & 8.3 & 7.4 & +0.91 & $-$0.16 \\
\addlinespace
Human-AI: Full & 0 & 220 & 72.2 & 77.3 & $-$5.04 & $-$10.47 \\
Human-AI: Full & 1 & 3,542 & 46.2 & 13.9 & +32.27 & +12.94 \\
Human-AI: Full & 2+ & 1,007 & 4.2 & 2.3 & +1.91 & +1.73 \\
\addlinespace
Human-AI: Intentional & 0 & 390 & 73.0 & 87.4 & $-$14.48 & $-$10.71 \\
Human-AI: Intentional & 1 & 3,552 & 45.2 & 14.8 & +30.42 & +12.32 \\
Human-AI: Intentional & 2+ & 1,180 & 4.2 & 3.1 & +1.10 & +0.83 \\
\addlinespace
Human-AI: Outer & 0 & 168 & 70.2 & 86.3 & $-$16.08 & $-$13.93 \\
Human-AI: Outer & 1 & 3,406 & 55.5 & 44.7 & +10.78 & $-$2.24 \\
Human-AI: Outer & 2+ & 2,007 & 12.8 & 12.6 & +0.29 & $-$1.69 \\
\addlinespace
AI-AI & 0 & 3,748 & 62.2 & 75.2 & $-$12.99 & $-$12.51 \\
AI-AI & 1 & 19,424 & 12.5 & 11.9 & +0.62 & +1.06 \\
AI-AI & 2+ & 39,307 & 0.5 & 0.6 & $-$0.08 & 0.00 \\
\bottomrule
\end{tabular}
\end{table}

\paragraph{Pinned and calibrated zeros.} A zero gap arises in two ways. A \emph{pinned} zero occurs when a player plays almost only cards with $\hat{p} = 0$: then $\hat{p} = y = 0$ on nearly every play, and the gap cannot move whatever the partner does. A \emph{calibrated} zero occurs when a player plays uncertain cards and they fail at the literal rate. Neither kind shows that the partner's choices were uninformative. The converse of Corollary~\ref{cor:nonzero} does not hold, and an aggregate zero can also arise from cancellation between strata. A calibrated zero is evidence that outcomes on the selected plays matched the literal reference; a pinned zero is not evidence either way. In the AI-AI corpus, Simple, IGGI, Internal and Outer had pinned zeros (35{,}496 plays, all at $\hat{p} = 0$). Piers (+0.01~pp [$-$0.45, +0.46], mean $\hat{p}$ 8.3\%) and Van den Bergh (+0.40~pp [$-$0.10, +0.89], mean $\hat{p}$ 8.2\%) played uncertain cards on about 30\% of their plays. The calibrated-zero evidence in that corpus therefore comes from these two agents. On their plays with $\hat{p} > 0$ the gap was +0.04~pp [$-$1.54, +1.59] for Piers (3{,}076 plays) and +1.37~pp [$-$0.33, +3.05] for Van den Bergh (2{,}456 plays). Table~\ref{tab:pinned_zero} (Appendix~\ref{app:convgap_by_subject}) gives the share of plays at $\hat{p} = 0$ and the gap on the remaining plays for every setting. Appendix~\ref{app:llm_modern} applies the distinction to the LLM partners of Full.

\paragraph{Reading each side of a pair.} By Corollary~\ref{cor:nonzero}, a nonzero population gap on one player's plays indicates, under (i) and (iii), that the other player's choices depended on hidden cards. The two sides of a human-AI pair can therefore differ. With Intentional they did. One-touch human plays had a gap of +30.4~pp (participant-cluster CI [+29.3, +31.5]). Zero-touch human plays had a gap of $-$14.5~pp ([$-$16.9, $-$5.9]; 390 plays, mean $\hat{p}$ 73.0\%, failure 87.4\%). Intentional's own gap, by contrast, was a pinned zero: it played almost only cards that hints had proven playable (mean $\hat{p}$ at play 0.07\%; 4{,}975 of its 4{,}979 plays had $\hat{p} = 0$; 91.7\% were two-or-more-touch plays; Table~\ref{tab:full_mechanism}). Its own gap is therefore uninformative about whether the humans' hints depended on hidden cards. The same holds for the HanabiData agent Outer (2{,}631 of its 2{,}633 plays had $\hat{p} = 0$).

\paragraph{Negative zero-touch gaps.} A negative zero-touch gap is consistent with a partner that hints playable cards first. A card left untouched by such a partner is then less likely to be playable than the literal reference assumes. The same sign appeared on zero-touch human plays with Outer ($-$16.1~pp [$-$24.0, $-$8.1]) but was not established with Full ($-$5.0~pp [$-$7.6, +7.3]). It also appeared in the AI-AI games, where the zero-touch stratum had a gap of $-$13.0~pp (Table~\ref{tab:receiver_hint}). Of these 3{,}748 AI-AI plays, 3{,}223 were Flawed's (Appendix~\ref{app:convgap_by_subject}).

\paragraph{Card-choice model.} The choice model is a conditional logit with the play turn as stratum and the cards in hand as alternatives; exactly one card is played per turn, and every turn has weight one. All covariates are measured before the play. C0 uses $\hat{p}$ and an indicator for $\hat{p}=0$. C1 adds two timing variables (T) and three state variables (S), so C1 $=$ C0+T+S. The timing variables are hinted-last-turn (the partner's previous action was a hint touching the card) and turns since the card was last hinted (capped at 10; 10 if never hinted). The state variables are the card's hint count (the number of hint actions that positively touched it), a newest-card indicator and the hand position. C0-flex replaces the $\hat{p}$ terms of C0 by a flexible function of $\hat{p}$ (defined below), and C0-flex+T, C0-flex+S and C1-flex $=$ C0-flex+T+S add the same blocks. Coefficients are log-odds per unit of the covariate, so the turns-since-hint coefficient is per turn elapsed since the hint. Table~\ref{tab:receiver_clogit} gives C1 for every setting with game-cluster bootstrap intervals (2{,}000 draws). For human-AI play these intervals are too narrow (see ``Intervals and weighting'' above); person-cluster intervals are in Table~\ref{tab:receiver_clogit_robust}. The AI-AI $\hat{p}$ coefficient is close to degenerate because the agents played a lowest-$\hat{p}$ card on 99.8\% of turns, and C1 added little to C0 (pseudo-$R^2$ 0.906 to 0.908).

\begin{table}[!ht]
\centering
\caption{Conditional choice model C1 (linear $\hat{p}$): coefficients with 95\% game-cluster bootstrap intervals (2{,}000 draws), in-sample McFadden pseudo-$R^2$ for C0 and C1, and the likelihood-ratio statistic of C1 against C0 (5 df). Play-weighted. Person-cluster intervals for the human settings are in Table~\ref{tab:receiver_clogit_robust}.}
\label{tab:receiver_clogit}
\small
\setlength{\tabcolsep}{4pt}
\textit{(a) Literal failure probability and timing}\par\smallskip
\begin{tabular}{lllll}
\toprule
Setting & $\hat{p}$ & $\hat{p}=0$ & hinted last turn & turns since hint \\
\midrule
Human-human & $-$7.09 [$-$7.47, $-$6.75] & $-$1.48 [$-$1.76, $-$1.22] & +0.24 [+0.03, +0.44] & $-$0.27 [$-$0.30, $-$0.23] \\
Human-AI (all partners) & $-$3.92 [$-$4.13, $-$3.71] & $-$0.67 [$-$0.86, $-$0.49] & +0.23 [+0.09, +0.37] & $-$0.39 [$-$0.41, $-$0.36] \\
Human-AI: Full & $-$4.06 [$-$4.52, $-$3.64] & $-$1.06 [$-$1.41, $-$0.67] & +0.24 [$-$0.04, +0.52] & $-$0.40 [$-$0.46, $-$0.36] \\
Human-AI: Intentional & $-$3.74 [$-$4.10, $-$3.40] & $-$1.12 [$-$1.41, $-$0.82] & +0.22 [$-$0.03, +0.49] & $-$0.37 [$-$0.41, $-$0.33] \\
Human-AI: Outer & $-$3.87 [$-$4.22, $-$3.54] & $-$0.05 [$-$0.35, +0.27] & +0.26 [+0.05, +0.47] & $-$0.35 [$-$0.38, $-$0.32] \\
AI-AI & $-$86.45 [$-$115.65, $-$69.35] & $-$4.91 [$-$6.54, $-$3.82] & +0.41 [+0.16, +0.65] & $-$0.05 [$-$0.10, $-$0.02] \\
\bottomrule
\end{tabular}

\bigskip
\textit{(b) State variables and fit}\par\smallskip
\begin{tabular}{lllllr}
\toprule
Setting & hint count & newest & position & Pseudo-$R^2$ C0 / C1 & LR \\
\midrule
Human-human & +0.23 [+0.13, +0.34] & +1.02 [+0.89, +1.16] & $-$0.01 [$-$0.06, +0.05] & 0.558 / 0.702 & 4,184 \\
Human-AI (all partners) & +0.39 [+0.32, +0.46] & +0.24 [+0.13, +0.35] & +0.02 [$-$0.01, +0.05] & 0.374 / 0.647 & 13,562 \\
Human-AI: Full & +0.72 [+0.56, +0.93] & +0.26 [+0.06, +0.48] & 0.00 [$-$0.06, +0.05] & 0.366 / 0.697 & 5,090 \\
Human-AI: Intentional & +0.74 [+0.59, +0.90] & +0.50 [+0.30, +0.70] & $-$0.03 [$-$0.09, +0.03] & 0.368 / 0.663 & 4,849 \\
Human-AI: Outer & +0.15 [+0.06, +0.24] & +0.07 [$-$0.08, +0.24] & +0.07 [+0.02, +0.12] & 0.388 / 0.602 & 3,846 \\
AI-AI & +0.10 [+0.03, +0.17] & +0.52 [+0.43, +0.62] & $-$0.01 [$-$0.03, +0.02] & 0.906 / 0.908 & 434 \\
\bottomrule
\end{tabular}
\end{table}

\paragraph{Robustness: flexible $\hat{p}$ and person-level clustering.} A linear term in $\hat{p}$ may leave residual dependence on $\hat{p}$ that the timing variables absorb. We therefore also fitted C1-flex. Its flexible function of $\hat{p}$ consists of the indicator for $\hat{p}=0$, nine indicators for $\hat{p}$ in $(0, 0.1], \ldots, (0.8, 0.9]$, a linear term, and an indicator that the card attains the lowest $\hat{p}$ in hand. A restricted cubic spline in $\hat{p}$ (knots at 0.05, 0.35, 0.65, 0.85 and 0.98, with the same two indicators) gave timing coefficients within 0.02 of the bin-indicator model (turns since hint $-$0.27 and $-$0.39; hinted last turn +0.20 and +0.19). Intervals are percentile bootstraps over persons with the model refitted on each of 400 resamples. The person is the acting player in human-human play (155 players) and the participant in human-AI play (228 participants). Table~\ref{tab:receiver_clogit_robust} gives the results. The robustness analysis covers the two human settings, pooled over partners.

\begin{table}[!ht]
\centering
\caption{Timing and state coefficients of the choice model under a flexible function of $\hat{p}$ (C1-flex) and under linear $\hat{p}$ (C1), with 95\% person-cluster bootstrap intervals (400 refits; acting player in human-human play, participant in human-AI play). The last column refits the flexible model without the two persons who contribute the most turns (no interval).}
\label{tab:receiver_clogit_robust}
\footnotesize
\begin{tabular}{llllr}
\toprule
Setting & Coefficient & Flexible $\hat{p}$ (C1-flex) & Linear $\hat{p}$ (C1) & Two largest removed \\
\midrule
Human-human & turns since hint & $-$0.26 [$-$0.32, $-$0.21] & $-$0.27 [$-$0.33, $-$0.21] & $-$0.26 \\
Human-human & hinted last turn & +0.20 [$-$0.05, +0.44] & +0.24 [+0.00, +0.49] & +0.13 \\
Human-human & hint count & +0.51 [+0.33, +0.70] & +0.23 [+0.09, +0.40] & +0.51 \\
Human-human & newest & +1.00 [+0.85, +1.17] & +1.02 [+0.87, +1.20] & +1.03 \\
Human-human & position & $-$0.06 [$-$0.13, +0.02] & $-$0.01 [$-$0.08, +0.08] & $-$0.02 \\
\addlinespace
Human-AI (all partners) & turns since hint & $-$0.38 [$-$0.47, $-$0.36] & $-$0.39 [$-$0.47, $-$0.36] & $-$0.45 \\
Human-AI (all partners) & hinted last turn & +0.21 [$-$0.31, +0.71] & +0.23 [$-$0.25, +0.73] & +0.40 \\
Human-AI (all partners) & hint count & +0.41 [+0.17, +0.49] & +0.39 [+0.15, +0.47] & +0.22 \\
Human-AI (all partners) & newest & +0.23 [$-$0.46, +0.76] & +0.24 [$-$0.45, +0.78] & +0.04 \\
Human-AI (all partners) & position & +0.02 [$-$0.16, +0.09] & +0.02 [$-$0.16, +0.09] & $-$0.11 \\
\bottomrule
\end{tabular}
\end{table}

\paragraph{Held-out evaluation.} Models were also compared out of sample by grouped five-fold cross-validation, with folds formed from whole games or from whole persons. Table~\ref{tab:receiver_heldout} gives the gain in log-likelihood per held-out play turn. Adding the two timing variables to C0-flex gained +0.207 nats per turn in held-out human-human games and +0.203 in held-out players. The gains in human-AI play were +0.395 in held-out games and +0.383 in held-out participants. Given the state variables as well, the timing gain was +0.084 and +0.079 in human-human play and +0.190 and +0.160 in human-AI play. All of these gains were positive in five of five folds. For scale, uniform choice among five cards scores $-$1.61 nats per turn. In held-out games C0-flex scored $-$0.69 (human-human) and $-$0.97 (human-AI), and C0-flex+T scored $-$0.49 and $-$0.57. C0-flex added +0.02 to +0.04 nats per turn over C0 and did not change top-1 accuracy. The intervals in Table~\ref{tab:receiver_heldout} resample held-out clusters with the out-of-fold predictions fixed. They are conditional on the fitted fold models and the fold assignment and do not include refitting variability. In human-human play a person fold holds out acting players; their partners, and hence turns of the same games, can be in the training folds. Game folds hold out whole games but not persons.

\begin{table}[!ht]
\centering
\caption{Held-out gains from grouped five-fold cross-validation. Log-likelihood rows are in nats per held-out play turn; top-1 rows are the change in the fraction of turns on which the played card has the highest predicted utility. Model labels as in the text: C0, linear $\hat{p}$; C0-flex, flexible $\hat{p}$; T, the two timing variables; S, the three state variables. Brackets are 95\% cluster-bootstrap intervals conditional on the fitted fold models. All log-likelihood gains were positive in 5 of 5 folds.}
\label{tab:receiver_heldout}
\footnotesize
\setlength{\tabcolsep}{4pt}
\resizebox{\textwidth}{!}{\begin{tabular}{lllll}
\toprule
 & \multicolumn{2}{c}{Human-human} & \multicolumn{2}{c}{Human-AI (all partners)} \\
Contrast & Game folds (425) & Player folds (155) & Game folds (2{,}035) & Participant folds (228) \\
\midrule
Timing over flexible $\hat{p}$ (C0-flex+T $-$ C0-flex) & +0.207 [+0.185, +0.229] & +0.203 [+0.161, +0.241] & +0.395 [+0.379, +0.410] & +0.383 [+0.343, +0.428] \\
Timing given state (C1-flex $-$ C0-flex+S) & +0.084 [+0.071, +0.097] & +0.079 [+0.056, +0.097] & +0.190 [+0.177, +0.202] & +0.160 [+0.080, +0.234] \\
State over flexible $\hat{p}$ (C0-flex+S $-$ C0-flex) & +0.148 [+0.130, +0.166] & +0.147 [+0.119, +0.178] & +0.211 [+0.197, +0.226] & +0.202 [+0.153, +0.232] \\
Flexible over linear $\hat{p}$ (C0-flex $-$ C0) & +0.019 [+0.013, +0.024] & +0.018 [+0.011, +0.024] & +0.038 [+0.032, +0.043] & +0.034 [+0.021, +0.047] \\
\addlinespace
Top-1: timing over flexible $\hat{p}$ & +0.097 [+0.089, +0.106] & +0.098 [+0.087, +0.109] & +0.155 [+0.148, +0.162] & +0.154 [+0.142, +0.165] \\
Top-1: timing given state & +0.025 [+0.017, +0.032] & +0.026 [+0.015, +0.036] & +0.054 [+0.048, +0.060] & +0.027 [$-$0.013, +0.070] \\
\bottomrule
\end{tabular}}
\end{table}

\paragraph{What is and is not robust.} The turns-since-hint coefficient was robust. Under flexible $\hat{p}$ and person clustering it was $-$0.26 [$-$0.32, $-$0.21] in human-human play and $-$0.38 [$-$0.47, $-$0.36] in human-AI play, close to the linear-$\hat{p}$ estimates. The newest-card coefficient was robust in human-human play (+1.00 [+0.85, +1.17]). Three results were not robust. First, the hinted-last-turn coefficient had person-cluster intervals that included zero in both settings (+0.20 [$-$0.05, +0.44] and +0.21 [$-$0.31, +0.71]). Second, the newest-card coefficient in human-AI play (+0.23 [$-$0.46, +0.76]) did not survive participant clustering and was +0.04 without the two largest participants. Third, in participant-held-out human-AI folds the state variables did not transfer. C1-flex ($-$0.615 nats per turn) scored below C0-flex+T ($-$0.594). The top-1 gain of timing given state had an interval that included zero. The timing result rests on the held-out gains, because the in-sample pseudo-$R^2$ increases in Table~\ref{tab:receiver_clogit} combine the timing and state variables (5 df).

\paragraph{Scope of the timing result.} The supported statement is that hint timing predicted card choice after flexible adjustment for $\hat{p}$, in held-out games and held-out persons. The model does not hold the literal state fixed. A scalar $\hat{p}$ is not a sufficient statistic of the literal state: cards of equal $\hat{p}$ can differ in rank, in their distribution over identities and in future value, and recently hinted cards may differ systematically on these. The timing association is observational, because a recently hinted card is also the card the partner chose to hint. The lowest bin of $\hat{p}$, $(0, 0.1]$, is sparse (58 and 159 card rows).

\paragraph{Descriptive summaries.} The played card attained the lowest $\hat{p}$ in hand on the following fractions of play turns: human-human 0.803, human-AI (all partners) 0.715, Full 0.724, Intentional 0.725, Outer 0.698, AI-AI 0.998. When the most recent hint to the acting player had arrived one or two turns earlier, the played card was the last-hinted card with probability 0.833 in human-human play ($n$=5{,}799). In human-AI play (all partners) the probability was 0.923 ($n$=11{,}557). By partner it was 0.936 with Full ($n$=3{,}630), 0.933 with Intentional ($n$=3{,}676) and 0.904 with Outer ($n$=4{,}251). In the AI-AI games it was 0.942 ($n$=53{,}929).
\FloatBarrier
\section{Statistical Methods}
\label{app:stats}

All tests are two-sided. Unless stated otherwise, confidence intervals are 95\% percentile bootstrap intervals from 10{,}000 draws (seed 42). The estimand of every interval in this appendix is a play-weighted (pooled) quantity: the convention gap of Eq.~\ref{eq:convgap}, the failure rate, or a difference of two such quantities. Equal-participant estimates are a different estimand and are given in Appendix~\ref{app:sensitivity}. Three kinds of interval appear in these appendices, each labeled where it is given. The default is the percentile bootstrap interval, with the resampling unit named (Table~\ref{tab:clusterci}). The second kind is the analytic normal or $t$ interval from CR0 cluster standard errors (CR0: the cluster-robust variance without small-sample correction; Table~\ref{tab:hhcluster}). The third is the $t$ interval on per-participant differences (Appendices~\ref{app:within} and~\ref{app:sensitivity}).

\paragraph{Bootstrap levels.}
Plays within a game, and games of the same person, are dependent. We therefore computed each interval at several resampling levels. The play level resamples plays and treats them as independent. The game level resamples whole games. The person level resamples whole acting players (human-human; hashed account) or whole participants (human-AI). For human-human play we added the dyad (unordered pair of players) and the connected component of the player graph (players are linked when they played a game together). For the AI-AI corpus we added the pairing (49 ordered agent pairings) and the acting agent (7 agents). In every draw both terms of the gap, the mean literal failure probability and the failure rate, were recomputed from the resampled clusters as ratios of sums.

Table~\ref{tab:clusterci} gives the intervals. The positive gaps remained after clustering, with wider intervals. The human-human interval widened from 1.34~pp (plays) to 3.26~pp (acting players) and 4.57~pp (components). The human-AI interval widened from 1.15~pp (plays) to 2.73~pp (participants). The per-partner intervals also widened at the participant level for Intentional and Outer. The Full interval changed little, but all three participant-level intervals rest on few effective participants (Appendix~\ref{app:sensitivity}); they describe resampling variability of this pooled sample, not of a population of participants. The AI-AI gap was $-$0.64~pp; its interval excluded zero under play, game and pairing resampling and included zero under acting-agent resampling, for which seven clusters are too few for a reliable interval. We describe it as near zero.

\begin{table}[!ht]
\centering
\caption{Play-weighted convention gap and 95\% percentile bootstrap intervals (pp; 10{,}000 draws) under each resampling unit. Width is the interval width, computed from the unrounded bounds. Human-AI rows use human plays. Per-partner participant intervals resample participants jointly across partners, as in Table~\ref{tab:convgap_agent}.}
\label{tab:clusterci}
\small
\begin{tabular}{llrcr}
\toprule
Setting (gap) & Resampling unit & Clusters & 95\% CI & Width \\
\midrule
Human-human (+26.16) & play & 9{,}017 & [+25.49, +26.83] & 1.34 \\
 & game & 425 & [+25.32, +27.01] & 1.68 \\
 & acting player & 155 & [+24.54, +27.80] & 3.26 \\
 & dyad & 95 & [+24.06, +28.31] & 4.25 \\
 & component & 65 & [+23.86, +28.43] & 4.57 \\
Human-AI (+16.38) & play & 15{,}472 & [+15.80, +16.95] & 1.15 \\
 & game & 2{,}035 & [+15.72, +17.06] & 1.34 \\
 & participant & 228 & [+15.04, +17.76] & 2.73 \\
\quad partner Full (+24.1) & play & 4{,}769 & [+23.2, +25.1] & 1.9 \\
 & game & 682 & [+23.1, +25.2] & 2.1 \\
 & participant & 129 & [+22.9, +24.9] & 2.0 \\
\quad partner Intentional (+20.2) & play & 5{,}122 & [+19.3, +21.2] & 1.9 \\
 & game & 646 & [+19.2, +21.3] & 2.1 \\
 & participant & 131 & [+18.0, +22.5] & 4.4 \\
\quad partner Outer (+6.2) & play & 5{,}581 & [+5.2, +7.2] & 2.0 \\
 & game & 707 & [+5.2, +7.2] & 2.0 \\
 & participant & 122 & [+4.9, +7.5] & 2.7 \\
AI-AI ($-$0.64) & play & 62{,}479 & [$-$0.79, $-$0.49] & 0.29 \\
 & game & 4{,}900 & [$-$0.79, $-$0.49] & 0.29 \\
 & pairing & 49 & [$-$1.15, $-$0.23] & 0.92 \\
 & acting agent & 7 & [$-$2.18, +0.16] & 2.34 \\
\bottomrule
\end{tabular}
\end{table}

\paragraph{Dyadic dependence in human-human play.}
In human-human play each person appears as the actor on some plays and as the partner on others, so clustering by acting player alone understates the dependence. We identified the partner of every play from the game logs. This gave 159 players in any role (155 acting players and 4 who appear only as partners), 95 unordered dyads and 65 connected components of the player graph; the largest component holds 12.9\% of plays. For each dependence structure we computed the analytic variance of the pooled gap from the residuals $r_i=(\hat{p}_i-y_i)-\hat{G}$ with cluster scores $u_c=\sum_{i\in c} r_i$ and $V=\sum_c u_c^2/N^2$ (CR0, no small-sample factor). The effective number of clusters is $(\sum_c u_c^2)^2/\sum_c u_c^4$. Table~\ref{tab:hhcluster} gives the normal interval and a $t$ interval whose degrees of freedom are the effective number of clusters minus one. This degrees-of-freedom rule is a heuristic without a coverage result, and we use the $t$ interval only as a sensitivity interval. Both intervals are analytic. The bootstrap intervals for the same resampling units (Table~\ref{tab:clusterci}) agree with the normal intervals to within 0.1~pp for games, acting players, dyads and components.

The standard error rose from 0.84~pp (acting player) to 1.11~pp (dyad), 1.16~pp (dyadic-robust, which allows correlation between any two plays that share a player) and 1.18~pp (components). Two-way clustering by actor and partner (0.86~pp) barely differed from acting-player clustering, because it ignores that the same person is the actor on some plays and the partner on others. The connected components share no player, so they are independent under any within-player, within-dyad or within-game dependence, and every other structure is nested in them. Their weakness is concentration (5.8 effective clusters), which makes the normal interval and the component bootstrap optimistic. The effective-df $t$ interval for components is $[+23.09, +29.24]$~pp, wider than the others. Under every structure the human-human gap stayed positive; the lowest lower bound was +23.1~pp.

\begin{table}[!ht]
\centering
\caption{Human-human play-weighted convention gap (+26.16~pp; 9{,}017 plays, 425 games) under alternative dependence structures. SE is the CR0 standard error of the pooled gap; intervals are analytic 95\% intervals (pp; normal, and $t$ with the effective number of clusters minus one as degrees of freedom), not bootstrap intervals. For two-way clustering the cluster counts are actors / partners.}
\label{tab:hhcluster}
\small
\begin{tabular}{lrrccc}
\toprule
Dependence structure & Clusters & Effective & SE & Normal CI & $t$ CI (effective df) \\
\midrule
None (independent plays) & 9{,}017 & 9{,}017 & 0.34 & [+25.49, +26.84] & -- \\
Game & 425 & 175.5 & 0.43 & [+25.32, +27.00] & [+25.32, +27.01] \\
Acting player & 155 & 12.6 & 0.84 & [+24.52, +27.81] & [+24.33, +28.00] \\
Two-way actor $\times$ partner & 155 / 157 & 12.6 / 13.0 & 0.86 & [+24.47, +27.85] & -- \\
Unordered dyad & 95 & 5.1 & 1.11 & [+24.00, +28.33] & [+23.13, +29.20] \\
Dyadic-robust (any shared player) & 159 & 10.9 & 1.16 & [+23.89, +28.44] & [+23.58, +28.75] \\
Connected components & 65 & 5.8 & 1.18 & [+23.86, +28.47] & [+23.09, +29.24] \\
\bottomrule
\end{tabular}
\end{table}

\paragraph{Partner contrasts with shared participants.}
Participants in the human-AI study met more than one AI partner: 69 of the 182 participants in the Full and Outer conditions played with both, and they supply 88\% of the plays in each. A standard error of the form $\sqrt{\mathrm{SE}_A^2+\mathrm{SE}_B^2}$, from separate per-condition standard errors, is valid only for disjoint clusters; here it would drop the covariance between the two condition means that shared participants induce. We therefore use the joint influence of each participant $c$ in the union of clusters,
\[
u_c=\sum_{i\in A,c}\frac{y_i-\bar{y}_A}{N_A}-\sum_{i\in B,c}\frac{y_i-\bar{y}_B}{N_B},\qquad \mathrm{SE}=\Bigl(\sum_c u_c^2\Bigr)^{1/2},
\]
with a missing condition contributing zero (CR0). For the gap, $y_i$ is replaced by $\hat{p}_i - y_i$. The estimand is the play-weighted difference of pooled means under both formulas. For disjoint clusters the two formulas coincide, so game-clustered values do not depend on the choice. The joint participant bootstrap resamples participants from the union and carries each drawn participant's plays in both conditions.

The choice of formula matters here. The participant-clustered $z$ of the Full-versus-Outer difference in human failure rate (14.4\% versus 34.4\%, $-$20.0~pp) is $-17.5$ with the joint formula and would be $-6.3$ with the disjoint-cluster formula. The latter is small because it treats the two largest participants' Full plays and Outer plays as independent samples; with these two participants removed, the disjoint-cluster and joint values are $-14.9$ and $-15.9$. The joint $z$ needs a caveat. Its variance is carried by few participants: the effective number of clusters is 2.9, and two participants supply 73\% of the variance. A normal reference distribution is not reliable in that case, and the percentile bootstrap shares the weakness. As a conservative companion we report the leave-one-participant-out jackknife, which gives $z=-13.5$ for this contrast. For the contrasts with Outer we state only $p<0.001$. The Full-Intentional contrasts are weaker: the failure-rate contrast does not reach $|z|=1.96$ ($z=-1.8$, jackknife $-1.4$), and the gap contrast gives $z=+3.4$ but jackknife $z=+2.6$. Table~\ref{tab:partnercontrast} gives all three contrasts for the failure rate and for the gap, with the values after removing the two largest participants.

\begin{table}[!ht]
\centering
\caption{Play-weighted partner contrasts on human plays in human-AI games (pp). CI: 95\% joint participant bootstrap. $z$: joint-influence CR0 with participant clusters; Eff.: effective number of participant clusters; Jack.\ $z$: leave-one-participant-out jackknife. The last column gives the difference and its 95\% CI after removing the two largest participants (Appendix~\ref{app:sensitivity}).}
\label{tab:partnercontrast}
\footnotesize
\setlength{\tabcolsep}{4pt}
\begin{tabular}{llcrrrc}
\toprule
Outcome & Contrast & Difference [95\% CI] & $z$ & Eff. & Jack.\ $z$ & Without two largest \\
\midrule
Failure rate & Full $-$ Outer & $-$20.04 [$-$22.19, $-$17.63] & $-$17.5 & 2.9 & $-$13.5 & $-$19.12 [$-$21.34, $-$16.51] \\
 & Intentional $-$ Outer & $-$16.79 [$-$18.98, $-$14.91] & $-$16.7 & 4.2 & $-$13.8 & $-$17.41 [$-$19.94, $-$14.71] \\
 & Full $-$ Intentional & $-$3.25 [$-$6.20, $-$0.26] & $-$1.8 & 1.7 & $-$1.4 & $-$1.71 [$-$3.33, +0.05] \\
Gap & Full $-$ Outer & +17.94 [+16.44, +19.13] & +30.1 & 6.7 & +25.7 & +17.55 [+15.53, +19.39] \\
 & Intentional $-$ Outer & +14.05 [+12.50, +16.12] & +16.5 & 3.2 & +13.5 & +14.93 [+12.62, +17.10] \\
 & Full $-$ Intentional & +3.89 [+1.63, +5.87] & +3.4 & 1.8 & +2.6 & +2.63 [+0.85, +4.46] \\
\bottomrule
\end{tabular}
\end{table}

Full and Intentional differed from Outer on both outcomes under every statistic in Table~\ref{tab:partnercontrast}. The Full-versus-Intentional contrast rests on fewer than two effective clusters, and its sign reverses under equal-participant weighting (Appendices~\ref{app:sensitivity} and~\ref{app:convgap_detail}). Its failure-rate bootstrap interval excludes zero, but its $z$ statistics do not reach 1.96 and the interval without the two largest participants includes zero. Both of its gap intervals exclude zero.

\paragraph{Success on zero-touch plays.}
The success rate on zero-touch plays (no hint action had positively touched the played card) was 77.7\% in human-human play ($n=484$ plays) and 15.7\% on human plays in human-AI games ($n=778$). The two corpora share no game and no person, so the independent-groups formula applies. The two-proportion statistic was $z=27.0$ with unpooled and $z=21.9$ with pooled play-level standard errors. With CR0 standard errors it was $z=21.7$ clustered by game (145 and 435 games) and $z=15.8$ clustered by person (72 players and 82 participants; 6.2 effective clusters; jackknife $z=14.4$). The person-clustered value is the one to read; given the few effective clusters we state only $p<0.001$.

\paragraph{Which statistics are cluster-aware.}
The following statistics are cluster-aware:
\begin{itemize}
\item the setting-level and per-partner gap intervals (Tables~\ref{tab:clusterci} and~\ref{tab:hhcluster});
\item the partner contrasts (Table~\ref{tab:partnercontrast}) and the comparison of the previous paragraph;
\item the sensitivity analyses of Appendix~\ref{app:sensitivity} (game- and participant-cluster bootstrap);
\item the pair-type gaps of Appendix~\ref{app:skill} (player-cluster bootstrap);
\item the hint-level statistics of Appendix~\ref{app:hintquality} (participant-cluster bootstrap of ratio-of-sums statistics, with participants resampled jointly across partners);
\item the card-choice models of Appendix~\ref{app:receiver} (game- and person-cluster bootstrap, with held-out games and persons);
\item the game-level intervals on the AI-AI, OBL and LLM games (for AI-AI also the pairing and acting-agent levels of Table~\ref{tab:clusterci}), which are conditional on the fixed agents, checkpoints and pairings.
\end{itemize}
The within-participant contrasts of Appendix~\ref{app:within} use the participant as the unit ($t$ interval and exact sign test on per-participant differences). They estimate an equal-participant quantity, not the pooled one.

The following statistics are descriptive only, without adjustment for clustering:
\begin{itemize}
\item play-level intervals wherever a table labels them as such;
\item the play-level one-way ANOVA of the gap over pair types (Appendix~\ref{app:skill});
\item the game-score ANOVAs, Tukey comparisons and Welch tests (Appendix~\ref{app:gamescore}), whose unit is the game although participants recur across partner groups;
\item the logistic regressions of Appendix~\ref{app:logistic}.
\end{itemize}

\paragraph{Game score analyses (Appendix~\ref{app:gamescore}).} All game-score ANOVAs are one-way with the final score of a game as the dependent variable. The factor is the setting (3 levels), the pair type (6 levels), the AI partner (3 levels) or the AI-AI agent (7 levels). Tukey HSD is used for post-hoc comparisons, and effect sizes are $\eta^2$. These tests treat games as independent and are descriptive (see above).

\paragraph{Logistic regression (Appendix~\ref{app:logistic}).} The dependent variable is the failure (0 or 1) of each play. Model~1 (M1) has the literal failure probability $\hat{p}$ as its only predictor. Model~2 (M2) adds four context covariates: the number of hint actions that positively touched the played card, the life tokens remaining and the number of cards left in the deck. The fourth is the game phase, in three bins of the turn number (at most 15, 16 to 35, above 35). Model~3 (M3) adds a categorical covariate to M2: the skill level of the acting player for human-human data and the AI partner for human-AI data. Models were fitted by maximum likelihood with \texttt{statsmodels.Logit}; pseudo-$R^2$ is McFadden's. The fits are in-sample and unclustered.

\paragraph{Change of the per-game gap with experience (Appendix~\ref{app:learning}).} Bayesian hierarchical models with random slopes assess whether the per-game gap changes with experience. Partial pooling shrinks individual slopes toward the population mean, which suits the small number of games per person. The per-game gap is selected on the whole game and falls outside Proposition~\ref{prop:zero}; these models are descriptive.

\textit{Human-human model.} The dependent variable is the per-game convention gap (mean $\hat{p}$ minus failure rate over the plays of one player in one game). The model is
\[
\text{gap}_{ij} = \alpha + \alpha_i + (\beta + \beta_i) \times \text{game\_seq}_{ij} + \varepsilon_{ij},
\]
where $\alpha$ is the population intercept, $\alpha_i \sim \mathcal{N}(0, \sigma_\alpha)$ the random intercept of player $i$, $\beta$ the population slope, $\beta_i \sim \mathcal{N}(0, \sigma_\beta)$ the random slope, and $\varepsilon_{ij} \sim \mathcal{N}(0, \sigma)$ the residual. Priors were the weakly informative defaults of \texttt{bambi}. The game index enters in units of 10 games (slopes are reported per game). Sampling used 4 chains of 2{,}000 draws (3{,}000 tuning) with \texttt{target\_accept} = 0.99 (no divergent transition, $\hat{R} \leq 1.00$). There were $n = 415$ game-player observations from 20 players with at least 10 completed games (Appendix~\ref{app:bots}).

\textit{Human-AI model.} The corresponding human-AI model, with the AI partner as a fixed effect and its interaction with game sequence (10 participants, 23 participant-by-AI cells with at least 10 games), gave no usable fit, so no human-AI slope is reported (Appendix~\ref{app:learning}).
\FloatBarrier
\section{Gap by Skill Stratum (Descriptive)}
\label{app:skill}

The convention gap was positive in every pair-type stratum of the human-human data (Table~\ref{tab:skill}). The play-weighted gap was +29.1~pp for Expert-Expert pairs (95\% player-cluster bootstrap CI [+27.5, +30.7]) and +21.1~pp for Beginner-Beginner pairs ([+18.5, +23.6]). The intervals of E-I (+25.2~pp) and I-I (+25.4~pp) overlap that of E-E, and I-B (+27.2~pp, 9 players) has no interval, so E-E and B-B form the only pair of strata whose intervals do not overlap. These are acting-player intervals, which understate the dependence between plays (Appendix~\ref{app:stats}). Across the five pair types other than Expert-Beginner the gap ranged from +21.1 to +29.1~pp. The Expert-Beginner type (3 games, 69 plays, 3 players) had +38.0~pp and is too small to interpret. A play-level one-way ANOVA of the per-play residual $\hat{p}_i - y_i$ over the five strata gave $F(4, 8943) = 16.41$ ($N$=8{,}948 plays, $p < 0.001$); it ignores the clustering of plays in players (Appendix~\ref{app:stats}). Mean $\hat{p}$ was similar for E-E and I-I pairs (31.6\% vs.\ 30.2\%), whereas the observed failure rate differed (2.5\% vs.\ 4.8\%).

\paragraph{Selection caveat.} Skill levels are defined from each player's average end-of-game score, which is an outcome of the same games. The strata are therefore selected on later outcomes and fall outside the scope of Proposition~\ref{prop:zero}: a stratum's gap need not be zero even for literal play, and the ordering of the strata may partly reflect this selection. The within-stratum values are descriptive.

\paragraph{Imputed pair labels.} Eleven games (127 plays) carry a pair type imputed from the sole surviving player. In 10 games (121 plays) the partner's account was removed at Tier~2 of the bot screen, and in game 1762152 (6 plays; Appendix~\ref{app:bots}) the partner was a verified bot that never played. In all 11 the remaining player's plays are kept and both seats are labeled with that player's skill level (9 games I-I, 1 E-E, 1 B-B). The lower block of Table~\ref{tab:skill} excludes these games. The exclusion changed three strata by at most 0.4~pp (E-E +29.0, I-I +25.0, B-B +21.2~pp) and left the ordering unchanged. Skill levels computed from the remaining 414 games were the same for every player. The effect of the same exclusion on the pooled human-human gap is given in Appendix~\ref{app:sensitivity}.

\begin{table}[!ht]
\centering
\caption{Convention gap by pair type (human-human plays; play-weighted within stratum). Players are classified by individual average end-of-game score: expert ($\geq$22), intermediate ($\geq$15 and $<$22), beginner ($<$15). Corpus: completed games after the two-tier bot exclusion (Appendix~\ref{app:bots}). Upper block: all 425 games, in which 11 games (127 plays) carry a pair type imputed from the sole surviving player. Lower block: those 11 games excluded (414 games). CI: 95\% percentile bootstrap over acting players (10{,}000 resamples). Strata with fewer than 10 acting players (I-B, 9 players; E-B, 3 players and 3 games) are point estimates without a CI.}
\label{tab:skill}
\small
\begin{tabular}{lrrcccc}
\toprule
Pair Type & $N$ plays & Players & Mean $\hat{p}$ & Failure & Gap & 95\% CI \\
\midrule
\multicolumn{7}{l}{\emph{All 425 games (imputed labels kept)}} \\
Expert-Expert (E-E)             & 3,268 & 61 & 31.6\% &  2.5\% & +29.1~pp & [+27.5, +30.7] \\
Expert-Intermediate (E-I)       &   194 & 12 & 28.8\% &  3.6\% & +25.2~pp & [+19.7, +30.4] \\
Intermediate-Intermediate (I-I) & 3,708 & 44 & 30.2\% &  4.8\% & +25.4~pp & [+22.4, +28.8] \\
Intermediate-Beginner (I-B)     &   279 &  9 & 36.1\% &  9.0\% & +27.2~pp & -- \\
Beginner-Beginner (B-B)         & 1,499 & 39 & 33.1\% & 12.0\% & +21.1~pp & [+18.5, +23.6] \\
Expert-Beginner (E-B)           &    69 &  3 & 42.4\% &  4.3\% & +38.0~pp & -- \\
\midrule
\multicolumn{7}{l}{\emph{11 imputed-label games excluded (414 games)}} \\
Expert-Expert (E-E)             & 3,255 & 60 & 31.6\% &  2.5\% & +29.0~pp & [+27.4, +30.5] \\
Expert-Intermediate (E-I)       &   194 & 12 & 28.8\% &  3.6\% & +25.2~pp & [+19.7, +30.4] \\
Intermediate-Intermediate (I-I) & 3,600 & 43 & 29.7\% &  4.7\% & +25.0~pp & [+22.0, +28.5] \\
Intermediate-Beginner (I-B)     &   279 &  9 & 36.1\% &  9.0\% & +27.2~pp & -- \\
Beginner-Beginner (B-B)         & 1,493 & 39 & 33.0\% & 11.9\% & +21.2~pp & [+18.6, +23.7] \\
Expert-Beginner (E-B)           &    69 &  3 & 42.4\% &  4.3\% & +38.0~pp & -- \\
\bottomrule
\end{tabular}
\end{table}
\FloatBarrier
\section{Convention Gap Across All Settings}
\label{app:convgap_by_subject}

\begin{figure}[!t]
\centering
\includegraphics[width=\textwidth]{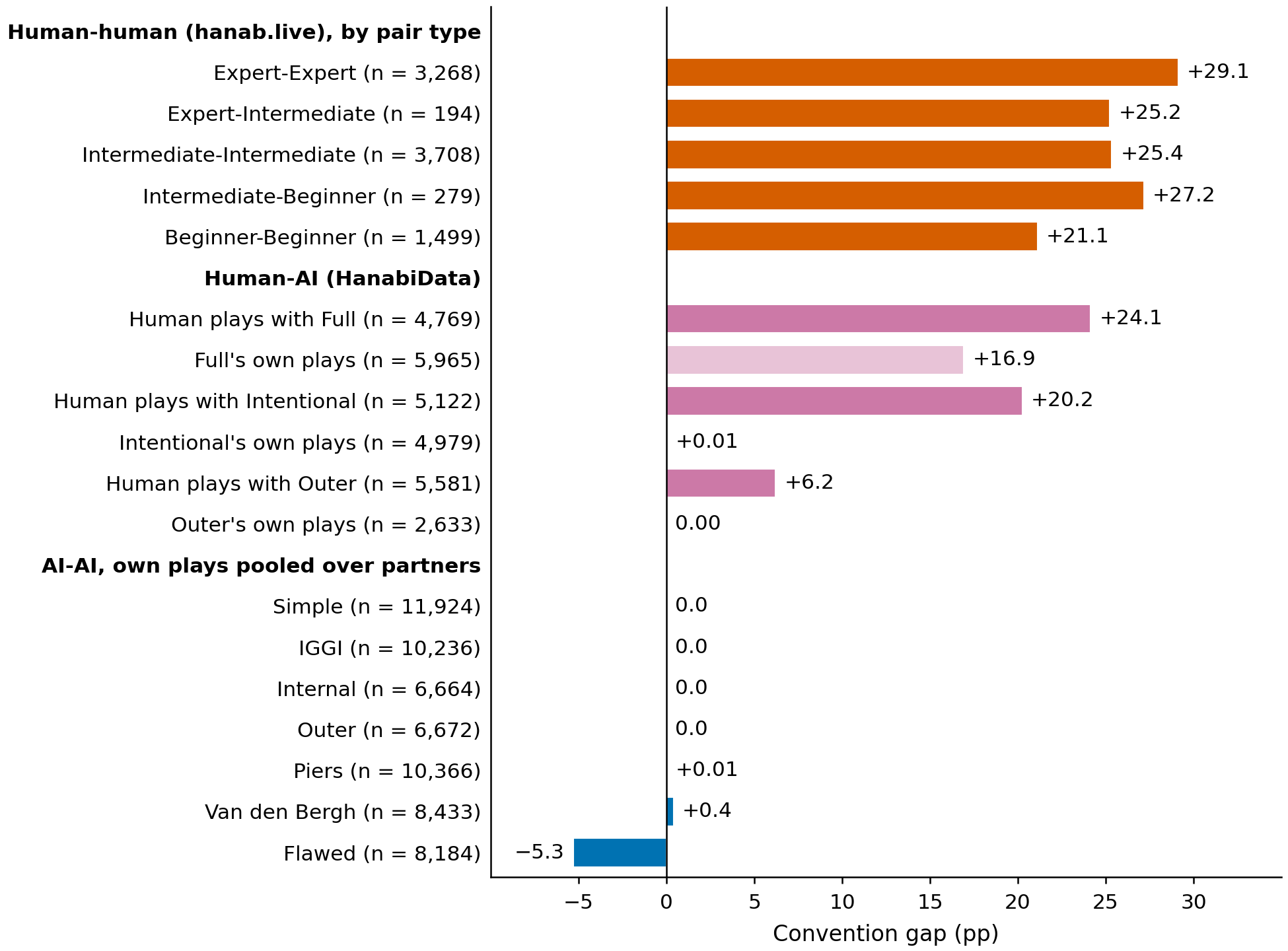}
\caption{Play-weighted convention gap (mean $\hat{p}$ minus failure rate, pp) by type of acting player; $n$ is the number of plays. Orange: human-human by pair type (skill levels defined by individual average score: expert $\geq$22, intermediate $\geq$15 and $<$22, beginner $<$15; the Expert-Beginner type, 3 games, is not shown). Pink: human-AI games (HanabiData); dark bars are human plays and light bars are the AI's own plays with each partner. Blue: each of the seven agents in the AI-AI games (six reimplemented rule-based agents, among them our Outer, and the Simple baseline; not the HanabiData agents, Appendix~\ref{app:agents_identity}), own plays pooled over all partners. Among AI plays in human-AI games, Full's own-play gap was +16.9~pp; the own-play zeros of Intentional and Outer are pinned (Appendix~\ref{app:full_mechanism}). Among the AI-AI agents Flawed had $-$5.3~pp; Piers (+0.01~pp) and Van den Bergh (+0.40~pp) are calibrated zeros, and the four bars at 0.0 are pinned (Appendix~\ref{app:receiver}). No intervals are drawn.}
\label{fig:convgap_all}
\end{figure}

Figure~\ref{fig:convgap_all} gives the play-weighted convention gap by type of acting player in the three corpora. The human-human gaps by pair type were +21.1 to +29.1~pp across the five pair types shown (the Expert-Beginner type, 3 games and 69 plays, had +38.0~pp). The gaps on human plays in human-AI games were +6.2 to +24.1~pp. The two ranges overlap: the smallest pair-type gap (Beginner-Beginner, +21.1~pp) was below the human gap with Full (+24.1~pp). In the AI-AI games (62{,}479 plays, 4{,}900 games; Appendix~\ref{app:agents}), the pooled gap was $-$0.64~pp [$-$0.79, $-$0.49] (game-cluster CI). By acting agent the gaps were 0 for Simple, IGGI, Internal and our Outer, +0.01~pp [$-$0.45, +0.46] for Piers and +0.40~pp [$-$0.10, +0.89] for Van den Bergh. Flawed had $-$5.31~pp [$-$6.14, $-$4.50] (8{,}184 plays, mean $\hat{p}$ 41.8\%, failure 47.1\%). The four zeros are pinned (Appendix~\ref{app:receiver}): these agents played only cards with $\hat{p} = 0$ (35{,}496 plays, no failures). Excluding Flawed's own plays, the pooled AI-AI gap was +0.06~pp over 54{,}295 plays.

We also binned $\hat{p}$ (width 0.1, right-closed as in main-text Figure~\ref{fig:calibration}b). Within these bins Piers and Van den Bergh showed departures (mean $\hat{p}$ minus failure rate) of both signs between $-$1.6 and +5.5~pp. The largest was Van den Bergh's +5.5~pp at $\hat{p} \in (0.2, 0.3]$ ($n$=477). All 4{,}439 AI-AI plays with $\hat{p} > 0.5$ were Flawed's, and in the three bins between 0.5 and 0.8 they failed 9.6 to 12.6~pp more often than $\hat{p}$ predicts. Flawed's negative gap was concentrated in its zero-touch plays (3{,}223 of its 8{,}184 plays, 39\%; mean $\hat{p}$ 68.6\%, failure 82.3\%, gap $-$13.7~pp). On its one-touch plays the gap was +0.2~pp. This pattern is consistent with partners that hint playable cards first, so that a card they leave untouched is less likely to be playable than the literal reference assumes. We did not test that explanation directly.

\paragraph{Plays at $\hat{p} = 0$ and the gap on the remaining plays.} A play of a card with $\hat{p} = 0$ cannot fail under the literal reference, and no such play failed in any corpus, so these plays contribute exactly zero to the gap. Table~\ref{tab:pinned_zero} gives their share and the gap on the plays with $\hat{p} > 0$. Four of the seven AI-AI agents never played a card with $\hat{p} > 0$. Piers and Van den Bergh are the only AI-AI agents with plays at $\hat{p} > 0$ besides Flawed; their intervals on these plays included zero (calibrated zeros, Appendix~\ref{app:receiver}). The human gaps on these plays were +45.2~pp (human-human) and +25.2~pp (human-AI). The own-play zeros of Intentional and Outer in HanabiData are pinned (Appendix~\ref{app:full_mechanism}).

\begin{table}[!ht]
\centering
\caption{Share of plays with $\hat{p} = 0$ and the play-weighted gap restricted to plays with $\hat{p} > 0$. Mean $\hat{p}$ and failure in \%, gap in pp, with 95\% percentile cluster-bootstrap intervals (10{,}000 draws). The last column gives the cluster and the number of clusters that have any play. The bootstrap resamples only clusters with at least one play at $\hat{p} > 0$. For human-AI these are 225, 120 and 128 participants (all partners, Full, Intentional); for AI-AI, 2{,}985, 975 and 920 games (all agents, Piers, Van den Bergh). In the other rows with an interval every cluster has a play at $\hat{p} > 0$. ``--'': fewer than five plays with $\hat{p} > 0$.}
\label{tab:pinned_zero}
\footnotesize
\setlength{\tabcolsep}{4pt}
\begin{tabular}{lrrrrrll}
\toprule
 & & $\hat{p} = 0$ & \multicolumn{4}{c}{Plays with $\hat{p} > 0$} & \\
\cmidrule(lr){4-7}
Setting & $N$ & (\%) & $N$ & Mean $\hat{p}$ & Failure & Gap [95\% CI] & Cluster \\
\midrule
Human-human & 9,017 & 42.1 & 5,219 & 54.3 & 9.1 & +45.20 [+43.24, +47.11] & acting player (155) \\
\addlinespace
Human-AI, human plays (all partners) & 15,472 & 35.1 & 10,044 & 60.2 & 34.9 & +25.23 [+21.83, +28.70] & participant (228) \\
Human-AI, human plays with Full & 4,769 & 33.7 & 3,164 & 58.1 & 21.7 & +36.38 [+33.80, +39.24] & participant (129) \\
Human-AI, human plays with Intentional & 5,122 & 36.8 & 3,235 & 60.0 & 27.9 & +32.06 [+26.59, +37.50] & participant (131) \\
Human-AI, human plays with Outer & 5,581 & 34.7 & 3,645 & 62.2 & 52.7 & +9.49 [+7.11, +12.10] & participant (122) \\
\addlinespace
Human-AI, own plays of Full & 5,965 & 38.4 & 3,673 & 52.4 & 25.0 & +27.48 [+21.18, +30.21] & participant (122) \\
Human-AI, own plays of Intentional & 4,979 & 99.9 & 4 & -- & -- & -- & participant (126) \\
Human-AI, own plays of Outer & 2,633 & 99.9 & 2 & -- & -- & -- & participant (112) \\
\addlinespace
AI-AI, all agents & 62,479 & 81.7 & 11,421 & 43.6 & 47.1 & $-$3.50 [$-$4.30, $-$2.71] & game (4{,}900) \\
AI-AI: Piers & 10,366 & 70.3 & 3,076 & 28.1 & 28.0 & +0.04 [$-$1.54, +1.59] & game (1{,}127) \\
AI-AI: Van den Bergh & 8,433 & 70.9 & 2,456 & 28.2 & 26.9 & +1.37 [$-$0.33, +3.05] & game (1{,}091) \\
AI-AI: Flawed & 8,184 & 28.0 & 5,889 & 58.0 & 65.4 & $-$7.38 [$-$8.56, $-$6.23] & game (1{,}300) \\
AI-AI: IGGI & 10,236 & 100.0 & 0 & -- & -- & -- & game (1{,}131) \\
AI-AI: Internal & 6,664 & 100.0 & 0 & -- & -- & -- & game (1{,}084) \\
AI-AI: Outer & 6,672 & 100.0 & 0 & -- & -- & -- & game (1{,}067) \\
AI-AI: Simple & 11,924 & 100.0 & 0 & -- & -- & -- & game (1{,}130) \\
\bottomrule
\end{tabular}
\end{table}

\paragraph{Calibration by bin of $\hat{p}$.} Table~\ref{tab:calibration_bins} lists the values plotted in main-text Figure~\ref{fig:calibration}b. Human-human plays failed less often than $\hat{p}$ predicts in every bin above the first; the first bin consists almost entirely of plays with $\hat{p} = 0$.

\begin{table}[!ht]
\centering
\caption{Calibration by bin of $\hat{p}$ (values of main-text Figure~\ref{fig:calibration}b). Each cell gives $N$; mean $\hat{p}$ (\%); failure rate (\%) with its 95\% game-cluster bootstrap interval (2{,}000 draws). Human-AI: human plays. The first row is given to two decimals. No AI-AI play had $\hat{p} > 0.8$.}
\label{tab:calibration_bins}
\footnotesize
\setlength{\tabcolsep}{4pt}
\begin{tabular}{llll}
\toprule
Bin of $\hat{p}$ & Human-human & Human-AI & AI-AI \\
\midrule
$[0, 0.1]$ & 3{,}840; 0.09; 0.00 [0.00, 0.00] & 5{,}523; 0.15; 0.13 [0.04, 0.23] & 51{,}332; 0.05; 0.04 [0.02, 0.06] \\
$(0.1, 0.2]$ & 390; 16.4; 2.1 [0.7, 3.8] & 747; 16.4; 14.1 [11.6, 16.6] & 1{,}816; 16.3; 15.6 [14.0, 17.3] \\
$(0.2, 0.3]$ & 229; 25.5; 4.8 [2.2, 7.7] & 415; 25.7; 10.6 [7.5, 13.8] & 1{,}265; 25.7; 21.2 [18.9, 23.5] \\
$(0.3, 0.4]$ & 629; 34.7; 6.4 [4.3, 8.5] & 975; 35.3; 16.6 [14.1, 18.9] & 3{,}280; 35.3; 35.1 [33.4, 36.8] \\
$(0.4, 0.5]$ & 1{,}130; 49.4; 6.8 [5.3, 8.4] & 1{,}476; 49.2; 19.2 [17.0, 21.4] & 347; 47.9; 41.5 [36.2, 47.0] \\
$(0.5, 0.6]$ & 561; 58.7; 8.6 [6.2, 11.1] & 953; 58.8; 26.4 [23.6, 29.3] & 489; 58.0; 70.6 [66.3, 74.7] \\
$(0.6, 0.7]$ & 1{,}153; 66.2; 8.8 [7.3, 10.7] & 2{,}088; 66.4; 31.3 [29.1, 33.5] & 2{,}405; 66.8; 78.9 [77.1, 80.7] \\
$(0.7, 0.8]$ & 887; 74.7; 11.0 [9.0, 13.2] & 1{,}817; 74.6; 36.3 [33.9, 38.9] & 1{,}545; 72.2; 81.8 [79.7, 83.8] \\
$(0.8, 0.9]$ & 126; 84.4; 17.5 [11.3, 24.2] & 273; 84.6; 51.3 [44.8, 57.3] & none \\
$(0.9, 1]$ & 72; 99.5; 97.2 [92.6, 100.0] & 1{,}205; 99.9; 99.8 [99.6, 100.0] & none \\
\bottomrule
\end{tabular}
\end{table}

The values plotted in main-text Figure~\ref{fig:hintcount_bars} are listed in Table~\ref{tab:hintcount}.

\paragraph{Hint strata.} Table~\ref{tab:hintcount} gives, for each stratum of the number of hint actions that positively touched the played card, the number of plays, mean $\hat{p}$, failure rate, gap and the ratio of gap to mean $\hat{p}$. The counts are per hint action, not per attribute type. The ratio expresses the gap as a fraction of its ceiling, the stratum's mean $\hat{p}$ (Appendix~\ref{app:robustness}). In human-human play the raw gap was largest in the zero-touch stratum (+46.2~pp against +35.6~pp at one touch), which also had the highest ceiling. As a fraction of the ceiling the order was reversed (0.67 against 0.87). Zero-touch plays were 5.4\% of the human-human plays and contributed 2.5~pp of the pooled +26.2~pp (share of plays times stratum gap); one-touch plays contributed 23.2~pp and 2+ plays 0.5~pp. In human-AI play the one-touch stratum had the largest raw gap.

In the 2+ stratum mean $\hat{p}$ was 0.5 to 12.8\%, so the gap there cannot exceed 0.5 to 12.8~pp whatever the partner did. On most 2+ plays both the color and the rank of the played card had been hinted: 2{,}419 of 2{,}649 human-human plays (91.3\%), 3{,}618 of 4{,}194 human plays in human-AI games (86.3\%) and 38{,}066 of 39{,}307 AI-AI plays (96.8\%). The card's identity was then known, so these plays contribute zero to the gap.

\begin{table}[!ht]
\centering
\caption{Convention gap by the number of hint actions that positively touched the played card (values plotted in main-text Figure~\ref{fig:hintcount_bars}, with the per-partner rows added). Mean $\hat{p}$ and failure in \%, gap in pp, play-weighted. Game CI: 95\% game-cluster bootstrap interval of the gap; participant CI: participant-cluster interval (human-AI only; 10{,}000 draws). Gap/$\hat{p}$ is the gap divided by mean $\hat{p}$, its ceiling. Human-AI rows are human plays.}
\label{tab:hintcount}
\footnotesize
\setlength{\tabcolsep}{4pt}
\begin{tabular}{llrrrrllr}
\toprule
Setting & Positive touches & $N$ & Mean $\hat{p}$ & Failure & Gap & Game CI & Participant CI & Gap/$\hat{p}$ \\
\midrule
Human-human & 0 & 484 & 68.5 & 22.3 & +46.2 & [+41.0, +50.6] & -- & 0.67 \\
Human-human & 1 & 5,884 & 41.1 & 5.5 & +35.6 & [+34.6, +36.5] & -- & 0.87 \\
Human-human & 2+ & 2,649 & 3.3 & 1.7 & +1.6 & [+1.2, +2.0] & -- & 0.49 \\
\addlinespace
Human-AI (all partners) & 0 & 778 & 72.2 & 84.3 & $-$12.2 & [$-$14.8, $-$9.5] & [$-$13.6, $-$5.1] & $-$0.17 \\
Human-AI (all partners) & 1 & 10,500 & 48.9 & 24.2 & +24.7 & [+23.8, +25.5] & [+23.9, +25.8] & 0.50 \\
Human-AI (all partners) & 2+ & 4,194 & 8.3 & 7.4 & +0.9 & [+0.5, +1.4] & [+0.2, +1.4] & 0.11 \\
\addlinespace
Human-AI: Full & 0 & 220 & 72.2 & 77.3 & $-$5.0 & [$-$10.4, +0.6] & [$-$7.6, +7.3] & $-$0.07 \\
Human-AI: Full & 1 & 3,542 & 46.2 & 13.9 & +32.3 & [+31.1, +33.4] & [+30.8, +33.1] & 0.70 \\
Human-AI: Full & 2+ & 1,007 & 4.2 & 2.3 & +1.9 & [+1.4, +2.5] & [+1.1, +2.7] & 0.46 \\
\addlinespace
Human-AI: Intentional & 0 & 390 & 73.0 & 87.4 & $-$14.5 & [$-$17.9, $-$11.0] & [$-$16.9, $-$5.9] & $-$0.20 \\
Human-AI: Intentional & 1 & 3,552 & 45.2 & 14.8 & +30.4 & [+29.3, +31.5] & [+29.3, +31.5] & 0.67 \\
Human-AI: Intentional & 2+ & 1,180 & 4.2 & 3.1 & +1.1 & [+0.7, +1.6] & [+0.7, +1.8] & 0.26 \\
\addlinespace
Human-AI: Outer & 0 & 168 & 70.2 & 86.3 & $-$16.1 & [$-$21.8, $-$10.2] & [$-$24.0, $-$8.1] & $-$0.23 \\
Human-AI: Outer & 1 & 3,406 & 55.5 & 44.7 & +10.8 & [+9.3, +12.2] & [+9.3, +12.6] & 0.19 \\
Human-AI: Outer & 2+ & 2,007 & 12.8 & 12.6 & +0.3 & [$-$0.5, +1.1] & [$-$1.1, +1.2] & 0.02 \\
\addlinespace
AI-AI & 0 & 3,748 & 62.2 & 75.2 & $-$13.0 & [$-$14.4, $-$11.6] & -- & $-$0.21 \\
AI-AI & 1 & 19,424 & 12.5 & 11.9 & +0.6 & [+0.2, +1.0] & -- & 0.05 \\
AI-AI & 2+ & 39,307 & 0.5 & 0.6 & $-$0.08 & [$-$0.14, $-$0.03] & -- & $-$0.15 \\
\bottomrule
\end{tabular}
\end{table}

\paragraph{Zero-touch cards and negative constraints.} A card that no hint has positively touched is not a card without literal constraints. A hint to the same hand that does not touch the card excludes the hinted color or rank for it, and $\hat{p}$ uses that exclusion. Table~\ref{tab:zerotouch_negative} counts the zero-touch plays on which the played card's set of possible colors or possible ranks had fewer than five members. This held for 77.9\% of zero-touch plays in human-human games, 44.6\% of zero-touch human plays in human-AI games and 53.2\% in the AI-AI games. In human-human play the zero-touch plays with a negative constraint failed at 14.9\% against 48.6\% without one, a difference of $-$33.7~pp [$-$45.0, $-$22.3], and their gap was 30.7~pp larger [+18.8, +42.4]. In human-AI play the gap difference was +1.8~pp, with a participant-cluster interval that included zero ([$-$8.4, +5.7]). Among one-touch plays, the attribute that the hint did not name had been narrowed by negative constraints on 26.0\% (human-human), 34.0\% (human-AI, human plays) and 31.3\% (AI-AI) of plays.

\begin{table}[!ht]
\centering
\caption{Plays of cards with no positive touch, split by whether negative constraints had restricted the played card (possible-color or possible-rank set with fewer than five members). Mean $\hat{p}$ and failure in \%, gap in pp, play-weighted, with 95\% bootstrap intervals (game cluster for human-human and AI-AI; participant cluster for human-AI). Participant-cluster intervals are asymmetric because plays are concentrated on two participants (Appendix~\ref{app:sensitivity}).}
\label{tab:zerotouch_negative}
\footnotesize
\begin{tabular}{llrrrl}
\toprule
Setting & Negative constraint & $N$ & Mean $\hat{p}$ & Failure & Gap [95\% CI] \\
\midrule
Human-human & with & 377 & 67.9 & 14.9 & +53.0 [+48.1, +56.9] \\
Human-human & without & 107 & 70.9 & 48.6 & +22.3 [+11.3, +33.1] \\
Human-AI (human plays) & with & 347 & 72.7 & 83.9 & $-$11.1 [$-$13.2, $-$5.1] \\
Human-AI (human plays) & without & 431 & 71.7 & 84.7 & $-$13.0 [$-$14.4, $-$1.4] \\
AI-AI & with & 1,993 & 56.3 & 75.0 & $-$18.7 [$-$20.4, $-$17.0] \\
AI-AI & without & 1,755 & 69.0 & 75.5 & $-$6.5 [$-$8.6, $-$4.5] \\
\bottomrule
\end{tabular}
\end{table}

By partner, the zero-touch human plays with a negative constraint were 96 of 220 with Full (43.6\%), 157 of 390 with Intentional (40.3\%) and 94 of 168 with Outer (56.0\%).

\paragraph{Strata within bins of $\hat{p}$.} The strata differ in mean $\hat{p}$, so a difference in gap between strata could reflect the ceiling alone. Table~\ref{tab:hint_qbin} therefore compares the strata within bins of $\hat{p}$. Within the bins where both the zero-touch and the one-touch strata have support ($\hat{p} > 0.4$), one-touch plays had a larger gap than zero-touch plays in both human corpora. Table~\ref{tab:hint_standardized} gives the strata after reweighting to a common distribution over those bins. The standardized gap was +43.8~pp at zero touches and +54.9~pp at one touch in human-human play (difference $-$11.1~pp [$-$17.2, $-$6.1]), and $-$12.1 and +35.5~pp in human-AI play (difference $-$47.6~pp [$-$52.2, $-$40.1]). Reweighted over the bins that the one-touch and 2+ strata share, the gaps were +25.9 and +22.2~pp in human-human play (difference +3.8~pp [+0.9, +7.0]) and +18.5 and +5.4~pp in human-AI play (difference +13.1~pp [+9.3, +17.4]). The raw one-touch minus 2+ difference in human-human play was +34.0~pp, so the small raw gap at 2+ touches mostly reflects the low ceiling. The 2+ comparison rests on few plays with $\hat{p} > 0$ (137 human-human, 581 human-AI). The $\hat{p} = 0$ bin contributes a gap of zero to both strata. In the top bin the strata are not at equal $\hat{p}$. The within-bin grid was not computed for the AI-AI corpus or for the three partners separately.

\begin{table}[!ht]
\centering
\caption{Hint strata within bins of $\hat{p}$. Each cell gives $N$; failure (\%); gap in pp with its 95\% bootstrap interval (game cluster for human-human; participant cluster for human-AI human plays). Cells with fewer than 30 plays are marked ``$<$30''; ``none'': no plays. At $\hat{p} = 0$ failure and gap are zero by construction.}
\label{tab:hint_qbin}
\footnotesize
\setlength{\tabcolsep}{4pt}
\resizebox{\textwidth}{!}{\begin{tabular}{lllll}
\toprule
Setting & Bin of $\hat{p}$ & Zero-touch & One-touch & 2+ \\
\midrule
Human-human & $=0$ & $<$30 & 1,294; 0; 0 & 2,500; 0; 0 \\
Human-human & $(0, 0.2]$ & none & 420; 1.9; +13.8 [+12.1, +15.1] & $<$30 \\
Human-human & $(0.2, 0.4]$ & $<$30 & 817; 5.6; +26.6 [+24.8, +28.3] & 32; 12.5; +19.2 [+6.9, +29.4] \\
Human-human & $(0.4, 0.6]$ & 53; 17.0; +37.4 [+26.3, +46.8] & 1,598; 6.8; +45.7 [+44.4, +46.9] & 40; 20.0; +31.3 [+18.0, +43.0] \\
Human-human & $(0.6, 0.8]$ & 370; 20.5; +49.3 [+43.6, +54.0] & 1,639; 7.5; +62.4 [+61.0, +63.7] & 31; 3.2; +66.6 [+59.3, +70.8] \\
Human-human & $(0.8, 1]$ & 48; 45.8; +41.0 [+26.0, +55.5] & 116; 32.8; +55.7 [+48.2, +63.4] & 34; 94.1; +4.9 [+0.0, +13.3] \\
\addlinespace
Human-AI & $=0$ & $<$30 & 1,812; 0; 0 & 3,613; 0; 0 \\
Human-AI & $(0, 0.2]$ & $<$30 & 747; 13.8; +1.8 [$-$2.3, +5.4] & 94; 9.6; +5.4 [$-$2.3, +10.1] \\
Human-AI & $(0.2, 0.4]$ & $<$30 & 1,290; 14.0; +18.5 [+16.8, +21.9] & 89; 23.6; +8.7 [$-$1.3, +15.9] \\
Human-AI & $(0.4, 0.6]$ & 61; 68.9; $-$14.0 [$-$27.9, +6.0] & 2,254; 19.4; +33.5 [+31.9, +35.9] & 114; 49.1; +4.1 [$-$10.5, +15.6] \\
Human-AI & $(0.6, 0.8]$ & 555; 85.6; $-$15.3 [$-$20.1, $-$10.1] & 3,224; 23.9; +46.3 [+44.9, +48.5] & 126; 55.6; +15.5 [+7.2, +26.8] \\
Human-AI & $(0.8, 1]$ & 147; 91.2; 0.0 [$-$3.3, +7.1] & 1,173; 89.8; +7.7 [+6.1, +10.2] & 158; 98.7; +0.8 [$-$0.6, +2.7] \\
\bottomrule
\end{tabular}}
\end{table}

\begin{table}[!ht]
\centering
\caption{Hint strata standardized to a common distribution of $\hat{p}$. Each comparison reweights both strata to the pooled distribution of the two strata over the bins of Table~\ref{tab:hint_qbin} in which both have at least 30 plays. For zero versus one touch these are the three bins with $\hat{p} > 0.4$. For one versus 2+ touches they are all bins except $(0, 0.2]$ in human-human play and all bins in human-AI play. ``In support'' is the number of the stratum's plays in those bins (and their share of the stratum). Mean $\hat{p}$ and failure in \%, gap and difference in pp, play-weighted within bins, with 95\% bootstrap intervals (game cluster for human-human; participant cluster for human-AI human plays; 10{,}000 draws). The raw gap is the unstandardized stratum gap of Table~\ref{tab:hintcount}.}
\label{tab:hint_standardized}
\footnotesize
\setlength{\tabcolsep}{4pt}
\resizebox{\textwidth}{!}{\begin{tabular}{lllrrrlrl}
\toprule
Setting & Comparison & Positive touches & In support & Mean $\hat{p}$ & Failure & Standardized gap & Raw gap & Difference \\
\midrule
Human-human & 0 vs.\ 1 & 0 & 471 (97.3\%) & 63.9 & 20.1 & +43.78 [+37.76, +48.72] & +46.2 & $-$11.11 [$-$17.21, $-$6.11] \\
Human-human & 0 vs.\ 1 & 1 & 3,353 (57.0\%) & 63.2 & 8.3 & +54.89 [+53.75, +55.99] & +35.6 & \\
Human-human & 1 vs.\ 2+ & 1 & 5,464 (92.9\%) & 30.0 & 4.1 & +25.92 [+25.07, +26.77] & +35.6 & +3.76 [+0.86, +7.00] \\
Human-human & 1 vs.\ 2+ & 2+ & 2,637 (99.5\%) & 29.9 & 7.8 & +22.17 [+18.85, +25.12] & +1.6 & \\
\addlinespace
Human-AI & 0 vs.\ 1 & 0 & 763 (98.1\%) & 69.2 & 81.4 & $-$12.15 [$-$16.24, $-$4.74] & $-$12.2 & $-$47.60 [$-$52.19, $-$40.08] \\
Human-AI & 0 vs.\ 1 & 1 & 6,651 (63.3\%) & 69.6 & 34.2 & +35.45 [+34.44, +36.87] & +24.7 & \\
Human-AI & 1 vs.\ 2+ & 1 & 10,500 (100\%) & 37.3 & 18.8 & +18.49 [+17.24, +19.43] & +24.7 & +13.12 [+9.33, +17.44] \\
Human-AI & 1 vs.\ 2+ & 2+ & 4,194 (100\%) & 37.7 & 32.3 & +5.37 [+1.46, +8.58] & +0.9 & \\
\bottomrule
\end{tabular}}
\end{table}

\paragraph{Partner comparison at equal $\hat{p}$, hint stratum and deck stage.} Mean $\hat{p}$ on the plays that humans selected was similar across the three AI partners (38.5\%, 37.9\% and 40.6\%), but these are selected plays and their composition differed. We therefore reweighted each partner's human plays by direct standardization to the pooled three-partner distribution over cells of three covariates. The first is the $\hat{p}$ bin (the six bins of Table~\ref{tab:hint_qbin}), and the second the hint stratum (zero-touch, one-touch, 2+). The third is the \emph{deck stage}, the tercile of the number of cards left in the deck (more than 34, 23 to 34, at most 22). The standardization uses the cells in which every partner had at least one play: 43 of the 48 occupied cells, with 99.63\% of the plays. Table~\ref{tab:partner_standardized} gives the standardized gaps, failure rates and contrasts. The contrasts of Full and of Intentional with Outer were +17.6 and +17.1~pp after standardization. The contrast of Full with Intentional was +3.9~pp raw and +0.5~pp after standardization, with an interval that included zero. The standardized failure rates were 15.9\%, 16.5\% and 33.6\% with Full, Intentional and Outer. The adjustment equalizes three observed covariates of selected plays only; other features of the state are not equalized. It is descriptive, not a controlled design.

\begin{table}[!ht]
\centering
\caption{Gap and failure rate of human plays by AI partner, before and after direct standardization (play-weighted; gaps and differences in pp). The standardization is to the pooled distribution over $\hat{p}$ bin $\times$ hint stratum $\times$ deck stage. ``Cells'' is the number of cells in the common support out of the occupied cells, with the share of the pooled plays they contain. The third gap row requires at least 10 plays per partner in a cell. The two rows marked ``two largest removed'' omit the two participants with the most plays (Appendix~\ref{app:sensitivity}), leaving 7{,}271 plays from 226 participants. Brackets are 95\% joint participant-bootstrap intervals (10{,}000 draws of all participants of the row: 228, or 226 with the two largest removed; weights and support re-estimated in every draw). Raw rows are point estimates; their intervals are in Table~\ref{tab:partnercontrast}. The standardized rows use 4{,}764, 5{,}121 and 5{,}529 plays with Full, Intentional and Outer, and the standardized mean $\hat{p}$ was 38.9\%, 39.0\% and 39.0\%. Intervals of the standardized gaps: Full [+22.1, +24.2], Intentional [+21.2, +23.8], Outer [+4.3, +7.0].}
\label{tab:partner_standardized}
\footnotesize
\setlength{\tabcolsep}{3pt}
\resizebox{\textwidth}{!}{\begin{tabular}{llllllll}
\toprule
Estimate & Cells (plays) & Full & Intentional & Outer & Full $-$ Outer & Intentional $-$ Outer & Full $-$ Intentional \\
\midrule
\multicolumn{8}{l}{\emph{Convention gap (pp)}} \\
Raw & -- & +24.14 & +20.25 & +6.20 & +17.94 & +14.05 & +3.89 \\
Standardized & 43/48 (99.63\%) & +23.03 & +22.48 & +5.39 & +17.64 [+16.17, +18.75] & +17.09 [+15.63, +18.01] & +0.55 [$-$0.32, +1.53] \\
Standardized, $\geq$10 plays per cell & 27/48 (96.04\%) & +23.38 & +22.95 & +5.70 & +17.68 [+16.18, +18.92] & +17.26 [+15.93, +18.30] & +0.43 [$-$0.44, +1.30] \\
\addlinespace
Raw, two largest removed & -- & +23.80 & +21.18 & +6.25 & +17.55 & +14.93 & +2.63 \\
Standardized, two largest removed & 39/46 (99.11\%) & +23.26 & +22.66 & +6.08 & +17.18 [+15.32, +19.00] & +16.58 [+14.74, +18.49] & +0.60 [$-$0.51, +1.61] \\
\midrule
\multicolumn{8}{l}{\emph{Failure rate (\%; differences in pp)}} \\
Standardized & 43/48 (99.63\%) & 15.9 [11.9, 19.5] & 16.5 [12.3, 20.8] & 33.6 [28.9, 37.8] & $-$17.7 [$-$18.8, $-$16.2] & $-$17.1 [$-$18.0, $-$15.7] & $-$0.7 [$-$1.6, +0.3] \\
\bottomrule
\end{tabular}}
\end{table}
\FloatBarrier
\section{Logistic Regression}
\label{app:logistic}

These are play-level outcome models (failure of the played card regressed on $\hat{p}$); they are fitted in-sample without cluster adjustment and are descriptive (Appendix~\ref{app:stats}). M1 has $\hat{p}$ as its only predictor. M2 adds four context covariates: the number of hint actions that positively touched the played card, the remaining lives, the number of cards in the deck, and the game phase (turn number at most 15, 16 to 35, above 35). M3 adds the acting player's skill level (human-human) or the AI partner (human-AI).

\begin{table}[!ht]
\centering
\caption{Logistic regression of play failure on the literal failure probability $\hat{p}$ (play level, in-sample, no cluster adjustment). M1: $\hat{p}$ only; M2: +context (hint actions on the played card, lives, cards in deck, game phase); M3: +skill of the acting player (human-human) or +partner (human-AI). Human-AI rows: human plays. AI-AI: plays of the AI-AI agents (M3 has no additional term and equals M2).}
\label{tab:logistic}
\small
\begin{tabular}{llrccc}
\toprule
Setting & Model & $N$ & AUC & Pseudo-$R^2$ & $\beta$ ($\hat{p}$) \\
\midrule
Human-human & M1 & 9,017 & 0.815 & 0.194 & 5.31 \\
Human-human & M2 (+context) & 9,017 & 0.833 & 0.219 & 5.03 \\
Human-human & M3 (+skill) & 9,017 & 0.852 & 0.249 & 4.92 \\
AI-AI & M1 & 62,479 & 0.978 & 0.612 & 9.67 \\
AI-AI & M2 (+context) & 62,479 & 0.979 & 0.616 & 8.72 \\
Human-AI & M1 & 15,472 & 0.871 & 0.348 & 6.01 \\
Human-AI & M2 (+context) & 15,472 & 0.886 & 0.371 & 5.80 \\
Human-AI & M3 (+partner) & 15,472 & 0.912 & 0.428 & 6.07 \\
\bottomrule
\end{tabular}
\end{table}

\begin{figure}[!ht]
\centering
\includegraphics[width=0.85\textwidth]{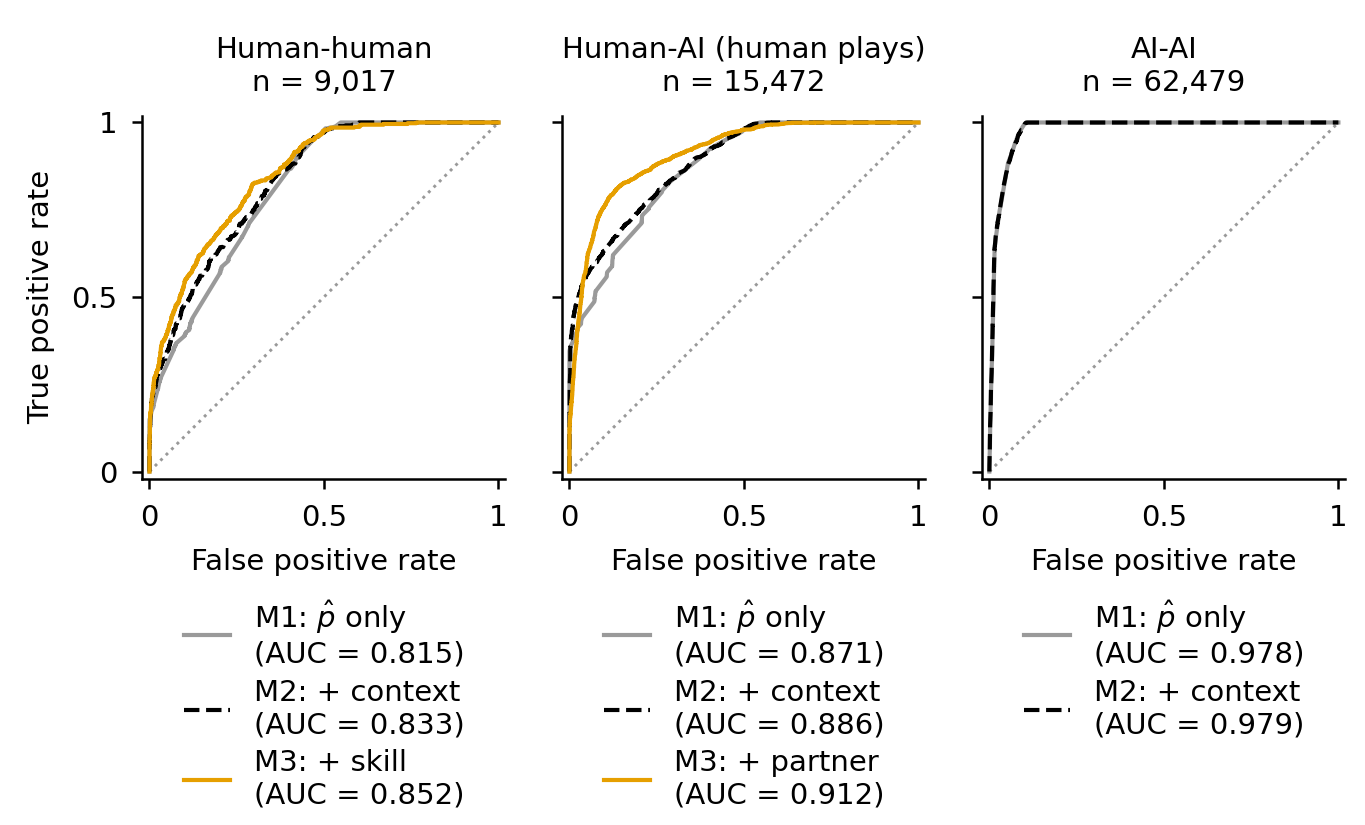}
\caption{In-sample ROC curves of the play-level logistic models of Table~\ref{tab:logistic} (failure of the played card; $N$ = 9{,}017 human-human, 15{,}472 human-AI human plays and 62{,}479 AI-AI plays). For the AI-AI plays M3 has no additional term and equals M2, so it is not drawn.}
\label{fig:roc}
\end{figure}

With $\hat{p}$ as the only predictor, the in-sample AUC was 0.978 for the AI-AI plays ($N$=62{,}479) and 0.815 for human-human plays ($N$=9{,}017) (Table~\ref{tab:logistic}; ROC curves in Figure~\ref{fig:roc}). The coefficient of $\hat{p}$ was 9.67 for the AI-AI agents and 5.31 for human-human plays. By this measure $\hat{p}$ separated failures from successes better on the AI-AI agents' plays than on human plays. The two AUCs are computed over very different distributions of $\hat{p}$, however. In the AI-AI corpus 81.7\% of plays had $\hat{p} = 0$, against 42.1\% of human-human plays (Table~\ref{tab:pinned_zero}). Such plays never fail (Appendix~\ref{app:convgap_by_subject}) and are ranked below every failure, so a large share of them raises the AUC. Restricted to plays with $\hat{p} > 0$, the AUC of $\hat{p}$ was 0.796 for the AI-AI plays (11{,}421 plays), 0.666 for human-human plays (5{,}219) and 0.763 for human plays in human-AI games (10{,}044). On these plays the AUC remained higher for the AI-AI plays than for human-human plays (0.796 against 0.666; no intervals computed). In the human-AI data, adding partner type to the context model raised the AUC from 0.886 (M2) to 0.912 (M3) and the pseudo-$R^2$ from 0.371 to 0.428 ($\Delta R^2 = 0.057$).

\FloatBarrier
\section{Robustness Checks}
\label{app:robustness}

\paragraph{Lives-remaining stratification.}
We stratified plays by the number of life tokens remaining at play time (Table~\ref{tab:lives}). In every setting the play-weighted gap was most positive (least negative) with three lives and lowest on last-life plays. The failure rate rose as lives fell in every setting. The strata are defined before the play, so they are selections of the kind Proposition~\ref{prop:zero} covers. The human-human gap changed little when last-life plays were excluded: +26.6~pp against +26.2~pp on all plays. The corresponding values were +17.4 against +16.4~pp for human plays in human-AI games and $-$0.4 against $-$0.6~pp in the AI-AI corpus. Last-life plays were 6.4\% of human-human plays, 17.9\% of human-AI human plays and 8.4\% of AI-AI plays. The table gives point estimates of pooled values without intervals; it is descriptive.

\begin{table}[!ht]
\centering
\caption{Play-weighted convention gap by life tokens remaining at play time (point estimates). Mean $\hat{p}$ bounds the gap of the stratum (see ``The ceiling of the gap''). Failure rate is the observed failure rate of the plays in the stratum.}
\label{tab:lives}
\small
\begin{tabular}{lrcccc}
\toprule
Setting & Lives & $N$ plays & Mean $\hat{p}$ & Failure rate & Gap \\
\midrule
Human-human & 3 & 6{,}439 & 31.7\% & 4.0\% & +27.7~pp \\
Human-human & 2 & 2{,}004 & 29.8\% & 6.5\% & +23.3~pp \\
Human-human & 1 & 574 & 34.2\% & 15.0\% & +19.2~pp \\
Human-AI (human plays) & 3 & 8{,}577 & 34.4\% & 16.4\% & +17.9~pp \\
Human-AI (human plays) & 2 & 4{,}118 & 44.2\% & 28.0\% & +16.2~pp \\
Human-AI (human plays) & 1 & 2{,}777 & 45.9\% & 34.1\% & +11.8~pp \\
AI-AI & 3 & 47{,}043 & 4.8\% & 5.0\% & $-$0.2~pp \\
AI-AI & 2 & 10{,}162 & 15.8\% & 17.2\% & $-$1.4~pp \\
AI-AI & 1 & 5{,}274 & 21.0\% & 24.1\% & $-$3.2~pp \\
\bottomrule
\end{tabular}
\end{table}

\paragraph{Completed-game filter.}
The corpus keeps only games that reached a terminal state (\S\ref{sec:datasources}), a filter that conditions on how a game ended. To test whether this selection produces the positive gap, we computed the gap on the plays of the truncated human-human games excluded by the filter, under the same two-tier bot exclusion. These 164 games had 1{,}798 plays, mean $\hat{p}$ 35.7\%, failure rate 7.2\% and gap +28.5~pp (95\% game-cluster bootstrap CI [+26.6, +30.4], 10{,}000 draws), against +26.2~pp on the 425 completed games. Pooling both sets gives +26.6~pp [+25.8, +27.3] over 589 games and 10{,}815 plays.

\paragraph{The ceiling of the gap.}
The convention gap cannot exceed the mean literal failure probability $\bar{\hat{p}}$ of the plays it is computed on; the bound is reached when no play in the group fails. The ratio of the gap to $\bar{\hat{p}}$ therefore shows how close a group's failure rate was to zero, relative to the failure rate the literal reference model assigns to the same plays (Table~\ref{tab:ceiling}). It is a description of the selected plays and not a measure of information. The ratio was 83.2\% for the human-human corpus, 92.1\% for expert-expert pairs and 63.7\% for beginner-beginner pairs (descriptive; Appendix~\ref{app:skill}). For human plays in human-AI games it was 62.7\% with Full, 53.5\% with Intentional and 15.3\% with Outer. In the AI-AI corpus it was $-$8.0\%; the mean $\hat{p}$ of 8.0\% there bounds the AI-AI gap on these plays at +8.0~pp, whatever their outcomes. Plays with $\hat{p}=0$ contribute exactly zero to the gap (Table~\ref{tab:pinned_zero}). A group with a low $\bar{\hat{p}}$ has little room for a positive gap, so gaps of groups with different $\bar{\hat{p}}$ should be compared with this bound in view.

\begin{table}[!ht]
\centering
\caption{The play-weighted convention gap as a fraction of its ceiling, the mean literal failure probability $\bar{\hat{p}}$ (point estimates; $N$ is the number of plays).}
\label{tab:ceiling}
\small
\begin{tabular}{lrccc}
\toprule
Setting & $N$ & Mean $\hat{p}$ & Gap & Gap / bound \\
\midrule
Human-human (all) & 9{,}017 & 31.4\% & +26.2~pp & 83.2\% \\
Human-human, expert-expert & 3{,}268 & 31.6\% & +29.1~pp & 92.1\% \\
Human-human, beginner-beginner & 1{,}499 & 33.1\% & +21.1~pp & 63.7\% \\
Human-AI (human plays, all) & 15{,}472 & 39.1\% & +16.4~pp & 41.9\% \\
Human-AI, partner Full & 4{,}769 & 38.5\% & +24.1~pp & 62.7\% \\
Human-AI, partner Intentional & 5{,}122 & 37.9\% & +20.2~pp & 53.5\% \\
Human-AI, partner Outer & 5{,}581 & 40.6\% & +6.2~pp & 15.3\% \\
AI-AI (all) & 62{,}479 & 8.0\% & $-$0.6~pp & $-$8.0\% \\
\bottomrule
\end{tabular}
\end{table}

\FloatBarrier
\section{Sensitivity to Weighting, Exclusions and Corpus Definition}
\label{app:sensitivity}

This appendix reports how the gaps of \S\ref{sec:gap_exists} and \S\ref{sec:partner_dependence} change under other weightings, exclusions and corpus definitions. Unless stated otherwise the estimand is the play-weighted (pooled) convention gap of Eq.~\ref{eq:convgap} with $\hat{p}$ of Eq.~\ref{eq:posterior} (the full-hand computation, Appendix~\ref{app:joint}), in percentage points, and intervals are 95\% percentile cluster bootstrap intervals (10{,}000 draws). The only other intervals are the $t$ intervals on per-participant differences (Appendix~\ref{app:within} and the within-participant paragraph below).

\paragraph{Participant concentration in the human-AI corpus.}
The 15{,}472 human plays come from 2{,}035 games and 228 participants. Participation was concentrated. The two largest participants supplied 4{,}354 and 3{,}847 plays, together 53.0\% of the plays and 1{,}106 of the games. The median participant played 2 games (mean 8.9); 106 participants played one game, 83 played at least three and 22 played at least ten. The Kish effective number of participants under play weights, $(\sum_c w_c)^2/\sum_c w_c^2$ with $w_c$ the number of plays of participant $c$, is 6.9. The two largest participants supplied 52.3\%, 51.9\% and 54.6\% of the plays with Full, Intentional and Outer.

The play-weighted gap therefore describes the pooled plays and is not the gap of an average participant. Table~\ref{tab:convgap_agent} (Appendix~\ref{app:convgap_detail}) sets it beside the equal-participant gap, the mean over participants of each participant's own pooled gap, and Table~\ref{tab:sens_participants} gives both after removing the two largest participants. The equal-participant gap and the paired within-participant differences below are means of per-participant ratios. A participant's gap is a ratio whose denominator, the participant's number of plays, depends on earlier outcomes. Its mean over participants therefore need not be exactly zero under (i)--(iii). These estimands are descriptive and lie outside the population statement of Proposition~\ref{prop:zero}, which covers the pooled gap. We use them as a weighting sensitivity, not as a test against zero. The gap was positive with every partner under both estimands, and Full and Intentional were above Outer under both. The equal-participant differences were +14.49~pp [+11.36, +17.59] for Full minus Outer and +14.78~pp [+11.71, +17.77] for Intentional minus Outer. The Full-minus-Intentional difference was +3.89~pp play-weighted and $-$0.29~pp [$-$3.46, +2.83] equal-participant. Paired within the 69 participants who met both, it was $-$3.43~pp [$-$7.76, +0.91] (32 positive, 37 negative; $t$ interval, Table~\ref{tab:within}; Appendix~\ref{app:convgap_detail}). The play-weighted intervals rest on a Kish-effective number of about seven participants (above).

\begin{table}[!ht]
\centering
\caption{Human plays in human-AI games without the two largest participants: play-weighted and equal-participant convention gap (pp) with 95\% participant-bootstrap intervals (participants resampled jointly across partners). The values with all participants are in Table~\ref{tab:convgap_agent}.}
\label{tab:sens_participants}
\small
\begin{tabular}{lrrrcc}
\toprule
Cell & Plays & Games & Participants & Play-weighted gap [95\% CI] & Equal-participant gap [95\% CI] \\
\midrule
All partners & 7{,}271 & 929 & 226 & +16.80 [+15.21, +18.21] & +14.57 [+13.03, +16.07] \\
Full & 2{,}276 & 331 & 127 & +23.80 [+22.14, +25.12] & +19.35 [+16.94, +21.75] \\
Intentional & 2{,}462 & 294 & 129 & +21.18 [+19.28, +22.84] & +19.72 [+17.45, +22.05] \\
Outer & 2{,}533 & 304 & 120 & +6.25 [+4.55, +7.90] & +4.91 [+2.78, +7.11] \\
\bottomrule
\end{tabular}
\end{table}

\paragraph{Leave-heavy-out.}
Table~\ref{tab:sens_leaveout} removes the largest participants one at a time and together. The overall play-weighted gap stayed between +15.9 and +17.1~pp, and the gap with Outer between +5.7 and +6.7~pp. Full and Intentional stayed above Outer in every row. The play-weighted Full-minus-Intentional difference stayed positive (+2.6 to +4.4~pp; Appendix~\ref{app:convgap_detail}). The contrasts with their intervals and test statistics are in Table~\ref{tab:partnercontrast} of Appendix~\ref{app:stats}.

\begin{table}[!ht]
\centering
\caption{Human plays in human-AI games: play-weighted convention gap (pp, point estimates) after removing participants ranked by number of plays.}
\label{tab:sens_leaveout}
\small
\begin{tabular}{lrrrcccc}
\toprule
Removed & Participants & Plays & Games & All & Full & Intentional & Outer \\
\midrule
None & 228 & 15{,}472 & 2{,}035 & +16.38 & +24.14 & +20.25 & +6.20 \\
Largest & 227 & 11{,}118 & 1{,}469 & +15.94 & +23.86 & +19.46 & +5.67 \\
Second largest & 227 & 11{,}625 & 1{,}495 & +17.06 & +24.19 & +21.63 & +6.73 \\
Two largest & 226 & 7{,}271 & 929 & +16.80 & +23.80 & +21.18 & +6.25 \\
Five largest & 223 & 5{,}964 & 753 & +16.13 & +23.28 & +20.50 & +5.85 \\
\bottomrule
\end{tabular}
\end{table}

\paragraph{Exclusion of preset-deck games.}
In the source study a participant's first game used one of five preset decks (recorded seeds 1 to 5). We call these 240 games (2{,}110 human plays) the preset-deck games. A deck fixed in advance is not a uniformly random shuffle. These games therefore do not satisfy condition~(i) of Proposition~\ref{prop:zero} literally, and the proposition does not guarantee a zero gap for them under (ii) and (iii). The exclusion restricts the estimate to games to which (i) applies. We exclude by recorded seed, because seeds 1, 3 and 5 each correspond to two deck orders in the logs. Without these games the human-AI gap was +16.69~pp (+16.38~pp with them), and Full and Intentional stayed above Outer (Table~\ref{tab:sens_exclusions}). Within the excluded block the gap was +14.44~pp, also positive with every partner. The own-play gap of Full without these games is in Appendix~\ref{app:full_mechanism}. The scope of this check is limited: all 228 participants have a game in the block and only 122 have human plays outside it, so the estimate without them describes a smaller set of participants and their later games.

\paragraph{Assignment of the AI type.} The server code released with the source study \citep{eger2017intentional} draws the AI type uniformly at random for every game, including games started with the ``play again'' link. The code uses one of five preset decks (seeds 1 to 5) for a participant's first game and a random seed from 6 to 10{,}000 afterwards. The logs agree with this. The 2{,}040 games divide into 686 with Full, 647 with Intentional and 707 with Outer ($\chi^2$ test of equal shares, $p = 0.26$). Of the 228 participant identifiers, 227 have exactly one preset-deck game and one has 13, and the recorded seeds of the other 1{,}800 games lie between 13 and 9{,}997. Besides the counts given above, 39 participants played two games. One AI type was met by 122 participants, two by 58 and all three by 48. If the AI type is an independent uniform draw for every game, the expected numbers are 121.9, 54.5 and 51.5 (exact expectation for the observed numbers of games), and the central 95\% of 2{,}000 simulations are 116--128, 46--63 and 45--58. The logs are therefore consistent with a uniform random draw of the AI type for every game, as the code specifies, but the play-weighted contrasts remain concentrated on few participants.

\paragraph{Self-hint games.}
Excluding the 4 games with self-directed hints (Appendix~\ref{app:rules}) removed 15 human plays and changed the gap by less than 0.01~pp. Excluding both these games and the preset-deck games gave +16.69~pp over 13{,}356 plays (combined row of Table~\ref{tab:sens_exclusions}). $\hat{p}$ is computed from the replayed state, so the discards at eight information tokens (Appendix~\ref{app:rules}) enter it as recorded.

\begin{table}[!ht]
\centering
\caption{Human plays in human-AI games: play-weighted convention gap (pp) under exclusions. Intervals are 95\% percentile cluster bootstrap intervals (10{,}000 draws) for the all-partner gap with games and with participants as clusters. The all-games row repeats the intervals of Table~\ref{tab:clusterci}. Partner columns are point estimates.}
\label{tab:sens_exclusions}
\footnotesize
\setlength{\tabcolsep}{4pt}
\begin{tabular}{lrrrcccc}
\toprule
Set & Plays & Games & Part. & Gap & CI (games) & CI (participants) & Full / Int. / Outer \\
\midrule
All games & 15{,}472 & 2{,}035 & 228 & +16.38 & [+15.72, +17.06] & [+15.04, +17.76] & +24.14 / +20.25 / +6.20 \\
Without preset-deck games & 13{,}362 & 1{,}795 & 122 & +16.69 & [+15.95, +17.43] & [+15.16, +18.59] & +24.82 / +20.63 / +5.93 \\
Without 4 self-hint games & 15{,}457 & 2{,}031 & 227 & +16.38 & [+15.69, +17.03] & [+15.02, +17.74] & +24.14 / +20.23 / +6.21 \\
Without both & 13{,}356 & 1{,}792 & 122 & +16.69 & [+15.96, +17.43] & [+15.17, +18.60] & +24.82 / +20.63 / +5.95 \\
Only preset-deck games & 2{,}110 & 240 & 228 & +14.44 & [+12.75, +16.10] & [+12.78, +16.03] & +19.35 / +17.89 / +7.78 \\
\bottomrule
\end{tabular}
\end{table}

\paragraph{Within-participant contrasts under the exclusions.}
Table~\ref{tab:within} (Appendix~\ref{app:within}) gives the within-participant contrasts on all games and without the preset-deck games. Here the estimand is different. The unit is the participant, the quantity is the difference between that participant's own pooled gaps with two partners, and every participant with at least the stated number of games in both cells counts once. Excluding the 4 self-hint games as well changes no mean difference in the lower block of Table~\ref{tab:within} by more than 0.1~pp and no interval limit by more than 0.2~pp, and leaves every $n$ and sign count unchanged. Full minus Outer and Intentional minus Outer persisted within participants under both exclusions. With at least three games per cell the difference was positive for 21 of 21 and 18 of 18 participants. Without the preset-deck games the Full-minus-Intentional contrast was $-$1.9~pp with at least one game per cell and +1.5~pp with at least three; both intervals included zero (Appendix~\ref{app:convgap_detail}).

\paragraph{Human-human corpus: rule options and partner status.}
The selection of the hanab.live corpus (two-player ``No Variant'' completed games) does not condition on the platform's optional rule settings. Of the 425 games, 96 carry a non-default option: \texttt{deckPlays} 48, timed 27, \texttt{speedrun} 9, \texttt{allOrNothing} 7, \texttt{emptyClues} 4, \texttt{cardCycle} 1 and \texttt{detrimentalCharacters} 1. The counts sum to 97 because one game carries two options (timed and \texttt{cardCycle}). Four of these options change the rules of play: \texttt{allOrNothing}, \texttt{emptyClues}, \texttt{cardCycle} and \texttt{detrimentalCharacters}. They occur in 13 games (7 + 4 + 1 + 1, no game carrying two of them). The timed and \texttt{speedrun} options concern the clock and the interface and leave the rules unchanged. \texttt{deckPlays} permits a blind play of the bottom card of the deck. A game with a \texttt{deckPlays} blind play cannot be replayed by our engine and is among the 25 unreplayable games of Appendix~\ref{app:bots}. The 48 \texttt{deckPlays} games in the corpus therefore contain no such play and follow the standard rules in every recorded action. Under \texttt{allOrNothing} play can continue past the end of the standard final round; our replay stops at the standard end of the game (Appendix~\ref{app:pipeline}). The 13 rule-changing games and the 48 \texttt{deckPlays} games are disjoint (61 games). Table~\ref{tab:sens_hh} gives the gap without these games, and without every game that carries a non-default option.

The table also gives the gap by the status of the partner account under the bot screen of Appendix~\ref{app:bots}. In 11 games the partner is a verified bot (1 game) or an account flagged by the screening heuristics (10 games), so that only one player's plays were retained. A flag is a screening result and does not establish that the account is a bot. The gap changed by at most 0.8~pp across these definitions, and every interval excluded zero.

\begin{table}[!ht]
\centering
\caption{Human-human corpus: play-weighted convention gap (pp) under alternative corpus definitions. CI: 95\% cluster bootstrap with games or acting players as clusters (10{,}000 draws). ``--'': acting-player interval not computed for the three option-based definitions.}
\label{tab:sens_hh}
\small
\begin{tabular}{lrrrccc}
\toprule
Set & Plays & Games & Players & Gap & CI (games) & CI (acting players) \\
\midrule
All games (main text) & 9{,}017 & 425 & 155 & +26.16 & [+25.32, +27.01] & [+24.54, +27.80] \\
Without 13 rule-changing games & 8{,}730 & 412 & 147 & +26.30 & [+25.42, +27.14] & -- \\
Without those and \texttt{deckPlays} (61 games) & 7{,}594 & 364 & 133 & +25.42 & [+24.50, +26.32] & -- \\
Default options only (without 96 games) & 6{,}903 & 329 & 126 & +25.40 & [+24.46, +26.31] & -- \\
Without the verified-bot-partner game & 9{,}011 & 424 & 155 & +26.18 & [+25.34, +27.02] & [+24.56, +27.81] \\
Without flagged-partner games & 8{,}896 & 415 & 153 & +25.98 & [+25.14, +26.82] & [+24.36, +27.64] \\
Both players unflagged & 8{,}890 & 414 & 153 & +26.00 & [+25.17, +26.83] & [+24.38, +27.65] \\
\bottomrule
\end{tabular}
\end{table}
\FloatBarrier
\section{Convention Gap by AI Partner Type}
\label{app:convgap_detail}

Table~\ref{tab:convgap_agent} and the per-partner rows of Table~\ref{tab:hintcount} (Appendix~\ref{app:convgap_by_subject}) support the per-partner analysis in \S\ref{sec:partner_dependence}. All values except the human-human reference row of Table~\ref{tab:convgap_agent} concern human plays in the HanabiData human-AI games (15{,}472 plays, 2{,}035 games, 228 participants).

\paragraph{Two estimands.} The play-weighted gap pools all plays, so that every play counts once. The equal-participant gap is the mean over participants of each participant's own pooled gap, so that every participant counts once. The two differ because participation was concentrated. Appendix~\ref{app:sensitivity} discusses the two estimands; Table~\ref{tab:convgap_agent} gives both (equal-participant: Full +19.4, Intentional +19.7, Outer +4.9~pp).

\begin{table}[ht]
\centering
\caption{Convention gap of human plays by AI partner. Mean $\hat{p}$, failure and the play-weighted (PW) gap pool all plays of the cell. The equal-participant (EP) gap is the mean of the participants' own gaps. CIs are 95\% percentile bootstrap intervals (10{,}000 resamples) over whole games or whole participants; participants are resampled jointly across partners. The human-human row is given for reference (clusters: games and acting players; EP not computed).}
\label{tab:convgap_agent}
\footnotesize
\setlength{\tabcolsep}{4pt}
\begin{tabular}{lrrccclll}
\toprule
Partner & Plays & Part. & Mean $\hat{p}$ & Failure & PW gap & Game CI & Participant CI & EP gap [participant CI] \\
\midrule
Full & 4{,}769 & 129 & 38.5\% & 14.4\% & +24.1 & [+23.1, +25.2] & [+22.9, +24.9] & +19.4 [+17.1, +21.7] \\
Intentional & 5{,}122 & 131 & 37.9\% & 17.6\% & +20.2 & [+19.2, +21.3] & [+18.0, +22.5] & +19.7 [+17.5, +22.0] \\
Outer & 5{,}581 & 122 & 40.6\% & 34.4\% & +6.2 & [+5.2, +7.2] & [+4.9, +7.5] & +4.9 [+2.8, +7.1] \\
All partners & 15{,}472 & 228 & 39.1\% & 22.7\% & +16.4 & [+15.7, +17.1] & [+15.0, +17.8] & +14.6 [+13.0, +16.1] \\
\midrule
Human-human & 9{,}017 & 155 & 31.4\% & 5.3\% & +26.2 & [+25.3, +27.0] & [+24.5, +27.8] & -- \\
\bottomrule
\end{tabular}
\end{table}

\paragraph{Partner contrasts.} Table~\ref{tab:partnercontrast} (Appendix~\ref{app:stats}) gives the play-weighted partner differences with a joint participant bootstrap. A resampled participant carries the plays with both partners, so the interval allows for the 64 to 69 participants who appear in both cells of a contrast. The play-weighted gap with Full exceeded that with Outer by +17.94~pp (95\% CI [+16.44, +19.13]), and the gap with Intentional exceeded it by +14.05~pp ([+12.50, +16.12]). Both differences kept their size when the largest or the two largest participants were removed (Table~\ref{tab:sens_leaveout}), and the equal-participant differences agreed (+14.5 and +14.8~pp; Appendix~\ref{app:sensitivity}). Because the clustered variance of these two contrasts rests on 6.7 and 3.2 effective participant clusters, Table~\ref{tab:partnercontrast} also gives a leave-one-participant-out jackknife $z$.

\paragraph{Full versus Intentional.} We claim no ordering of Full and Intentional. The play-weighted gap difference was +3.9~pp ([+1.6, +5.9]), and 85\% of its clustered variance came from two participants. The equal-participant difference was $-$0.3~pp ([$-$3.5, +2.8]), and the paired within-participant difference was $-$3.4~pp ($n$=69; Appendix~\ref{app:within}). The sign therefore depends on the weighting. The same holds for the failure rate of the human plays: the play-weighted difference was $-$3.2~pp with a jackknife $z$ of $-$1.4, and the paired equal-participant difference was +0.7~pp. After standardization the gap difference was +0.5~pp with an interval that included zero (Table~\ref{tab:partner_standardized}, Appendix~\ref{app:convgap_by_subject}).

\paragraph{By hint stratum.} The per-partner rows of Table~\ref{tab:hintcount} split the human plays by the number of hint actions that positively touched the played card. The gap of a stratum cannot exceed its mean $\hat{p}$, so the three strata have different ceilings, and mean $\hat{p}$ and failure are given beside each gap. Zero-touch cards are not cards without literal information, because they often carry negative constraints from hints to other cards (Appendix~\ref{app:convgap_by_subject}). The Full versus Outer difference was largest on one-touch plays (+32.3 versus +10.8~pp). On zero-touch plays the gap was negative with Intentional and Outer; with Full the participant-cluster interval included zero. On 2+ plays the gap was at most +1.9~pp with every partner, with mean $\hat{p}$ between 4\% and 13\%.

\paragraph{Risk-standardized partner contrasts.} The AI type was drawn at random for each game (Appendix~\ref{app:sensitivity}), but the plays compared are those the humans selected (Appendix~\ref{app:convgap_by_subject}). Table~\ref{tab:partner_standardized} (Appendix~\ref{app:convgap_by_subject}) gives the gaps and failure rates after direct standardization over $\hat{p}$ bin, hint stratum and deck stage. The differences between Full or Intentional and Outer remained large (+17.6 and +17.1~pp, against +17.9 and +14.0~pp raw).

\paragraph{Plays with $\hat{p} > 0$.} A play with $\hat{p} = 0$ contributes exactly zero to the gap, and about a third of the human plays were of this kind (33.7\% with Full, 36.8\% with Intentional, 34.7\% with Outer). On the plays with $\hat{p} > 0$ the play-weighted gap was +36.4~pp with Full, +32.1~pp with Intentional and +9.5~pp with Outer (Table~\ref{tab:pinned_zero}).

Main-text Table~\ref{tab:partner_main} gives the gap of human plays and of the AI's own plays for each of the three HanabiData agents; Appendix~\ref{app:full_mechanism} details the own-play values.

\section{Within-Participant Partner Analysis}
\label{app:within}

\paragraph{Participant $\times$ partner structure.}
The HanabiData metadata link each of the 2{,}040 analyzed games to a participant and contain 228 unique participant identifiers. The dataset description reports 240 players; 228 of them have a completed game in the released logs, and all counts use these 228. Of these, 53.5\% played exactly one AI type, 25.4\% two and 21.1\% all three (counts in Appendix~\ref{app:sensitivity}). The AI type was drawn at random for every game by the study server (Appendix~\ref{app:sensitivity}). Participation was self-paced and heavy-tailed rather than a balanced within-subject design. The median participant played 2 games in total (maximum 568; 566 among the 2{,}035 games with a human play), and the median participant$\times$partner cell contains 1 game (mean 5.3, maximum 207). Appendix~\ref{app:sensitivity} describes the concentration of plays on two participants.

\paragraph{Equal-participant paired contrasts.}
For each participant who played two AI types, we computed that participant's own gap and failure rate with each partner and took the paired difference (Table~\ref{tab:within}). Every participant has weight one, so this is a different estimand from the play-weighted contrasts of Appendix~\ref{app:convgap_detail} (a descriptive estimand; Appendix~\ref{app:sensitivity}). The difference between Full or Intentional and Outer persisted within participants. Full$-$Outer was positive for 57 of 69 participants (mean +16.0~pp, 95\% CI [+12.2, +19.8]; sign test $p < 0.001$). Intentional$-$Outer was positive for 58 of 64 (mean +18.8~pp, [+14.7, +22.9]; $p < 0.001$). With at least three games per cell, Full$-$Outer was positive for 22 of 22 participants and Intentional$-$Outer for 20 of 21. Among the 48 participants who played all three AI types, gap(Full) $>$ gap(Outer) for 83.3\%, and both Full and Intentional exceeded Outer for 77.1\%.

\paragraph{What pairing does and does not remove.} Pairing removes stable differences between the participants who met each partner. It does not remove learning, order, carry-over, deck or selective-continuation effects, because participation was self-paced and the partner sequence was not a balanced design. The paired contrasts are therefore evidence against a purely between-participant composition effect for the Outer difference; they are not a controlled within-subject experiment.

\paragraph{Full versus Intentional.} The paired difference was $-$3.4~pp with at least one game per cell ($n$=69; 32 positive, 37 negative) and +1.8~pp with at least three ($n$=18; 13 positive, 5 negative); neither interval excluded zero. The play-weighted difference was positive (+3.9~pp; Table~\ref{tab:partnercontrast}); Appendix~\ref{app:convgap_detail} discusses this dependence on the weighting.

\paragraph{Exclusion of preset-deck games.} The preset-deck games are the first games of participants, which used one of five preset decks, recorded as seeds 1 to 5 (240 games, 2{,}110 human plays; Appendix~\ref{app:sensitivity}). The lower block of Table~\ref{tab:within} repeats the paired contrasts without these games. It describes a smaller set of participants and their later games (122 participants; Appendix~\ref{app:sensitivity}). The contrasts with Outer persisted (Full$-$Outer +21.1~pp, 44 of 47 positive; Intentional$-$Outer +18.1~pp, 43 of 48 positive; 21 of 21 and 18 of 18 with at least three games per cell). Full$-$Intentional was $-$1.9 and +1.5~pp at the two thresholds (Appendix~\ref{app:convgap_detail}).

\begin{table}[!ht]
\centering
\caption{Equal-participant paired contrasts of the convention gap ($\Delta$gap) and failure rate ($\Delta$failure) by AI partner, at two minimum-games-per-cell thresholds. Each participant with both cells contributes one difference of that participant's own pooled values. CI: 95\% $t$ interval on the mean paired difference. Sign: participants with positive/negative $\Delta$gap (ties dropped; one tie in the row with $n = 51$); $p$: two-sided exact sign test. Upper two blocks: all games. Lower block: preset-deck games excluded (122 participants have at least one game without a preset deck; $\Delta$failure not computed). Excluding the 4 self-hint games as well changes no mean difference in the lower block by more than 0.1~pp (Appendix~\ref{app:sensitivity}).}
\label{tab:within}
\footnotesize
\begin{tabular}{llrccccc}
\toprule
Threshold & Contrast & $n$ & $\Delta$gap & 95\% CI & Sign $+/-$ & Sign $p$ & $\Delta$failure \\
\midrule
\multicolumn{8}{l}{\emph{All games}} \\
$\geq$1 game & Full $-$ Outer & 69 & +16.0~pp & [+12.2, +19.8] & 57/12 & $3.7\times10^{-8}$ & $-$22.3~pp \\
$\geq$1 game & Intentional $-$ Outer & 64 & +18.8~pp & [+14.7, +22.9] & 58/6 & $9.0\times10^{-12}$ & $-$21.3~pp \\
$\geq$1 game & Full $-$ Intentional & 69 & $-$3.4~pp & [$-$7.8, +0.9] & 32/37 & 0.63 & +0.7~pp \\
\midrule
$\geq$3 games & Full $-$ Outer & 22 & +19.7~pp & [+16.4, +23.1] & 22/0 & $4.8\times10^{-7}$ & $-$22.6~pp \\
$\geq$3 games & Intentional $-$ Outer & 21 & +18.4~pp & [+14.0, +22.7] & 20/1 & $2.1\times10^{-5}$ & $-$22.7~pp \\
$\geq$3 games & Full $-$ Intentional & 18 & +1.8~pp & [$-$2.4, +6.0] & 13/5 & 0.096 & +0.1~pp \\
\midrule
\multicolumn{8}{l}{\emph{Preset-deck games excluded}} \\
$\geq$1 game & Full $-$ Outer & 47 & +21.1~pp & [+16.4, +25.8] & 44/3 & $2.5\times10^{-10}$ & -- \\
$\geq$1 game & Intentional $-$ Outer & 48 & +18.1~pp & [+13.7, +22.6] & 43/5 & $1.4\times10^{-8}$ & -- \\
$\geq$1 game & Full $-$ Intentional & 51 & $-$1.9~pp & [$-$7.6, +3.8] & 26/24 & 0.89 & -- \\
$\geq$3 games & Full $-$ Outer & 21 & +20.8~pp & [+17.4, +24.2] & 21/0 & $9.5\times10^{-7}$ & -- \\
$\geq$3 games & Intentional $-$ Outer & 18 & +18.9~pp & [+14.0, +23.9] & 18/0 & $7.6\times10^{-6}$ & -- \\
$\geq$3 games & Full $-$ Intentional & 18 & +1.5~pp & [$-$3.2, +6.1] & 13/5 & 0.096 & -- \\
\bottomrule
\end{tabular}
\end{table}

\section{Own-Play Statistics of the Three HanabiData Agents}
\label{app:full_mechanism}

Table~\ref{tab:full_mechanism} reports statistics at the moment of each play action for the three HanabiData agents (Full, Intentional, Outer) in the human-AI games. $\hat p$ is the literal failure probability of the played card (\S\ref{sec:posterior}). One-touch and two-or-more-touch (2+) plays are plays of a card that one or at least two hint actions positively touched. The statistics concern the three agents as implemented by \citet{eger2017intentional}; the agents of our AI-AI corpus (Appendix~\ref{app:agents}) are different implementations (Appendix~\ref{app:agents_identity}).

Full played cards with a mean $\hat p$ of 32.3\% and failed on 15.4\% of its plays (917 of 5{,}965), which gives an own-play gap of +16.9~pp play-weighted (95\% participant-cluster CI [+13.5, +18.5]) and +10.1~pp equal-participant ([+7.1, +13.0]). The size of Full's gap thus depends on the weighting; its sign does not. Without the two participants with most plays the play-weighted value was +14.5~pp ([+12.1, +16.6]). On its 3{,}673 plays with $\hat p > 0$, Full's gap was +27.5~pp (mean $\hat p$ 52.4\%, failure 25.0\%; 95\% participant-cluster CI [+21.2, +30.2]). Without the 240 preset-deck games (Appendix~\ref{app:sensitivity}), Full's play-weighted own-play gap was +17.4~pp (5{,}418 plays, 79 participants; [+13.9, +18.9]) and its equal-participant gap +12.5~pp. Full's gap was positive on one-touch plays (4{,}289 plays; mean $\hat p$ 43.3\%, failure 19.6\%, gap +23.8~pp); on 2+ plays it was $-$0.6~pp (1{,}670 plays).

Intentional and Outer played almost only cards with $\hat p = 0$: 4{,}975 of 4{,}979 plays (99.92\%) and 2{,}631 of 2{,}633 plays (99.92\%), against 2{,}292 of 5{,}965 (38.4\%) for Full. Their mean $\hat p$ was 0.068\% and 0.037\%, with 3 and 1 failures, so their own-play gaps were within 0.02~pp of zero. These zeros are pinned: a play with $\hat p = 0$ contributes exactly zero to the gap whatever the partner's hints convey, and only 4 and 2 plays had $\hat p > 0$. They therefore carry no information about what the human partners' hints conveyed.

A positive own-play gap shows that Full's plays failed less often than the literal reference predicts. If Full acts on the transcript alone (condition~(iii), not verified), Corollary~\ref{cor:nonzero} attributes the positive gap to its human partners' choices. This is consistent with the source study's description of Full as an agent that interprets hints as signals of intent. The gap does not by itself identify that mechanism: the comparison involves one implementation per agent and no ablation of the intent module. Full differs from the other two agents mainly in playing cards with $\hat p > 0$.

A larger gap is not better play. Among the three agents, Full had both the largest own-play gap and the highest own-play failure rate (15.4\%, against 0.06\% and 0.04\%).

\begin{table}[!ht]
\centering
\caption{Own-play statistics of each HanabiData agent in the human-AI games (the agent is the acting player; the partner is human). Gaps in pp with 95\% participant-cluster bootstrap CIs; the cluster is the human participant of the game. Play-weighted: pooled over plays. Equal-participant: mean over participants of the gap of the agent's plays in that participant's games. The zeros of Intentional and Outer are pinned (almost all plays at $\hat{p}=0$); the gap on plays at $\hat{p}>0$ is not computed for 4 and 2 plays (---).}
\label{tab:full_mechanism}
\small
\begin{tabular}{lrrr}
\toprule
Metric & Full & Intentional & Outer \\
\midrule
$n$ (play actions) & 5{,}965 & 4{,}979 & 2{,}633 \\
Games / participants & 669 / 122 & 627 / 126 & 617 / 112 \\
Mean $\hat{p}$ & 32.30\% & 0.068\% & 0.037\% \\
Plays at $\hat{p}=0$ & 38.4\% & 99.92\% & 99.92\% \\
Plays at $\hat{p}>0$ (count) & 3{,}673 & 4 & 2 \\
Gap on plays at $\hat{p}>0$ & +27.48 [+21.18, +30.21] & --- & --- \\
Plays at $\hat{p}>0.5$ & 30.6\% & 0.1\% & 0.0\% \\
Failure rate (failures) & 15.37\% (917) & 0.060\% (3) & 0.038\% (1) \\
Gap, play-weighted & +16.92 [+13.52, +18.46] & +0.01 [+0.00, +0.02] & 0.00 [$-$0.01, +0.00] \\
Gap, equal-participant & +10.06 [+7.12, +12.96] & 0.00 [0.00, 0.00] & 0.00 [0.00, 0.00] \\
Mean hint actions on played card & 1.31 & 2.04 & 1.88 \\
One-touch plays & 71.9\% & 8.3\% & 16.2\% \\
2+ plays & 28.0\% & 91.7\% & 83.7\% \\
\bottomrule
\end{tabular}
\end{table}
\section{Hint Quality Metrics}
\label{app:hintquality}

This appendix defines the hint-level metrics of \S\ref{sec:failure_metrics} and reports them with participant-clustered intervals. The unit is one hint action. The human-AI values are for the hints that the three HanabiData AIs gave to humans ($N = 20{,}088$: Full 6{,}412, Intentional 6{,}551, Outer 7{,}125).

\paragraph{Definitions.} Two playability measures must be distinguished. \emph{Per-hint playability} is the mean over hints of the fraction of touched cards that were playable when the hint was given; each hint has weight one. \emph{Pooled touched-card playability} is the number of playable touched cards divided by the number of touched cards; each hint is weighted by the number of cards it touches. This is the hint playability of \S\ref{sec:failure_metrics}. The two differ because hints that touch more cards count more in the pooled version. The \emph{candidate-elimination fraction} (``candidates eliminated'' in the tables) is $(C_{\mathrm{before}} - C_{\mathrm{after}})/C_{\mathrm{before}}$, averaged over hints, where $C$ counts the candidate identities that the hint history leaves open. We report it in two versions. In the \emph{whole-hand} version $C$ is summed over all cards of the target's hand, so it includes the negative constraints that the hint places on untouched cards. In the \emph{touched-card} version $C$ is summed over the touched cards only. Neither is Shannon information: every candidate counts equally whatever its probability, and the counts are per-card candidate sets, not joint hand assignments. \emph{Next-play failure} is the failure rate of the target's play on the turn immediately after the hint, whichever card was played; it is not conditioned on the played card being one that the hint touched.

\paragraph{Intervals and weighting.} Intervals are 95\% percentile bootstraps over participants (10{,}000 draws); a drawn participant brings their hints from every AI they met, so contrasts between AIs include the covariance from shared participants. Two participants received 50.5\% of the AI hints (5{,}971 and 4{,}165; participant concentration is described in Appendix~\ref{app:sensitivity}). The hint-weighted intervals are therefore dominated by two clusters and may under-cover. Beside each hint-weighted value we give the equal-participant value, in which every participant has weight one. An equal-participant contrast is the mean within-participant difference over participants who met both AIs (69 for Full and Outer, 64 for Intentional and Outer, 69 for Full and Intentional). The hint-weighted and equal-participant values are different estimands.

\begin{table}[!ht]
\centering
\caption{Hint metrics of the three AI hinters in human-AI games (\%). HW: hint-weighted, with 95\% participant-cluster bootstrap interval. EP: equal-participant mean (129, 131 and 122 participants; 126 for Full's next-play failure), with 95\% interval from resampling participants. Next-play failure uses the 3{,}631 / 3{,}676 / 4{,}253 hints followed by a human play.}
\label{tab:hintquality_levels}
\footnotesize
\setlength{\tabcolsep}{4pt}
\begin{tabular}{lllll}
\toprule
Metric & Weight & Full & Intentional & Outer \\
\midrule
Per-hint playability & HW & 61.9 [60.8, 63.6] & 63.9 [62.3, 65.4] & 49.6 [47.5, 50.7] \\
 & EP & 70.7 [67.7, 73.7] & 64.9 [62.6, 67.1] & 48.7 [46.2, 51.3] \\
Pooled touched-card playability & HW & 57.4 [56.2, 60.0] & 59.3 [57.2, 60.7] & 43.5 [41.2, 44.6] \\
 & EP & 68.6 [65.5, 71.8] & 60.6 [58.0, 63.2] & 42.5 [40.1, 45.0] \\
Candidates eliminated, whole hand & HW & 36.5 [36.0, 36.8] & 36.2 [35.9, 36.5] & 39.0 [38.7, 39.3] \\
 & EP & 36.1 [35.5, 36.7] & 36.1 [35.6, 36.6] & 38.8 [38.2, 39.3] \\
Candidates eliminated, touched cards & HW & 74.1 [73.5, 74.8] & 74.1 [73.3, 74.7] & 68.4 [67.0, 69.3] \\
 & EP & 75.3 [74.6, 76.0] & 74.2 [73.6, 74.8] & 67.6 [66.6, 68.6] \\
Next-play failure & HW & 7.6 [6.3, 9.0] & 8.3 [5.8, 11.2] & 30.8 [26.7, 35.5] \\
 & EP & 8.2 [5.7, 11.1] & 6.5 [4.9, 8.3] & 29.2 [25.9, 32.7] \\
\bottomrule
\end{tabular}
\end{table}

\begin{table}[!ht]
\centering
\caption{Contrasts between AI hinters (pp). HW: difference of hint-weighted values with 95\% joint participant-cluster bootstrap interval. EP: mean within-participant difference over participants who met both AIs. HW$^{-2}$: hint-weighted after removing the two participants with the most hints. The EP next-play-failure contrast of Full and Outer is over 68 participants.}
\label{tab:hintquality_contrasts}
\footnotesize
\setlength{\tabcolsep}{4pt}
\begin{tabular}{lllll}
\toprule
Metric & Weight & Full $-$ Outer & Intentional $-$ Outer & Full $-$ Intentional \\
\midrule
Per-hint playability & HW & +12.3 [+10.7, +15.5] & +14.2 [+12.2, +16.7] & $-$2.0 [$-$3.6, +0.5] \\
 & EP & +23.8 [+19.1, +28.8] & +17.3 [+13.5, +21.3] & +6.6 [+1.8, +11.3] \\
 & HW$^{-2}$ & +13.9 [+11.1, +17.0] & +14.8 [+12.4, +17.1] & $-$0.9 [$-$3.3, +2.0] \\
Pooled touched-card playability & HW & +13.9 [+12.3, +18.1] & +15.8 [+13.9, +18.0] & $-$1.9 [$-$3.7, +2.0] \\
 & EP & +27.7 [+22.8, +32.6] & +18.3 [+14.2, +22.5] & +7.9 [+2.7, +12.9] \\
 & HW$^{-2}$ & +16.5 [+13.4, +20.0] & +15.9 [+13.5, +18.4] & +0.6 [$-$2.2, +3.9] \\
Candidates eliminated, whole hand & HW & $-$2.6 [$-$3.1, $-$2.0] & $-$2.9 [$-$3.3, $-$2.3] & +0.3 [$-$0.3, +0.6] \\
 & EP & $-$3.0 [$-$4.0, $-$2.1] & $-$3.1 [$-$4.0, $-$2.2] & +0.8 [$-$0.2, +2.0] \\
 & HW$^{-2}$ & $-$2.8 [$-$3.3, $-$2.3] & $-$2.8 [$-$3.3, $-$2.3] & +0.0 [$-$0.6, +0.5] \\
Candidates eliminated, touched cards & HW & +5.7 [+5.1, +7.3] & +5.7 [+5.2, +6.7] & 0.0 [$-$0.4, +0.9] \\
 & EP & +7.4 [+6.0, +8.7] & +6.0 [+4.5, +7.5] & +1.6 [+0.6, +2.5] \\
 & HW$^{-2}$ & +6.8 [+5.8, +7.8] & +6.2 [+5.3, +7.2] & +0.6 [$-$0.1, +1.3] \\
Next-play failure & HW & $-$23.3 [$-$26.7, $-$19.3] & $-$22.5 [$-$24.7, $-$20.0] & $-$0.8 [$-$2.6, +1.6] \\
 & EP & $-$25.1 [$-$29.8, $-$20.6] & $-$25.5 [$-$30.0, $-$21.4] & $-$1.4 [$-$3.9, +0.9] \\
 & HW$^{-2}$ & $-$20.8 [$-$23.2, $-$18.1] & $-$21.6 [$-$24.2, $-$18.8] & +0.8 [$-$0.9, +2.5] \\
\bottomrule
\end{tabular}
\end{table}

\begin{figure}[ht]
\centering
\includegraphics[width=0.95\textwidth]{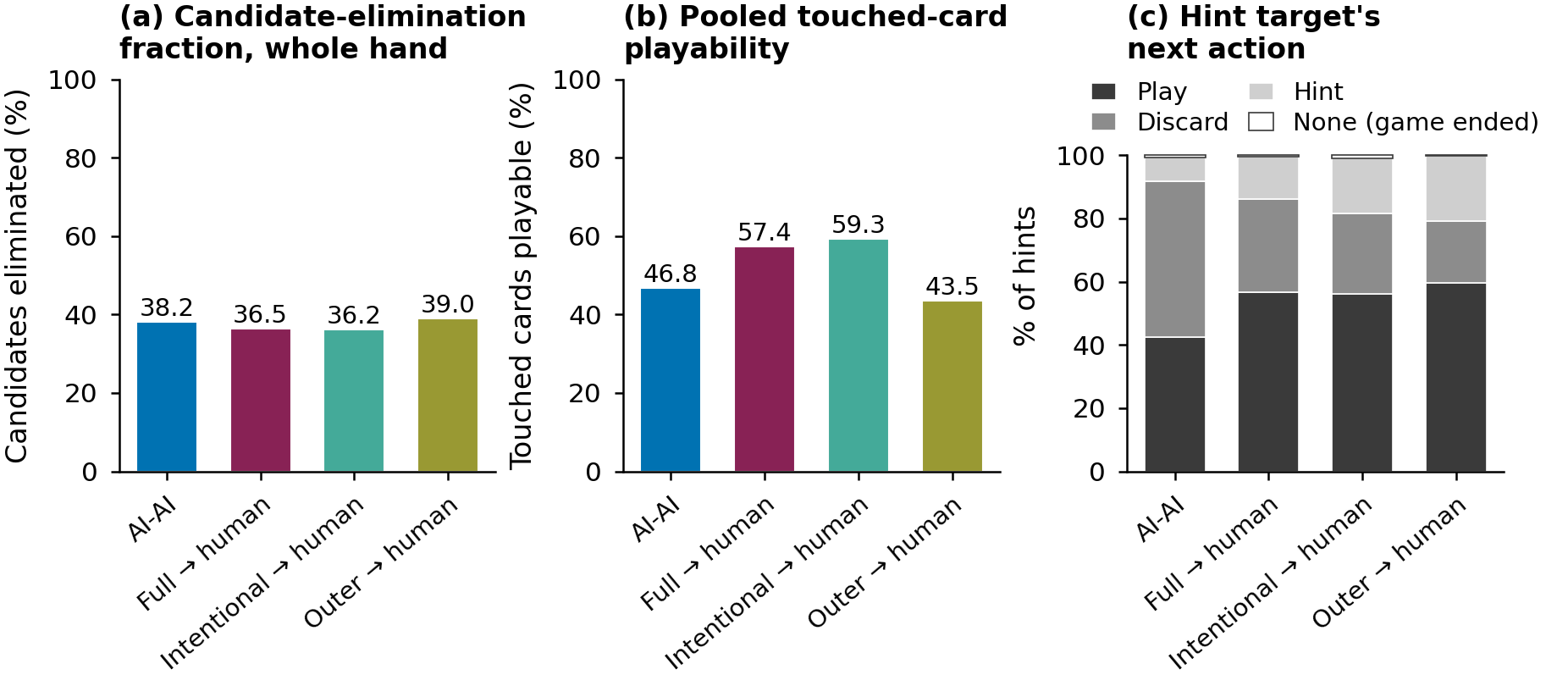}
\caption{Hint metrics of the three HanabiData AIs as hinters to humans ($N = 20{,}088$ hints, hint-weighted), with the pooled hinters of the AI-AI corpus (127{,}063 hints). \textbf{(a)} Whole-hand candidate-elimination fraction. \textbf{(b)} Pooled touched-card playability (hint playability). \textbf{(c)} The hint target's next action; ``None (game ended)'' marks hints after which the game ended before the target acted (0.2--1.0\% of hints). No intervals are drawn; the participant-cluster intervals are in Tables~\ref{tab:hintquality_levels} and~\ref{tab:hintquality_contrasts}.}
\label{fig:hintquality}
\end{figure}

\paragraph{Results.} Outer's hints touched playable cards less often than Full's or Intentional's under every weighting (Tables~\ref{tab:hintquality_levels} and~\ref{tab:hintquality_contrasts}; Figure~\ref{fig:hintquality}). The per-hint difference was +12.3~pp for Full and +14.2~pp for Intentional with hint weights, and +23.8 and +17.3~pp with equal participant weights. Full and Intentional were not separated by the hint-weighted values ($-$2.0~pp [$-$3.6, +0.5]), whereas the within-participant difference favored Full (+6.6~pp [+1.8, +11.3]).

The ordering of the AIs on the candidate-elimination fraction depended on the definition. Counted over the whole target hand, Outer's hints eliminated a larger fraction of candidates (39.0\% against 36.5\% and 36.2\%). Counted over the touched cards only, they eliminated a smaller fraction (68.4\% against 74.1\% and 74.1\%). Outer touched more cards per hint (1.78 against 1.52 and 1.52) and eliminated a smaller fraction of the candidates on each. The share of all eliminated candidates that lay on untouched cards was 35.8\% (Full), 36.2\% (Intentional) and 30.8\% (Outer).

The human play that followed an AI hint failed on 7.6\% of occasions after a hint from Full, 8.3\% after Intentional and 30.8\% after Outer; both contrasts with Outer excluded zero under all three weightings.

\paragraph{Next action after a hint.} A human play followed 56.6\% [53.7, 60.2] of Full's hints, 56.1\% [52.7, 60.1] of Intentional's and 59.7\% [55.7, 64.0] of Outer's. A discard followed 29.6\% [27.5, 31.7], 25.5\% [23.1, 28.7] and 19.5\% [15.7, 24.4], and a hint followed 13.4\% [10.3, 15.0], 17.4\% [13.8, 19.5] and 20.5\% [18.3, 21.6] (hint-weighted, participant-cluster intervals). The remaining hints (0.4\%, 1.0\% and 0.2\%) were followed by the end of the game before the human acted.

\paragraph{Zero-touch plays by partner.} Human plays of cards that no hint had positively touched were 4.6\% [1.5, 8.6] of human plays with Full (220 of 4{,}769), 7.6\% [2.6, 13.7] with Intentional (390 of 5{,}122) and 3.0\% [0.6, 6.1] with Outer (168 of 5{,}581). They failed at 77.3\% [70.6, 85.1], 87.4\% [83.5, 93.7] and 86.3\% [74.3, 91.7] (participant-cluster intervals).

\paragraph{Other hinters.} Table~\ref{tab:hintquality_other} gives the same metrics for human hinters and for the 127{,}063 hints of the AI-AI corpus (Appendix~\ref{app:agents}). The AI-AI rows are a descriptive reference (Appendix~\ref{app:agents_identity}). The touched-card candidate-elimination fraction was computed only for the AI hinters in human-AI games.

\begin{table}[!ht]
\centering
\caption{Hint metrics of other hinters (\%), hint-weighted, with 95\% cluster bootstrap intervals (game cluster for human-human and AI-AI; participant cluster for human hinters in human-AI games). ``Candidates eliminated'' is the whole-hand candidate-elimination fraction. The human-human row uses the 423 of the 425 games that contain at least one hint.}
\label{tab:hintquality_other}
\footnotesize
\setlength{\tabcolsep}{4pt}
\begin{tabular}{lrlll}
\toprule
Hinter & Hints & Per-hint playability & Pooled touched-card playability & Candidates eliminated, whole hand \\
\midrule
Human, human-human corpus & 9,616 & 51.0 [49.7, 52.2] & 44.9 [43.8, 46.1] & 40.7 [40.4, 41.0] \\
Human to Full & 7,334 & 57.2 [52.7, 59.3] & 50.2 [45.8, 52.6] & 40.1 [39.6, 41.2] \\
Human to Intentional & 10,275 & 58.8 [54.7, 60.8] & 51.5 [47.9, 53.6] & 39.6 [39.1, 40.7] \\
Human to Outer & 5,525 & 59.6 [53.1, 63.0] & 52.5 [46.3, 56.2] & 39.5 [39.0, 40.7] \\
\addlinespace
AI-AI, pooled & 127,063 & 53.6 [53.2, 53.9] & 46.8 [46.4, 47.1] & 38.2 [38.1, 38.4] \\
AI-AI: IGGI & 14,915 & 70.4 [69.6, 71.2] & 62.0 [61.3, 62.8] & 37.9 [37.7, 38.1] \\
AI-AI: Piers & 19,217 & 60.8 [60.1, 61.4] & 53.6 [52.9, 54.2] & 37.9 [37.7, 38.1] \\
AI-AI: Internal & 29,380 & 52.9 [52.2, 53.6] & 46.4 [45.7, 47.0] & 33.8 [33.6, 34.0] \\
AI-AI: Van den Bergh & 25,008 & 49.0 [48.3, 49.7] & 40.0 [39.4, 40.6] & 43.0 [42.8, 43.2] \\
AI-AI: Simple & 8,483 & 48.8 [47.5, 50.3] & 46.9 [45.7, 48.3] & 35.4 [35.2, 35.7] \\
AI-AI: Outer & 27,309 & 47.9 [47.2, 48.6] & 42.0 [41.4, 42.7] & 40.1 [39.9, 40.3] \\
AI-AI: Flawed & 2,751 & 31.8 [30.0, 33.9] & 33.4 [31.3, 35.6] & 36.1 [35.6, 36.7] \\
\bottomrule
\end{tabular}
\end{table}

\FloatBarrier
\section{Operational Categories of Failed Plays}
\label{app:failure}

We classified every failed play of a human player by an operational rule based on the number of hint actions that positively touched the played card, the literal failure probability $\hat{p}$ and the remaining lives (Table~\ref{tab:failure}). The table covers human plays only: human-human games and human plays with each of the three HanabiData AIs. This appendix is descriptive. The category names are labels of the rule and do not establish the cause of a failure. A touched card is one that at least one hint action positively touched (one-touch or 2+). The category ``touched card, $\hat{p} > 0.5$'' was the largest in all four columns (39.5\% to 51.4\%). With Outer it comprised 51.4\% of the 1{,}921 failures, and last-life failures a further 30.1\%. The shares are play-level descriptions of failed plays, that is, of plays selected on their outcome; no interval or test is attached.

\begin{table}[ht]
\centering
\caption{Operational categories of failed human plays (\% of the failures in each column; descriptive). H-H: human-human; ``vs'': human plays with that HanabiData AI. Touched: at least one hint action positively touched the played card. Rows follow the order of assignment: failures with one life left are assigned first, so the other rows exclude them. The bins of $\hat{p}$ are $[0.1, 0.5]$ and $[0.01, 0.1)$.}
\label{tab:failure}
\small
\begin{tabular}{lcccc}
\toprule
Category & H-H & vs Full & vs Intent. & vs Outer \\
\midrule
Failure with 1 life left & 18.1 & 29.2 & 18.6 & 30.1 \\
Zero-touch card, $\hat{p}>0.5$ & 17.4 & 12.2 & 24.5 & 3.4 \\
Touched card, $\hat{p}>0.5$ & 39.5 & 45.2 & 46.3 & 51.4 \\
Touched card, $\hat{p}$ in $[0.1, 0.5]$ & 24.4 & 12.7 & 9.9 & 14.9 \\
Touched card, $\hat{p}$ in $[0.01, 0.1)$ & 0.0 & 0.3 & 0.2 & 0.1 \\
Other (zero-touch with $\hat{p}\leq 0.5$, or touched with $\hat{p}<0.01$) & 0.6 & 0.4 & 0.6 & 0.1 \\
\midrule
Total failures & 476 & 686 & 903 & 1{,}921 \\
\bottomrule
\end{tabular}
\end{table}
\section{Game Score Analysis}
\label{app:gamescore}

This appendix gives the score statistics behind \S\ref{sec:score_missed}. The unit of every analysis here is one game, counted once per setting regardless of how many players or agents acted in it. ANOVA model specifications are in Appendix~\ref{app:stats}. Differences between settings are given with cluster-bootstrap intervals that resample participants for the human-AI games and games for the other two settings (10{,}000 draws). The ANOVA, Tukey HSD, Welch and $\chi^2$ tests treat games as independent and are descriptive (Appendix~\ref{app:stats}); we report $p < 0.001$ without smaller bounds.

\paragraph{Scoring conventions.}
Under \emph{strict} scoring, the official rule (Appendix~\ref{app:rules}), a game that ends on the third misplay (a \emph{strike-out}) scores zero. This convention is standard in reinforcement-learning benchmarks \citep{bard2020hanabi,jeon2023behavioral,dizdarevic2025ah2ac2,cohen2025rlhanabi}. Under \emph{partial} scoring, the score is the fireworks total at the end of play, so a strike-out game keeps the total reached before its third misplay. Partial scoring is common in human-subject studies and in some LLM benchmarks \citep{eger2017intentional,sidji2023hidden,liang2019implicit,ramesh2026sparks}. We obtain the partial score of a game as follows. For a game that did not strike out, it is the recorded final score, so the two conventions coincide for every such game. For a strike-out game, it is the fireworks total before the third misplay, read from the last play record (Appendix~\ref{app:pipeline}). We report both conventions.

\paragraph{Strict scores by setting.}
Mean strict scores were 18.5 in human-human (SD 9.6, $n = 425$ games), 10.7 in AI-AI (SD 6.8, $n = 4{,}900$) and 7.2 in human-AI games (SD 8.5, $n = 2{,}040$). The differences, with 95\% cluster-bootstrap intervals, were $+$11.29 [$+$9.18, $+$13.49] for human-human minus human-AI, $+$7.77 [$+$6.85, $+$8.70] for human-human minus AI-AI, and $-$3.52 [$-$5.51, $-$1.70] for human-AI minus AI-AI. A one-way ANOVA gave $F(2, 7362) = 433.29$, and all three Tukey HSD comparisons had $p < 0.001$.

\paragraph{Strict scores within settings.}
Within human-human, by skill pairing (E = expert, I = intermediate, B = beginner), scores ranged from E-E (23.8) and E-I (23.4) to I-B (13.6) and B-B (9.9); the E-B stratum has only three games (mean 16.7). A one-way ANOVA across the six pair types gave $F(5, 419) = 32.72$, $p < 0.001$. Within human-AI, over the 2{,}035 games containing a human play, the mean score was 7.2 with Full, 12.1 with Intentional and 2.9 with Outer ($F(2, 2032) = 243.21$, $p < 0.001$; all pairwise Tukey $p < 0.001$). Score thus separated Intentional from Full. The convention gap and the hint metrics placed Full and Intentional above Outer without a stable ordering between the two (\S\ref{sec:partner_dependence}).

\begin{figure}[!ht]
\centering
\includegraphics[width=\textwidth]{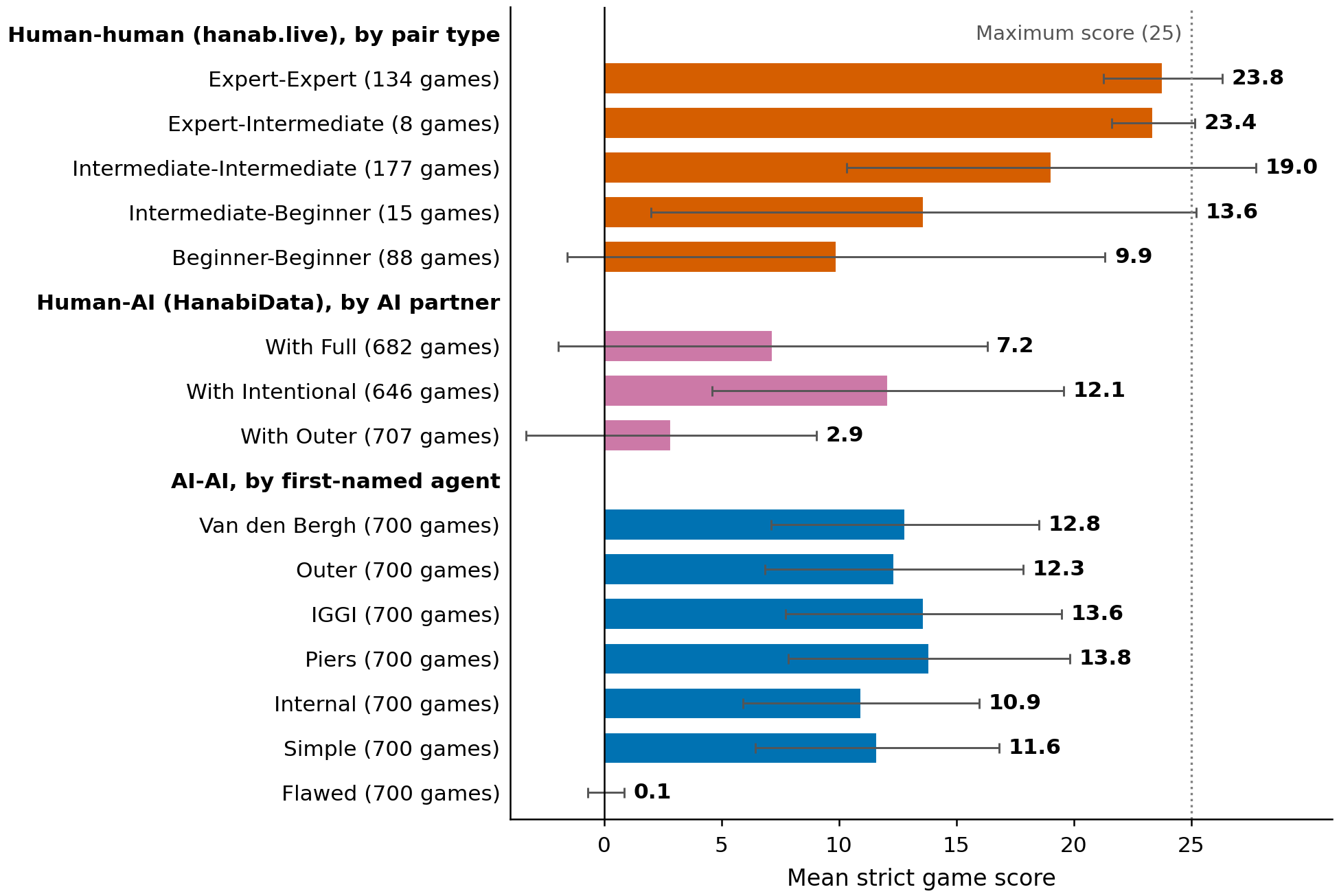}
\caption{Mean strict game score by group. Orange: human-human by skill pairing (the Expert-Beginner type, 3 games, is not shown). Pink: human-AI by AI partner type. Blue: AI-AI by first-named agent (700 games per agent, 100 per partner). Error bars show $\pm$1~SD across games, not a confidence interval; because the strict score is bounded by 0 and 25, a bar can extend beyond the attainable range. Human-AI groups contain the 2{,}035 games with at least one human play (682 / 646 / 707).}
\label{fig:game_scores_all}
\end{figure}

\begin{figure}[!ht]
\centering
\includegraphics[width=\textwidth]{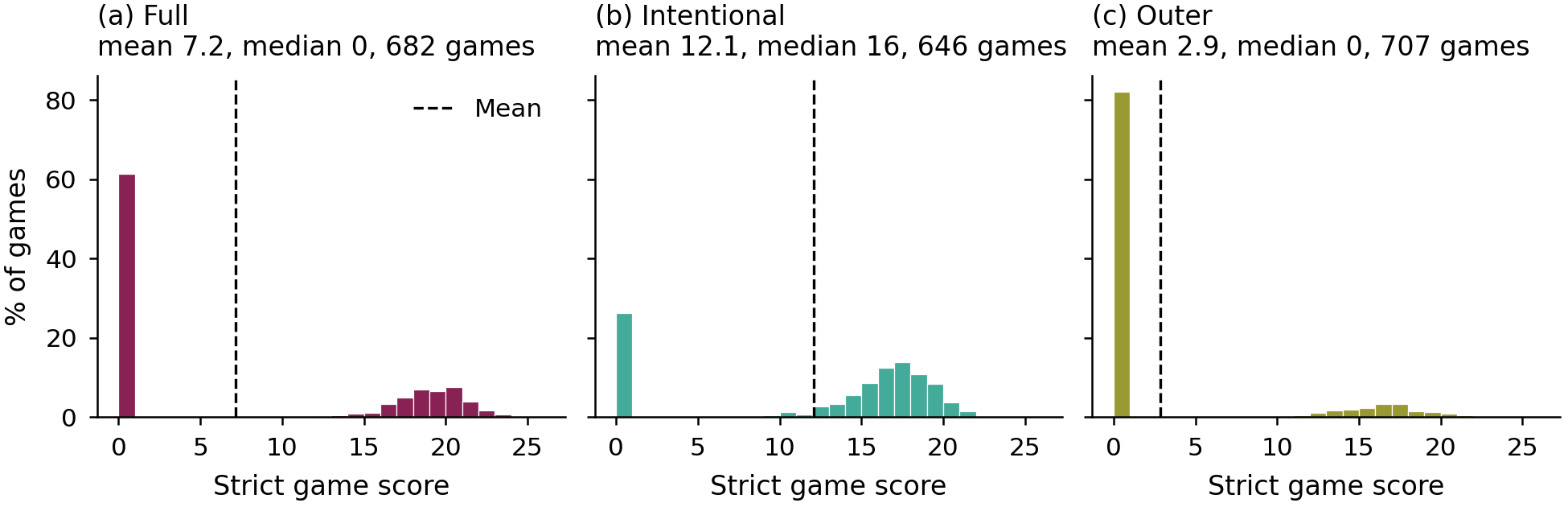}
\caption{Strict game score distributions for human-AI games by AI partner type: \textbf{(a)} Full, \textbf{(b)} Intentional, \textbf{(c)} Outer. Unit: one game; bars are one score point wide, so the vertical axis is the percentage of games at that score. The dashed line marks the mean. Strike-out rates were 61.3\%, 26.2\% and 82.0\%. The panels contain the 2{,}035 games with at least one human play (682 / 646 / 707).}
\label{fig:score_distributions}
\end{figure}

Within AI-AI play, we assigned each of the 4{,}900 games to the first-named agent of its pairing (700 games per agent, 100 per partner; Table~\ref{tab:aiai_scores}). Mean scores over each agent's 700 first-named games ranged from Piers (13.8) and IGGI (13.6) to Simple (11.6) and Internal (10.9), with Flawed near zero (0.07); $F(6, 4893) = 611.19$, $p < 0.001$. In the Tukey HSD comparisons, five pairs were not separated at the 5\% level. They were Van den Bergh and Outer ($p = 0.60$), Van den Bergh and IGGI ($p = 0.056$), IGGI and Piers ($p = 0.98$), Outer and Simple ($p = 0.14$), and Internal and Simple ($p = 0.15$). Every other pair had $p < 0.01$. Mean scores of the groups are shown in Figure~\ref{fig:game_scores_all} (the Expert-Beginner type, 3 games, is not drawn); the score distributions for the three human-AI partners are in Figure~\ref{fig:score_distributions}.

\begin{table}[ht]
\centering
\caption{AI-AI scores by first-named agent (700 games each: 100 per partner, including 100 self-play games). Entries are game means with the SD across games in parentheses. Strike-out is the percentage of games that ended on the third misplay. ``Completed'' is the mean score of games that did not strike out. Self-play is the strict mean over the 100 self-play games.}
\label{tab:aiai_scores}
\small
\begin{tabular}{lccccc}
\toprule
Agent & Strict & Partial & Strike-out (\%) & Completed & Self-play \\
\midrule
Van den Bergh & 12.81 (5.71) & 13.44 (4.49) & 13.4 & 14.79 & 13.99 \\
Outer & 12.33 (5.50) & 12.91 (4.32) & 14.3 & 14.39 & 14.41 \\
IGGI & 13.61 (5.87) & 14.19 (4.67) & 13.4 & 15.72 & 17.04 \\
Piers & 13.83 (6.00) & 14.31 (5.01) & 13.7 & 16.03 & 17.64 \\
Internal & 10.93 (5.03) & 11.48 (4.02) & 14.3 & 12.75 & 10.99 \\
Simple & 11.62 (5.19) & 12.03 (4.38) & 13.6 & 13.45 & 11.41 \\
Flawed & 0.07 (0.78) & 3.21 (2.38) & 99.1 & 8.00 ($n = 6$) & 0.00 \\
\bottomrule
\end{tabular}
\end{table}

\paragraph{Strike-out rate and score of completed games.}
The strict mean combines a strike-out rate with the score of the games that did not strike out. Strike-out rates were 20.5\% in human-human (87 of 425 games), 57.4\% in human-AI (1{,}170 of 2{,}040) and 26.0\% in AI-AI (1{,}273 of 4{,}900); $\chi^2(2) = 667.20$, $p < 0.001$. Among games that did not strike out, mean scores were 23.3 in human-human (SD 2.0, $n = 338$), 16.9 in human-AI (SD 2.6, $n = 870$) and 14.5 in AI-AI (SD 2.8, $n = 3{,}627$). A Welch ANOVA gave $F(2, 878) = 2786.72$, $p < 0.001$, and the three pairwise Welch tests gave $t = 45.4$ (human-human vs.\ human-AI), $t = 73.9$ (human-human vs.\ AI-AI) and $t = 24.4$ (human-AI vs.\ AI-AI), all $p < 0.001$. Human-AI games that did not strike out therefore scored 2.4 points higher than such AI-AI games, whereas the strict mean over all games was 3.5 points lower. This comparison conditions on an outcome (no strike-out), so it describes the selected games and is not a controlled contrast.

By partner, human-AI strike-out rates were 61.3\% with Full (418 of 682 games), 26.2\% with Intentional (169 of 646) and 82.0\% with Outer (580 of 707), and the completed-game means were 18.5, 16.4 and 15.9. In the AI-AI corpus, every strike-out occurred in a game with Flawed in one seat: 1{,}273 of the 1{,}300 games involving Flawed (97.9\%) ended in a strike-out, including all 100 Flawed self-play games.

\paragraph{Partial scores and the distribution diagnostic.}
Under partial scoring the setting means were 20.3 in human-human (SD 6.5), 12.1 in human-AI (SD 5.7) and 11.7 in AI-AI (SD 5.6). The differences, with 95\% cluster-bootstrap intervals, were $+$8.26 [$+$6.81, $+$9.78] for human-human minus human-AI, $+$8.68 [$+$8.04, $+$9.30] for human-human minus AI-AI, and $+$0.41 [$-$0.96, $+$1.65] for human-AI minus AI-AI. Under strict scoring the last difference was $-$3.52 [$-$5.51, $-$1.70] ($d = -0.46$). The human-AI deficit relative to AI-AI was therefore present under strict scoring; under partial scoring, with participants resampled, the interval includes zero. The tests that treat games as independent gave $F(2, 7362) = 461.95$ ($p < 0.001$) and, for the partial human-AI minus AI-AI difference, Tukey $p = 0.015$ and Welch $t = 2.77$ ($p = 0.006$, $d = 0.07$); they ignore the clustering of games within participants. The human-human versus human-AI difference was large under both conventions ($d = 1.35$ partial, $d = 1.24$ strict).

\begin{figure}[!ht]
\centering
\includegraphics[width=\textwidth]{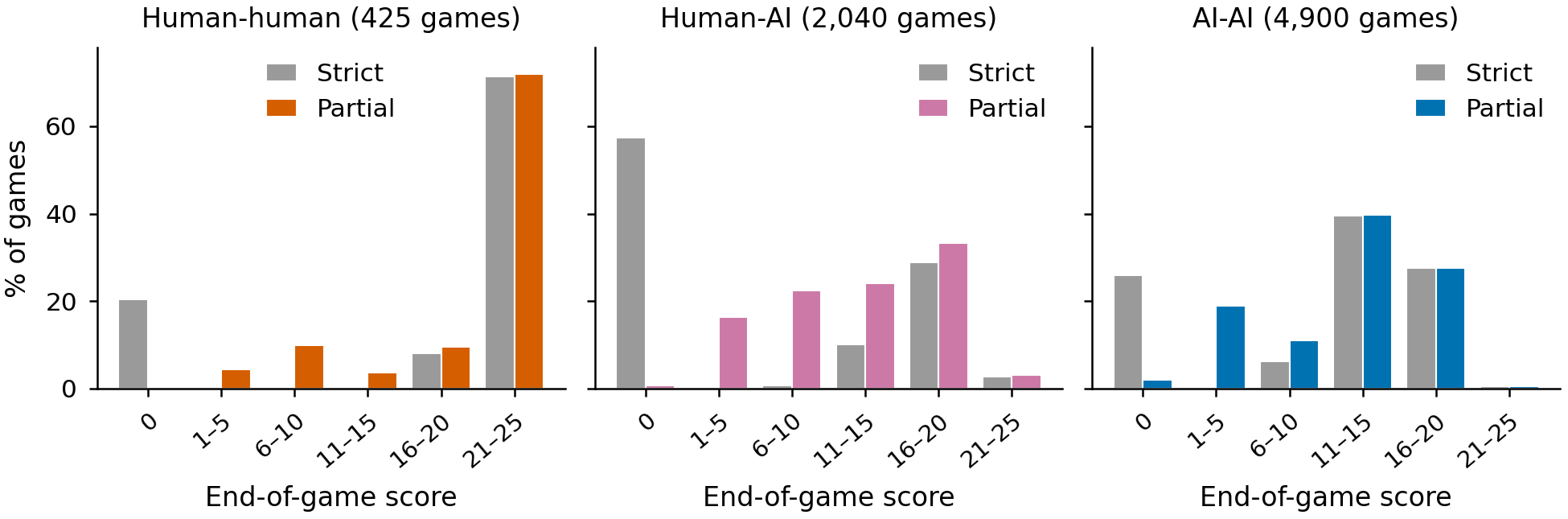}
\caption{Strict (grey) and partial (colored) end-of-game score distributions for the three settings, as percentages of games in the bins $[0,\,1\text{--}5,\,6\text{--}10,\,11\text{--}15,\,16\text{--}20,\,21\text{--}25]$. Unit: one game (425 human-human, 2{,}040 human-AI, 4{,}900 AI-AI). No intervals are shown.}
\label{fig:partial_diag}
\end{figure}

\begin{figure}[!ht]
\centering
\includegraphics[width=\textwidth]{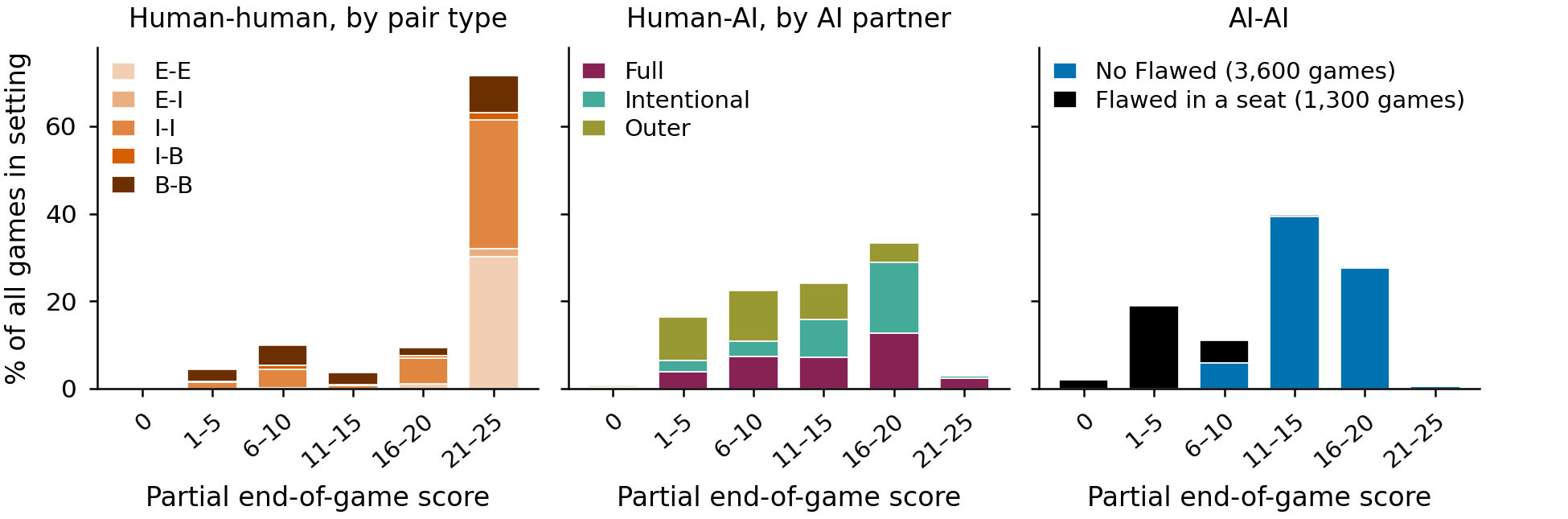}
\caption{Partial-score distributions by group: human-human pair types (E = expert, I = intermediate, B = beginner), human-AI partner types, and AI-AI games with and without Flawed in one seat (1{,}300 and 3{,}600 games). Bars are stacked by group and scaled to the percentage of all games in the setting, so each panel sums to 100\% (the human-human panel omits the three Expert-Beginner games, 0.7\%). The low AI-AI mode consists of games with Flawed in one seat. No intervals are shown.}
\label{fig:partial_group}
\end{figure}

Strict scoring places every strike-out game at zero. Figure~\ref{fig:partial_diag} shows where those games sit under partial scoring, in the bins $[0,\,1\text{--}5,\,6\text{--}10,\,11\text{--}15,\,16\text{--}20,\,21\text{--}25]$. The human-AI distribution had a single broad mode (0.7/16.4/22.4/24.1/33.3/3.0\% across the bins); its strike-out games ended at a mean fireworks total of 8.4. The human-human distribution had a main peak at 21--25 with a lower tail (0.2/4.5/9.9/3.8/9.6/72.0\%); its strike-out games ended at a mean total of 8.9. The AI-AI distribution kept two modes (2.0/19.0/11.1/39.7/27.7/0.6\%). Its strike-out games ended at a mean total of 3.5, and all of them involved Flawed (Figure~\ref{fig:partial_group}). The pooled AI-AI score, like the pooled AI-AI gap, is a mixture over the seven agents of the roster, and the human-AI values are a mixture over three partners (\S\ref{sec:score_missed}). Partial scoring gives credit to games that ended in failure, and strict scoring discards the progress made before the failure. We report both and base no claim on the choice between them.

\clearpage
\section{Cross-Play Matrices for the AI-AI Corpus}
\label{app:crossmatrix}

\begin{figure}[!ht]
\centering
\includegraphics[width=\textwidth]{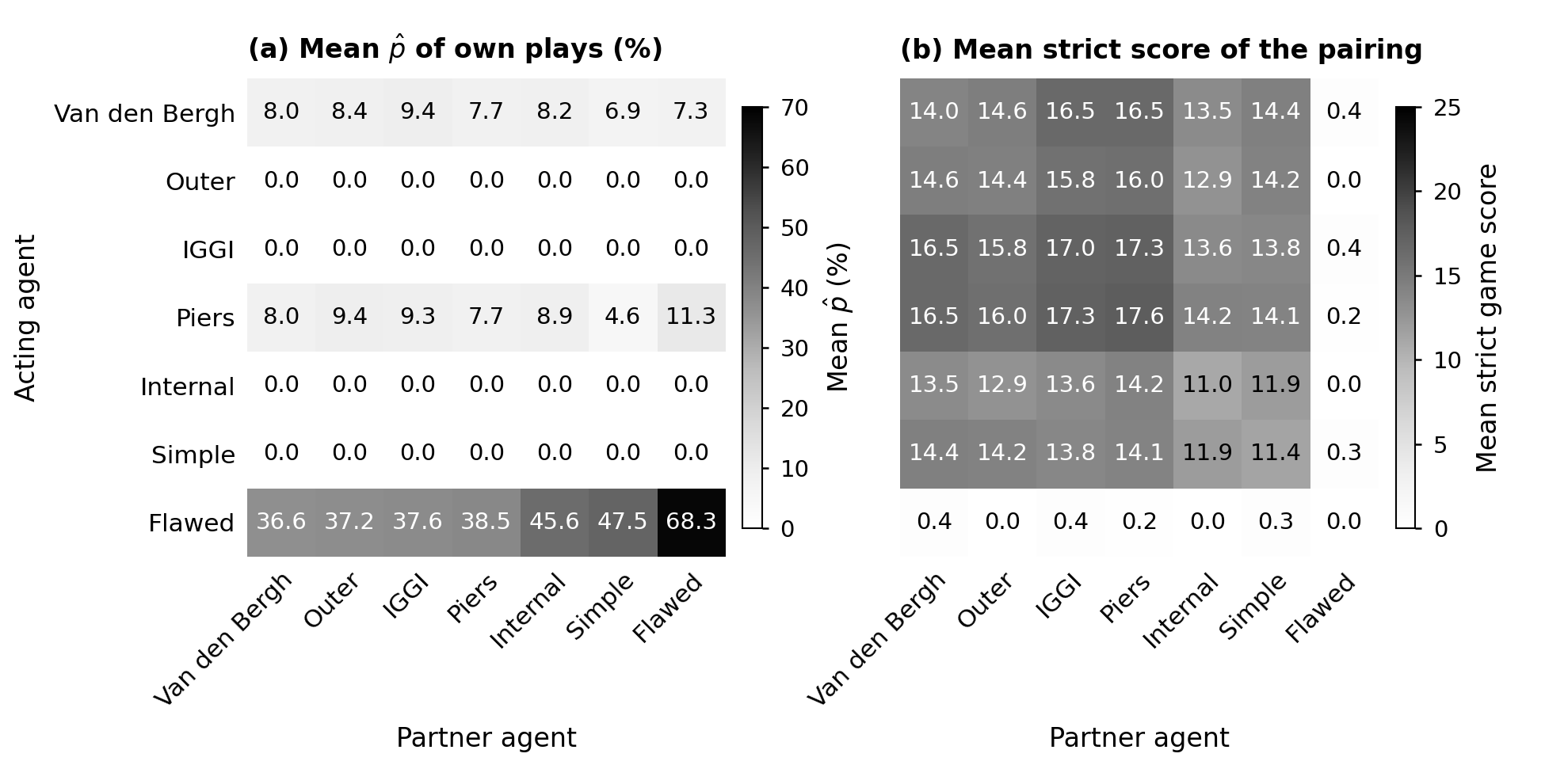}
\caption{Cross-play matrices for the 7$\times$7 pairings of the AI-AI corpus (rows: acting agent; columns: partner agent). \textbf{(a)} Mean literal failure probability $\hat{p}$ (\%) over the acting agent's plays with that partner. \textbf{(b)} Mean strict game score; each cell pools the 200 games of the unordered pairing (100 on the diagonal), so the matrix is symmetric. Cell values are means; no intervals are shown.}
\label{fig:crossplay}
\end{figure}

Figure~\ref{fig:crossplay} shows acting agent $\times$ partner matrices for the AI-AI corpus: seven agents, 49 ordered pairings, 100 games each (62{,}479 plays; Appendix~\ref{app:agents}). No agent of this corpus is a HanabiData agent (Appendix~\ref{app:agents_identity}), so the matrices are not compared with the human-AI data. This appendix is descriptive.

Panel~(a) gives the mean literal failure probability $\hat{p}$ on the acting agent's own plays. Four agents (Outer, IGGI, Internal, Simple) played only cards with $\hat{p} = 0$, so their rows are 0.0\% with every partner and they lost no life on their own plays. Van den Bergh and Piers accepted moderate risk: pooled over all their plays, mean $\hat{p}$ was 8.2\% and 8.3\%, with cells from 6.9\% to 9.4\% and from 4.6\% to 11.3\%. Flawed, which plays cards at a low playability threshold (Appendix~\ref{app:agents}), had a mean $\hat{p}$ of 36.6--38.5\% with Van den Bergh, Outer, IGGI and Piers, 45.6\% with Internal, 47.5\% with Simple and 68.3\% in self-play. The observed failure rates of these three agents were 4.8--8.7\% (Van den Bergh), 6.6--10.2\% (Piers) and 42.8--68.5\% (Flawed). The cell gaps were therefore between $-$1.7 and $+$2.6~pp for Van den Bergh and between $-$4.2 and $+$3.6~pp for Piers. For Flawed they were negative in every cell ($-$8.2 to $-$0.2~pp), and exactly zero for the four agents that played only cards with $\hat{p} = 0$. Cells with Flawed as partner contain few plays (24 to 105, against 382 to 2{,}464 in the other off-diagonal cells) because those games end early.

Panel~(b) gives the mean strict game score of each unordered pairing. Pairings among Van den Bergh, Outer, IGGI and Piers scored 14.0 to 17.6, with the maximum for Piers self-play (17.6) and Piers with IGGI (17.3). Pairings with Internal scored 11.0 to 14.2 and pairings with Simple 11.4 to 14.4. Every pairing with Flawed scored at most 0.4. Over all 1{,}300 games in which an agent took part (either seat; Appendix~\ref{app:gamescore} uses the 700 games in which the agent is first-named), the mean scores were 13.4 (Piers), 13.2 (IGGI), 12.8 (Van den Bergh), 12.4 (Outer), 11.4 (Simple), 11.0 (Internal) and 0.2 (Flawed). The agents that never played a card with $\hat{p} > 0$ thus span the score range from Internal to IGGI, so the mean risk taken and the score order the agents differently.

\section{Off-Belief Learning Agents: Convergent Evidence and Its Limits}
\label{app:obl}

\paragraph{What OBL is.}
Off-belief learning \citep{hu2021offbelief} trains a policy that interprets past actions as if they had been taken by a fixed policy $\pi_0$, while assuming that future play follows the policy being learned. At level 1 (OBL1), $\pi_0$ is uniformly random over the legal actions, so in the learner's belief a past action carries almost no information beyond its literal content. At level $k{+}1$, $\pi_0$ is the level-$k$ policy. This is not an ordinary best response to the level-$k$ partner. The hierarchy is designed to start from grounded play and to admit more interpretation of partner actions at each level.

\paragraph{Relation to condition (ii).}
A uniformly random \emph{legal} hinter does not satisfy condition (ii) exactly, because the number of legal hints depends on the receiver's hand. In the null control of Appendix~\ref{app:nullcontrol}, such a partner gave gaps of +0.10 to +0.53~pp across three receivers. The level-1 belief is therefore close to the literal reference but not identical to it. Moreover, the partner in OBL1 self-play is the trained OBL1 policy, not a uniformly random hinter. Neither this training reference nor the deployed partner policy guarantees (ii), so a zero gap at OBL1 is not a theoretical requirement.

\paragraph{Expected pattern.}
OBL's design suggests a small gap at OBL1 and larger gaps at higher levels, but no theorem orders the gap by level. Reward, convention use and $\mathbb{E}[\hat{p}] - \mathbb{E}[y]$ need not move together. A grounded receiver can show a nonzero gap, and the comparison is empirical.

\paragraph{Setup.}
We ran the released checkpoints in two-player games with per-step greedy acting, following the evaluation protocol of \citet{hu2021offbelief}. Self-play used one checkpoint per level (at each level the first released model; at OBL1, the first of the five), with 1{,}000 games per level. OBL1 cross-play used the ten unordered pairings among the five independently trained OBL1 models, with 100 games per pairing. Inter-level play paired each two of the five levels (ten pairings), with 1{,}000 games per pairing. The released set contains a fifth level (OBL5) that the OBL paper does not report \citep{hu2021offbelief}; we include it as a fifth point, and levels 1 to 4 alone show the same order of gaps. Seat 0 always moved first. In cross-play the first-listed model took seat 0. Each inter-level pairing was played twice on the same 1{,}000 deals, once with the lower level in seat 0 and once with the higher level in seat 0.

\paragraph{Verification.}
Every game was replayed through our engine and reproduced the recorded final score (all 26{,}000 games). Mean self-play scores were 21.09, 23.45, 23.97, 24.05 and 24.16 for OBL1--5, and the cross-play score was 20.90. The posterior code is checked against brute-force enumeration in \S\ref{sec:posterior}.

\paragraph{Results by level.}
The gap increased with level (main-text Table~\ref{tab:obl} and Figure~\ref{fig:obl_main}). It was +1.57~pp [+1.33, +1.81] at OBL1 and +12.40, +16.91, +19.88 and +21.68~pp at OBL2--5 (intervals in main-text Table~\ref{tab:obl}). The OBL1 gap was small relative to the higher levels, but its interval excluded zero. The mean $\hat{p}$ of the plays chosen rose from 7.8\% at OBL1 to 24.1\% at OBL5, while the failure rate fell from 6.3\% at OBL1 to 2.2--3.0\% at OBL2--5. Higher levels thus chose plays with a higher $\hat{p}$ and failed less often. In the six inter-level pairings among OBL2--5 (lower level in seat 0), the gap on each agent's own plays ranged from +13.44 to +20.44~pp, and every interval excluded zero (\texttt{output/sensitivity/obl\_per\_game\_bootstrap.md}).

\paragraph{Strata.}
At every level the one-touch stratum had the largest gap (+34.4~pp at OBL5, where mean $\hat{p}$ was 37.3\% and the failure rate 2.9\%; Table~\ref{tab:obl_hints}). The zero-touch strata were small ($n$ = 332--990; +21.0 [+16.8, +25.4]~pp over 394 plays at OBL5). At every level the interval of the one-touch gap lay above that of the zero-touch gap. The one-touch stratum also had the largest gap for human plays with AI partners. In human-human play the raw zero-touch gap was larger (+46.2 against +35.6~pp), and the one-touch gap was the larger of the two only after standardization to a common distribution of $\hat{p}$ (\S\ref{sec:hint_localization}). The small gap on 2+ plays (+0.2 to +1.3~pp in self-play) is mostly a ceiling effect: mean $\hat{p}$ in that stratum was 0.6--1.7\%, and the gap cannot exceed mean $\hat{p}$.

\begin{table}[!ht]
\centering
\caption{OBL convention gap by the number of positively touching hint actions on the played card. First line per condition: gap in pp [95\% CI]. Second line: plays (share of the condition's plays). Zero-touch cards can still carry negative constraints. Intervals: 95\% game-level bootstrap (10{,}000 draws).}
\label{tab:obl_hints}
\small
\begin{tabular}{lccc}
\toprule
Condition & Zero-touch & One-touch & 2+ \\
\midrule
OBL1 self-play & $-$11.7 [$-$14.3, $-$9.1] & +5.0 [+4.5, +5.5] & +0.2 [+0.1, +0.3] \\
 & 990 (4.3\%) & 8{,}951 (39.3\%) & 12{,}850 (56.4\%) \\
OBL2 self-play & +6.5 [+2.4, +10.4] & +22.1 [+21.7, +22.6] & +0.5 [+0.4, +0.6] \\
 & 375 (1.5\%) & 13{,}201 (54.5\%) & 10{,}628 (43.9\%) \\
OBL3 self-play & +16.8 [+12.4, +21.4] & +29.0 [+28.5, +29.4] & +0.7 [+0.6, +0.8] \\
 & 332 (1.4\%) & 13{,}887 (56.6\%) & 10{,}319 (42.1\%) \\
OBL4 self-play & +15.4 [+11.3, +19.5] & +32.8 [+32.3, +33.3] & +1.0 [+0.8, +1.1] \\
 & 390 (1.6\%) & 14{,}524 (58.7\%) & 9{,}817 (39.7\%) \\
OBL5 self-play & +21.0 [+16.8, +25.4] & +34.4 [+33.9, +34.9] & +1.3 [+1.1, +1.5] \\
 & 394 (1.6\%) & 15{,}058 (60.5\%) & 9{,}419 (37.9\%) \\
OBL1 cross-play & $-$15.1 [$-$17.5, $-$12.6] & +4.5 [+4.0, +5.1] & +0.1 [+0.1, +0.2] \\
 & 930 (4.1\%) & 8{,}783 (38.8\%) & 12{,}903 (57.1\%) \\
\bottomrule
\end{tabular}
\end{table}

\paragraph{Standardization.}
The growth is not only a change in the mix of plays. Standardized to the pooled stratum shares of the five levels (2.0\%, 54.2\% and 43.8\%), the gaps were +2.55, +12.35, +16.34, +18.50 and +19.66~pp at OBL1--5 (point values). The one-touch stratum alone rose from +5.00 to +34.42~pp. The monotone order is an empirical finding about these five checkpoints.

\paragraph{OBL1 cross-play.}
Pooled over the ten pairings of the five OBL1 models, the OBL1 gap in cross-play was +1.22~pp [+0.98, +1.45] (self-play +1.57~pp [+1.33, +1.81]), with a mean score of 20.90. This interval also excluded zero. Per pairing (100 games each) the gap ranged from +0.54 to +1.69~pp, and nine of the ten intervals excluded zero. The ten pairings share five trained models and are not ten independent replications.

\paragraph{OBL1 with higher-level partners.}
We held the OBL1 checkpoint fixed and varied the partner's level. With OBL1 in seat 0, the gap on OBL1's own plays was +1.01~pp [+0.68, +1.35] with OBL2, +0.61 [+0.25, +0.96] with OBL3, +0.33 [$-$0.03, +0.69] with OBL4 and +0.12 [$-$0.22, +0.45] with OBL5 (Table~\ref{tab:obl1_partner}). The last two intervals include zero. The aggregate against OBL5 is small because strata of opposite sign offset each other. Against OBL5 the gap was $-$19.48~pp [$-$22.78, $-$16.16] on zero-touch plays, +2.50~pp [+1.56, +3.44] on one-touch plays and +0.30~pp [+0.14, +0.46] on 2+ plays. Each of the three intervals excluded zero. Weighted by their shares of plays (4.4\%, 30.8\% and 64.8\%), the strata contributed $-$0.85, +0.77 and +0.20~pp to the aggregate of +0.12~pp. In OBL1 self-play with OBL1 in seat 0 (Table~\ref{tab:obl1_seat}) the contributions were $-$0.63, +2.00 and +0.13~pp (aggregate +1.49~pp). The one-touch gap was lower against OBL5 than in seat-0 self-play (+2.50 against +5.18~pp [+4.44, +5.89]; the intervals do not overlap), but it remained positive. An aggregate gap near zero therefore does not show that every class of plays is calibrated to the literal reference.

\begin{table}[!ht]
\centering
\caption{Gap on OBL1's own plays by partner and by the number of positively touching hints on the played card. First line per partner: gap in pp [95\% CI]. Second line: plays (share of OBL1's plays with that partner). Intervals: 95\% game-level bootstrap (10{,}000 draws). OBL1 is in seat 0 with OBL2--5 partners; the self-play row pools both seats. Table~\ref{tab:obl1_seat} gives both seats. Against OBL5, mean $\hat{p}$ and failure rate were 58.2\% and 77.6\% (zero-touch), 14.3\% and 11.8\% (one-touch), and 1.0\% and 0.7\% (2+).}
\label{tab:obl1_partner}
\footnotesize
\setlength{\tabcolsep}{4pt}
\begin{tabular}{lcccc}
\toprule
Partner & Zero-touch & One-touch & 2+ & All plays \\
\midrule
OBL1 (self-play) & $-$11.65 [$-$14.26, $-$9.07] & +5.00 [+4.46, +5.50] & +0.20 [+0.10, +0.30] & +1.57 [+1.33, +1.81] \\
 & 990 (4.3\%) & 8{,}951 (39.3\%) & 12{,}850 (56.4\%) & 22{,}791 \\
OBL2 & $-$17.74 [$-$20.49, $-$14.86] & +4.88 [+4.09, +5.68] & +0.42 [+0.26, +0.58] & +1.01 [+0.68, +1.35] \\
 & 645 (5.5\%) & 4{,}169 (35.7\%) & 6{,}862 (58.8\%) & 11{,}676 \\
OBL3 & $-$19.89 [$-$22.76, $-$16.91] & +4.37 [+3.41, +5.32] & +0.29 [+0.14, +0.44] & +0.61 [+0.25, +0.96] \\
 & 577 (5.0\%) & 3{,}746 (32.3\%) & 7{,}274 (62.7\%) & 11{,}597 \\
OBL4 & $-$19.73 [$-$22.77, $-$16.57] & +3.72 [+2.71, +4.68] & +0.21 [+0.04, +0.37] & +0.33 [$-$0.03, +0.69] \\
 & 536 (4.8\%) & 3{,}450 (30.7\%) & 7{,}256 (64.5\%) & 11{,}242 \\
OBL5 & $-$19.48 [$-$22.78, $-$16.16] & +2.50 [+1.56, +3.44] & +0.30 [+0.14, +0.46] & +0.12 [$-$0.22, +0.45] \\
 & 483 (4.4\%) & 3{,}411 (30.8\%) & 7{,}166 (64.8\%) & 11{,}060 \\
\bottomrule
\end{tabular}
\end{table}

\paragraph{Seat.}
Table~\ref{tab:obl1_seat} gives the same comparison for each seat. The decline held in both seats. From self-play to OBL5 the gap on OBL1's own plays fell from +1.49 to +0.12~pp with OBL1 in seat 0 and from +1.65 to $-$0.11~pp in seat 1. The one-touch gap fell from +5.18 to +2.50~pp and from +4.82 to +1.77~pp. In seat 1 the decline was not monotone (0.00~pp with OBL3, +0.27~pp with OBL4). In self-play the zero-touch gap was $-$15.1~pp in seat 0 and $-$8.4~pp in seat 1 (the intervals overlap). Within each seat it moved further from zero with higher-level partners, to between $-$17.7 and $-$19.9~pp in seat 0 and between $-$15.4 and $-$21.3~pp in seat 1.

\begin{table}[!ht]
\centering
\caption{Gap on OBL1's own plays by seat, in pp [95\% CI]. Seat 0 moves first. Self-play rows split OBL1 self-play by the seat of the acting player. Each partner was paired with OBL1 on the same 1{,}000 deals twice, once with OBL1 in seat 0 and once in seat 1. Plays: OBL1's plays. Intervals: 95\% game-level bootstrap (10{,}000 draws).}
\label{tab:obl1_seat}
\footnotesize
\setlength{\tabcolsep}{3pt}
\begin{tabular}{llrcccc}
\toprule
Partner & Seat & Plays & All plays & Zero-touch & One-touch & 2+ \\
\midrule
OBL1 (self-play) & 0 & 11{,}413 & +1.49 [+1.16, +1.82] & $-$15.13 [$-$18.63, $-$11.60] & +5.18 [+4.44, +5.89] & +0.22 [+0.09, +0.36] \\
 & 1 & 11{,}378 & +1.65 [+1.29, +2.00] & $-$8.40 [$-$12.01, $-$4.70] & +4.82 [+4.07, +5.57] & +0.17 [+0.04, +0.30] \\
\addlinespace
OBL2 & 0 & 11{,}676 & +1.01 [+0.68, +1.35] & $-$17.74 [$-$20.49, $-$14.86] & +4.88 [+4.09, +5.68] & +0.42 [+0.26, +0.58] \\
 & 1 & 12{,}097 & +0.86 [+0.52, +1.20] & $-$17.37 [$-$20.21, $-$14.43] & +4.52 [+3.75, +5.30] & +0.36 [+0.22, +0.50] \\
\addlinespace
OBL3 & 0 & 11{,}597 & +0.61 [+0.25, +0.96] & $-$19.89 [$-$22.76, $-$16.91] & +4.37 [+3.41, +5.32] & +0.29 [+0.14, +0.44] \\
 & 1 & 11{,}592 & 0.00 [$-$0.34, +0.34] & $-$21.28 [$-$24.02, $-$18.52] & +2.60 [+1.66, +3.53] & +0.35 [+0.22, +0.49] \\
\addlinespace
OBL4 & 0 & 11{,}242 & +0.33 [$-$0.03, +0.69] & $-$19.73 [$-$22.77, $-$16.57] & +3.72 [+2.71, +4.68] & +0.21 [+0.04, +0.37] \\
 & 1 & 11{,}440 & +0.27 [$-$0.06, +0.59] & $-$15.44 [$-$18.57, $-$12.30] & +2.62 [+1.77, +3.49] & +0.26 [+0.10, +0.43] \\
\addlinespace
OBL5 & 0 & 11{,}060 & +0.12 [$-$0.22, +0.45] & $-$19.48 [$-$22.78, $-$16.16] & +2.50 [+1.56, +3.44] & +0.30 [+0.14, +0.46] \\
 & 1 & 11{,}244 & $-$0.11 [$-$0.44, +0.23] & $-$18.50 [$-$21.63, $-$15.29] & +1.77 [+0.87, +2.63] & +0.23 [+0.06, +0.40] \\
\bottomrule
\end{tabular}
\end{table}

\paragraph{Composition and within-stratum change.}
We split the decline of OBL1's aggregate gap from self-play to each inter-level pairing into two terms by a Kitagawa decomposition \citep{kitagawa1955components}. The first is the change in the shares of the three hint strata; the second is the change of the gap within strata (Table~\ref{tab:obl_kitagawa}). Each decline is computed within one seat, against OBL1's self-play plays in the same seat. The composition term is evaluated at the self-play stratum gaps and the within-stratum term at the inter-level shares. Against OBL2, OBL3, OBL4 and OBL5 the within-stratum term was 28\%, 51\%, 59\% and 70\% of the decline in seat 0 and 62\%, 76\%, 72\% and 79\% in seat 1. With the symmetric weighting, which averages the two orders of evaluation, these shares were 25\%, 52\%, 62\% and 78\% in seat 0 and 58\%, 79\%, 77\% and 86\% in seat 1. This is an accounting identity and assigns no cause. One reading is that OBL1 benefits from the correlation between a lower-level partner's hints and playability, and that this correlation is weaker in the hints of higher-level partners. Changing the partner of a fixed policy does not isolate such a cause.

\begin{table}[!ht]
\centering
\caption{Change of the gap on OBL1's own plays from OBL1 self-play to play with a higher-level partner (pp), split into a composition term (shares of the three hint strata) and a within-stratum term. Each change is computed within one seat: the baseline is OBL1's self-play plays in the same seat.}
\label{tab:obl_kitagawa}
\small
\begin{tabular}{lrrrr}
\toprule
Partner & Total change & Composition & Within stratum & Within share (\%) \\
\midrule
\multicolumn{5}{l}{\textit{OBL1 in seat 0}} \\
OBL2 & $-$0.480 & $-$0.346 & $-$0.133 & 27.8 \\
OBL3 & $-$0.885 & $-$0.430 & $-$0.455 & 51.4 \\
OBL4 & $-$1.157 & $-$0.478 & $-$0.679 & 58.7 \\
OBL5 & $-$1.375 & $-$0.409 & $-$0.965 & 70.2 \\
\multicolumn{5}{l}{\textit{OBL1 in seat 1}} \\
OBL2 & $-$0.784 & $-$0.297 & $-$0.486 & 62.1 \\
OBL3 & $-$1.647 & $-$0.402 & $-$1.244 & 75.6 \\
OBL4 & $-$1.380 & $-$0.387 & $-$0.993 & 72.0 \\
OBL5 & $-$1.753 & $-$0.370 & $-$1.383 & 78.9 \\
\bottomrule
\end{tabular}
\end{table}

\paragraph{What the intervals cover.}
The intervals in this appendix are game-level bootstrap intervals. They reflect the sampling of games (deals and the resulting play). They are conditional on one checkpoint per level and on the five OBL1 models of cross-play, and they do not include training-run or model variability. The difference between two levels is a difference between two trained models, and the ten cross-play pairings share five models.
The OBL result is therefore an observed association across one checkpoint per level.

\paragraph{Availability.}
The OBL code and checkpoints are distributed by their authors under CC-BY-NC~4.0, so the checkpoints and the raw game logs are not redistributed. The public repository holds per-game statistics of all 26{,}000 OBL games (\texttt{data/processed/obl/}). For each game and seat they give the replayed score and the number of plays, the sum of $\hat{p}$ and the number of failures, overall and by hint stratum. The scripts in \texttt{obl/} compute $\hat{p}$ with the joint posterior of \S\ref{sec:posterior}. From the per-game statistics they recompute every estimate and mean score of this appendix, and every interval up to bootstrap Monte Carlo error of at most 0.02~pp (outputs in \texttt{output/sensitivity/obl\_*}). The package also holds JSON summaries of each condition. Generating the games requires the upstream OBL code, the released checkpoints and a GPU. The games with the higher level in seat 0 come from the packaged generation script, and the other games from scripts that differ from it in path handling only. \texttt{obl/README.md} gives the upstream commit and the SHA-256 values of the nine checkpoints. These values were recorded on the generation server after generation and do not show which files were used.

\section{Exact Null Control for the Proposition}
\label{app:nullcontrol}

\paragraph{Purpose.}
Proposition~\ref{prop:zero} states that the population gap is zero under (i)--(iii). We checked this by simulating games in which the three conditions hold by construction of the simulation: the deck is a uniform shuffle (i), the partner's policy satisfies (ii), and the receiver acts on its observation alone (iii). The control uses our two-player engine, which rejects illegal actions, and the posterior code of the paper, with $\hat{p}$ computed on the live simulator state.

\paragraph{Partner policies and why they satisfy (ii).}
A partner policy is the probability law of the partner's action given its view. Both null partners decide the action type and, for a play or a discard, their own slot from a view of the game with every hand removed (public hint knowledge, fireworks, discards, tokens and turn) plus private randomness. A hint names the color or the rank (fair coin) of a uniformly drawn slot of the receiver's hand. The probability of a given hint is therefore $P(\text{hint type} \mid \text{public}) \times |\text{touched slots}| / (2n)$ for a hand of $n$ cards. The receiver's hand enters only through the set of slots that the hint touches, which is the game's response and is part of the transcript. All shuffles in $D_t$ (the shuffles consistent with the receiver's transcript) thus give the recorded actions the same likelihood, which is what (ii) requires. N1 draws the action type at random, with probabilities that depend on the number of information tokens. N2 cycles through the action types by its own turn index.

\paragraph{Receivers.}
Six receivers that act on the transcript alone were used:
\begin{itemize}[leftmargin=*,nosep]
\item the AI-AI agents Flawed, Piers and Van den Bergh (Appendix~\ref{app:agents});
\item last-hinted, which plays the newest card touched by the hint just received;
\item lowest-$\hat{p}$, which plays its lowest-$\hat{p}$ card when $\hat{p} \le 0.35$;
\item risky-random, which plays a random slot on half of its turns.
\end{itemize}
Their mean $\hat{p}$ on plays ranged from 11\% to 71\%, so the cells differ strongly in which plays are selected.

\paragraph{Partners that violate (ii).}
Four kinds of partner violate (ii). V1 draws the hint uniformly from the \emph{legal} hints (which violates (ii) because the number of legal hints depends on the receiver's hand). V2 hints as N1 does (a uniformly drawn slot and attribute), but its play and discard decisions count the cards in the receiver's hand. V3 pairs the AI-AI agents that hint playable cards (Outer, IGGI) with lowest-$\hat{p}$. V4 pairs the same partners with last-hinted. Their cells are the contrast cells.

\paragraph{Estimation.}
Each cell is the pooled gap (mean $\hat{p}$ minus failure rate over the receiver's plays). Intervals are 95\% game-level bootstrap intervals (2{,}000 draws), and $z$ is the gap divided by the bootstrap SE. The prespecified replication and the run of the separately written game loop use the game-cluster linearized SE and a normal interval. The main run is one simulation of every cell at the sizes of Table~\ref{tab:nullcells}; Tables~\ref{tab:nullcells} and~\ref{tab:nullstrata} report it.

\begin{table}[!ht]
\centering
\caption{Exact null control: pooled gap on the receiver's plays in the 12 cells whose partner satisfies (ii) (top), and in the cells whose partner violates (ii) (bottom). Gap = mean $\hat{p}$ minus failure rate, in pp; 95\% game-level bootstrap intervals; $z$ = gap / bootstrap SE.}
\label{tab:nullcells}
\footnotesize
\begin{tabular}{llrrrrlr}
\toprule
Partner & Receiver & Games & Plays & Mean $\hat{p}$ (\%) & Failure (\%) & Gap (pp) [95\% CI] & $z$ \\
\midrule
N1 & Flawed & 40{,}000 & 207{,}220 & 57.64 & 57.56 & +0.08 [$-$0.12, +0.27] & +0.80 \\
N1 & Piers & 40{,}000 & 200{,}228 & 14.39 & 14.39 & $-$0.00 [$-$0.14, +0.13] & $-$0.03 \\
N1 & Van den Bergh & 40{,}000 & 145{,}022 & 12.67 & 12.66 & +0.02 [$-$0.13, +0.16] & +0.21 \\
N1 & last-hinted & 40{,}000 & 167{,}468 & 70.58 & 70.70 & $-$0.12 [$-$0.28, +0.03] & $-$1.49 \\
N1 & lowest-$\hat{p}$ & 20{,}000 & 128{,}053 & 10.89 & 10.98 & $-$0.09 [$-$0.24, +0.06] & $-$1.15 \\
N1 & risky-random & 40{,}000 & 168{,}500 & 70.54 & 70.66 & $-$0.12 [$-$0.32, +0.06] & $-$1.33 \\
N2 & Flawed & 40{,}000 & 200{,}233 & 59.77 & 59.75 & +0.02 [$-$0.18, +0.22] & +0.22 \\
N2 & Piers & 40{,}000 & 203{,}933 & 14.80 & 14.86 & $-$0.06 [$-$0.20, +0.08] & $-$0.86 \\
N2 & Van den Bergh & 40{,}000 & 180{,}046 & 11.66 & 11.67 & $-$0.01 [$-$0.14, +0.12] & $-$0.16 \\
N2 & last-hinted & 40{,}000 & 167{,}837 & 70.65 & 70.62 & +0.03 [$-$0.13, +0.19] & +0.40 \\
N2 & lowest-$\hat{p}$ & 20{,}000 & 122{,}594 & 11.10 & 10.96 & +0.14 [$-$0.01, +0.30] & +1.84 \\
N2 & risky-random & 40{,}000 & 168{,}863 & 70.50 & 70.49 & +0.01 [$-$0.16, +0.21] & +0.09 \\
\midrule
V1 & last-hinted & 40{,}000 & 165{,}164 & 72.20 & 71.76 & +0.44 [+0.28, +0.60] & +5.32 \\
V1 & lowest-$\hat{p}$ & 20{,}000 & 117{,}129 & 11.62 & 11.52 & +0.10 [$-$0.07, +0.27] & +1.16 \\
V1 & risky-random & 40{,}000 & 168{,}640 & 71.14 & 70.61 & +0.53 [+0.35, +0.72] & +5.70 \\
V2 & last-hinted & 40{,}000 & 167{,}885 & 70.56 & 70.51 & +0.04 [$-$0.12, +0.21] & +0.53 \\
V2 & lowest-$\hat{p}$ & 20{,}000 & 127{,}691 & 10.82 & 10.62 & +0.21 [+0.05, +0.35] & +2.66 \\
V2 & risky-random & 40{,}000 & 168{,}812 & 70.54 & 70.48 & +0.05 [$-$0.13, +0.23] & +0.59 \\
V3: Outer & lowest-$\hat{p}$ & 5{,}000 & 71{,}466 & 7.14 & 5.50 & +1.65 [+1.50, +1.79] & +22.15 \\
V3: IGGI & lowest-$\hat{p}$ & 5{,}000 & 49{,}312 & 6.11 & 4.90 & +1.21 [+1.04, +1.38] & +13.85 \\
V4: Outer & last-hinted & 10{,}000 & 59{,}586 & 51.49 & 50.35 & +1.15 [+0.86, +1.44] & +7.61 \\
V4: IGGI & last-hinted & 10{,}000 & 78{,}767 & 46.74 & 31.24 & +15.50 [+15.23, +15.78] & +109.59 \\
\bottomrule
\end{tabular}
\end{table}

\paragraph{Exact-null cells.}
Pooled over the 12 exact-null cells (440{,}000 games, 2{,}059{,}997 plays) the gap was $-$0.010~pp (SE 0.024, $z = -0.39$). The cell gaps ranged from $-$0.12 to +0.14~pp, every cell interval included zero, and the largest $|z|$ was 1.84 (Table~\ref{tab:nullcells}). The sum of $z^2$ over the cells was 10.3 on 12 degrees of freedom (expected value 12). The largest per-cell SE was 0.10~pp, so a cell deviation of 0.3~pp would lie about three SE from zero. Among the plays with $\hat{p} = 0$ or 1 in all 22 cells, none of the 944{,}648 plays with $\hat{p} = 0$ failed and none of the 424{,}316 plays with $\hat{p} = 1$ succeeded. This holds for any partner, because $\hat{p} = 0$ or 1 means that all consistent deals agree.

\paragraph{Strata.}
The proposition also applies to any selection of plays that is measurable in the transcript. We tested the hint-count strata (Table~\ref{tab:nullstrata}) and five bins of $\hat{p}$ ($(0, 0.2]$, $(0.2, 0.4]$, $(0.4, 0.6]$, $(0.6, 0.8]$ and $(0.8, 1)$) in every cell, 74 strata with at least 200 plays in all. Seven had $|z| > 1.96$ (about 3.7 expected by chance), and one had $|z| > 3$ (largest $|z|$ 3.46). For the hint-count strata alone the sum of $z^2$ was 42.5 on 34 degrees of freedom (expected value 34).

\begin{table}[!ht]
\centering
\caption{Exact-null cells: gap in pp ($z$; plays) by the number of positively touching hints on the played card. Values from the main run. last-hinted never plays a zero-touch card.}
\label{tab:nullstrata}
\footnotesize
\begin{tabular}{llccc}
\toprule
Partner & Receiver & Zero-touch & One-touch & 2+ \\
\midrule
N1 & Flawed & +0.05 (+0.40; 122{,}773) & +0.14 (+0.82; 70{,}436) & +0.01 (+0.04; 14{,}011) \\
N1 & Piers & $-$0.37 ($-$1.11; 15{,}138) & +0.02 (+0.24; 114{,}760) & +0.03 (+0.54; 70{,}330) \\
N1 & Van den Bergh & +0.20 (+0.48; 9{,}554) & +0.07 (+0.55; 68{,}303) & $-$0.07 ($-$1.04; 67{,}165) \\
N1 & last-hinted & -- & $-$0.14 ($-$1.61; 149{,}089) & +0.05 (+0.25; 18{,}379) \\
N1 & lowest-$\hat{p}$ & $-$0.28 ($-$0.75; 10{,}065) & $-$0.19 ($-$1.56; 75{,}056) & +0.12 (+1.71; 42{,}932) \\
N1 & risky-random & $-$0.24 ($-$1.57; 85{,}092) & $-$0.02 ($-$0.15; 56{,}836) & +0.01 (+0.04; 26{,}572) \\
N2 & Flawed & $-$0.02 ($-$0.15; 129{,}751) & +0.16 (+0.93; 64{,}348) & $-$0.58 ($-$1.37; 6{,}134) \\
N2 & Piers & +0.29 (+0.87; 15{,}245) & $-$0.12 ($-$1.17; 121{,}217) & $-$0.03 ($-$0.50; 67{,}471) \\
N2 & Van den Bergh & +0.15 (+0.37; 9{,}705) & $-$0.02 ($-$0.19; 80{,}458) & $-$0.02 ($-$0.30; 89{,}883) \\
N2 & last-hinted & -- & +0.02 (+0.18; 152{,}344) & +0.20 (+1.01; 15{,}493) \\
N2 & lowest-$\hat{p}$ & +1.10 (+2.90; 9{,}631) & $-$0.02 ($-$0.21; 74{,}457) & +0.23 (+3.46; 38{,}506) \\
N2 & risky-random & $-$0.04 ($-$0.28; 87{,}056) & +0.06 (+0.42; 57{,}761) & +0.07 (+0.43; 24{,}046) \\
\bottomrule
\end{tabular}
\end{table}

\paragraph{Follow-up runs of the flagged strata.}
The three strata with $|z| > 2.5$ in the main run were simulated again in follow-up runs with independent base seeds. Every follow-up run is reported; none is pooled with the main run or entered in Tables~\ref{tab:nullcells} and~\ref{tab:nullstrata}. Their sizes were chosen while the follow-up runs were in progress (20{,}000 to 40{,}000 completed games per run). One stratum changed sign, and one returned to zero. For N2\,$\times$\,lowest-$\hat{p}$ on 2+ plays the main run gave +0.23~pp ($z = +3.46$) and the pooled follow-up runs $-$0.05~pp [$-$0.09, $-$0.00] ($z = -2.21$). For N1\,$\times$\,risky-random with $\hat{p}$ in $(0.4, 0.6]$ the main run gave $-$1.08~pp ($z = -2.56$) and the follow-up run +0.20~pp [$-$0.60, +1.06], an interval that includes zero.

The zero-touch stratum of N2\,$\times$\,lowest-$\hat{p}$ stayed positive (Table~\ref{tab:nullresidual}). Six follow-up runs (184{,}000 games) gave +0.42~pp [+0.18, +0.67] ($z = +3.38$). A further run of 48{,}000 games with a separately written game loop gave +0.09~pp (SE 0.25). This value is compatible both with zero and with the follow-up runs (difference $z = 1.17$). Pooled by inverse variance, the follow-up runs and this run gave +0.36~pp (SE 0.11, $z = +3.19$). In the follow-up runs the gap of the whole cell was +0.03~pp [$-$0.02, +0.08]. The stratum holds about 0.5 plays per game.

\paragraph{Prespecified replication.}
A replication of this stratum was run with a design recorded in the package before it started (\texttt{null\_replication\_prespec.md}). The design fixed the cell, a size of exactly 200{,}000 games and a base seed used in no other run. It also fixed the primary statistic, the zero-touch gap with a two-sided 95\% normal interval from the game-cluster linearized SE, and the reading of either outcome. For every play, $\hat{p}$ was computed a second time from the deck order and the action log alone. This code rebuilds the hands, the hint constraints, the discard pile and the fireworks, and it evaluates the joint posterior by exact enumeration with integer weights. It does not use the simulator's record of the hint constraints. The two computations agreed exactly on all 1{,}221{,}389 plays. The zero-touch gap was $-$0.09~pp [$-$0.33, +0.16] (96{,}209 plays, $z = -0.70$). The one-touch and 2+ strata gave $-$0.02~pp [$-$0.09, +0.05] and $-$0.02~pp [$-$0.06, +0.02], and the whole cell $-$0.025~pp [$-$0.074, +0.025]. By the reading fixed in advance, the replication did not reproduce the excess.

\paragraph{Diagnostics of the excess.}
The partner policies do not read the receiver's hand (code inspection). Four diagnostics found no error in $\hat{p}$: replay invariance (below, including sets targeted at zero-touch plays), comparison with uniform sampling of $D_t$, exact enumeration of the hidden cards, and calibration of $\hat{p}$ on all slots of the same cell (\emph{Checks of the posterior code}). These diagnostics take $D_t$ from the simulator's record of the hint constraints on each slot. The transcript-based computation of the replication does not, and it agreed with the simulator on every play. The excess of the follow-up runs was not reproduced and remains unexplained. Pooled by inverse variance over all runs except the main run, the stratum gap was +0.16~pp (SE 0.08, $z = +1.90$; Table~\ref{tab:nullresidual}). Chance, together with the adaptive sizes of the follow-up runs, is one reading; no other cause was found.

\begin{table}[!ht]
\centering
\caption{The zero-touch stratum of N2\,$\times$\,lowest-$\hat{p}$: gap in pp by source. The main run is the one in which the stratum was flagged and is excluded from the pooled rows. The replication was specified after the follow-up runs and before it was run; its $\hat{p}$ was recomputed from the action log. The last row is a post hoc pooled summary.}
\label{tab:nullresidual}
\small
\begin{tabular}{lrrlr}
\toprule
Source & Games & Plays in stratum & Gap (pp) & $z$ \\
\midrule
Main run & 20{,}000 & 9{,}631 & +1.10 [+0.37, +1.82] & +2.90 \\
Follow-up runs pooled (six seeds) & 184{,}000 & 88{,}276 & +0.42 [+0.18, +0.67] & +3.38 \\
Separately written game loop$^{\dagger}$ & 48{,}000 & 22{,}846 & +0.09 (SE 0.25) & +0.37 \\
Follow-up and $\dagger$ runs, inverse variance & 232{,}000 & 111{,}122 & +0.36 (SE 0.11) & +3.19 \\
Prespecified replication & 200{,}000 & 96{,}209 & $-$0.09 [$-$0.33, +0.16] & $-$0.70 \\
All runs except the main run, inverse variance & 432{,}000 & 207{,}331 & +0.16 (SE 0.08) & +1.90 \\
\bottomrule
\end{tabular}

\smallskip
{\raggedright\footnotesize $^{\dagger}$This run enters the two inverse-variance rows. The six follow-up runs gave +1.10 (9{,}673 plays), +0.28, +0.26, +0.40, +0.38 and +0.35~pp (20{,}000 to 40{,}000 games each). Two further runs with separate scripts gave +0.50~pp ($n = 4{,}016$) and $-$0.13~pp ($n = 9{,}951$; plays with $0 < \hat{p} < 1$ only) and are not pooled. One follow-up run and the first further run were stopped for computing time (at 24{,}000 of 80{,}000 and 8{,}395 of 20{,}000 planned games); all their completed games are included. The code of the $\dagger$ run and of the two further runs is not in the package; their summaries are reported.\par}
\end{table}

\paragraph{Replay invariance.}
The gap tables test the proposition up to sampling error. A second check tests its premise without sampling error in the outcome. For sampled receiver plays (200 games per cell, up to three plays per game) we drew alternative deals from $D_t$. We kept the visible cards and re-dealt the receiver's hand and the undrawn deck subject to the hint constraints on each slot. We replayed the recorded actions on each alternative deal with the same private randomness. We compared the engine's response to every action, the partner's action type and own slot, the probability of each recorded hint under the partner's policy, the receiver's actions, and $\hat{p}$ of the played slot. Under (ii) and (iii) none of these may differ. In the exact-null cells there were 0 mismatches in 15{,}153 alternative deals; this count includes two sets targeted at zero-touch plays with $0 < \hat{p} < 1$ (2{,}000 games each). In the violating cells 4{,}573 of 10{,}708 alternative deals differed, in the hint probability for V1 and in the partner's choice for V2--V4. The engine's response, the receiver's action and $\hat{p}$ never differed in any cell. The alternative deals were drawn by rejection sampling (5{,}000 uniform re-deals per play), so heavily constrained hands are under-represented among them.

\paragraph{Checks of the posterior code.}
(a) \emph{Uniform sampling of $D_t$.} For zero-touch plays with $0 < \hat{p} < 1$ in the two lowest-$\hat{p}$ null cells (2{,}000 games each), $\hat{p}$ was compared with the fraction of unplayable identities among up to 300 deals drawn uniformly from $D_t$. The mean difference was $-$0.141~pp (SE 0.123; 662 plays) for N1 and $-$0.004~pp (SE 0.127; 628 plays) for N2. The mean squared standardized difference was 0.98 and 0.92 (1 expected). (b) \emph{Exact enumeration.} For every receiver play whose hidden pool (own hand plus undrawn deck) had at most 15 cards, $\hat{p}$ was compared with the fraction of unplayable identities over all assignments of the hidden physical cards that satisfy the hint constraints. The enumeration shares no code with the posterior. In five cells (600 games each; 5{,}239 plays checked; 2{,}536{,}798 assignments) no value differed (maximum absolute difference 0). This check covers late-game states only. (c) \emph{Calibration on all slots.} For N2\,$\times$\,lowest-$\hat{p}$ (8{,}000 games) we compared $\hat{p}$ of every own slot with $0 < \hat{p} < 1$, at every receiver turn, with whether the card was unplayable. Mean $\hat{p}$ minus the unplayable rate was +0.006~pp (SE 0.096; 936{,}608 slot-turns) over all slots, $-$0.006~pp (SE 0.110; 591{,}617) on zero-touch slots, +0.050~pp (SE 0.150) on one-touch slots and $-$0.064~pp (SE 0.288) on 2+ slots.

\paragraph{Results of the contrast cells.}
With partners that violate (ii), the pooled gaps were positive (+0.04 to +15.50~pp), and stratum gaps took either sign (Table~\ref{tab:nullcells}). A uniformly random legal hinter (V1) gave +0.10 to +0.53~pp across three receivers; two of the three intervals excluded zero. Card counting by the partner (V2) gave +0.04 to +0.21~pp; one interval excluded zero. With partners that hint playable cards the gaps were +1.15 to +15.50~pp (four cells: the partners Outer and IGGI with the receivers lowest-$\hat{p}$ and last-hinted). For the receiver that plays the card just hinted, the gap with an IGGI partner was +15.50~pp [+15.23, +15.78] (+17.25~pp [+16.95, +17.56] on one-touch plays). For the lowest-$\hat{p}$ receiver with an Outer partner, the pooled gap was +1.65~pp. It combined $-$34.21~pp [$-$36.88, $-$31.62] on zero-touch plays ($n = 936$), +5.00~pp [+4.70, +5.28] on one-touch plays ($n = 32{,}928$) and $-$0.40~pp [$-$0.46, $-$0.33] on 2+ plays ($n = 37{,}602$). These sizes are specific to these partner and receiver policies.

\paragraph{What the control shows.}
\begin{itemize}[leftmargin=*,nosep]
\item With a partner policy that satisfies (ii), the receivers' plays were compatible with calibration of $\hat{p}$ at the pooled and cell level. The pooled gap and all 12 cell gaps were zero within sampling error for six receivers with very different selection behavior. Across the hint-count and $\hat{p}$-bin strata, 7 of 74 had $|z| > 1.96$ (3.7 expected). Of the three with $|z| > 2.5$, follow-up runs moved one to zero and one to a small gap of opposite sign. The prespecified replication did not reproduce the zero-touch residual of N2\,$\times$\,lowest-$\hat{p}$.
\item It is an end-to-end check of the posterior code (pool counting, positive and negative hint constraints, the falling-factorial joint enumeration) under the conditions of the proposition.
\item The zero does not come from the receiver. Three of the same receivers gave nonzero gaps, or nonzero stratum gaps of opposite sign, when the partner's choices depended on the receiver's hand beyond the slots a hint touches.
\end{itemize}

\paragraph{What it does not show.}
\begin{itemize}[leftmargin=*,nosep]
\item Nothing about OBL (Appendix~\ref{app:obl}).
\item Nothing about human play or the human-AI data, and nothing about why the rule-based agents of our AI-AI corpus have gaps close to zero. Their hints depend on the receiver's hand beyond the slots a hint touches, so they violate (ii); (ii) is sufficient for a zero gap and is not necessary.
\item It does not test violations of (i) (non-uniform shuffles) or of (iii) (receivers with side information).
\item It does not validate the replay pipeline that reconstructs states from logged games, because $\hat{p}$ is computed here on the live simulator state.
\item The null partners are not sensible Hanabi players. Their games are short (three strikes are reached quickly with most receivers), so the plays are mostly early- and mid-game plays, with two players only.
\end{itemize}

\paragraph{Reproduction.}
The module \texttt{src.sensitivity.null\_control} and the replication script reproduce every value in this appendix except those of the runs marked in the note to Table~\ref{tab:nullresidual}. The package also holds the per-game statistics of the main run and of the six follow-up runs (plays, sum of $\hat{p}$ and failures in each hint stratum). From them every cell gap and the zero-touch rows of Table~\ref{tab:nullresidual} can be recomputed without simulation (see the package README).
\section{Change of the Per-Game Gap with Experience}
\label{app:learning}

Model specifications are in Appendix~\ref{app:stats}. We fit Bayesian hierarchical models with random intercepts and random slopes per player (human-human) or per participant (human-AI), using partial pooling to address small per-player sample sizes. Game sequence enters the model in units of ten games, and slopes are reported converted to a per-game value. A fit is reported only if it has no divergent transition, $\hat{R} \leq 1.01$ and bulk and tail effective sample sizes of at least 400.

\textbf{Human-human.} Among 20 hanab.live players with $\geq$10 completed games ($n = 415$ game-player observations), the population slope was positive. Its posterior mean was $\beta = +0.00183$ per game, that is, +0.18~pp per game (95\% highest density interval [+0.006, +0.39]~pp per game; $P(\beta > 0) = 0.98$). The lower end of the interval is close to zero. The fit had no divergent transition, $\hat{R} \leq 1.00$ and effective sample sizes above 1{,}400 (bulk) and 2{,}500 (tail).

\textbf{Human-AI.} Among 23 participant $\times$ AI cells with at least ten games ($n = 1{,}376$ game-participant observations; 10 participants), the fit had 15 divergent transitions and did not meet the reporting criteria above. We therefore report no human-AI slope and draw no conclusion about learning with AI partners from this model.

The per-game gap used here is a mean over a game-dependent number of plays. As noted at the end of the proof of Proposition~\ref{prop:zero}, a single game's gap need not have mean zero, because the number of plays depends on earlier outcomes. This holds even when outcomes follow the literal reference model and no learning occurs. The estimate also uses completed games only, a selection on a later outcome that falls outside the scope of Corollary~\ref{cor:nonzero}. The slope is therefore descriptive.
\section{Own-Play Gap of a Port of Full with LLM Partners}
\label{app:llm_modern}

\paragraph{Question.}
In HanabiData, the gap on Full's own plays with human partners was $+16.9$~pp (play-weighted; Table~\ref{tab:full_mechanism}). Full's design interprets received hints as play signals (\S\ref{sec:discussion}). That observation rests on three rule-based agents from one 2017 study. As a separate observation with a different class of partner, we paired a port of Full with five large language model (LLM) configurations: GPT-5.4, GPT-5.4-mini with and without chain-of-thought (CoT), and Qwen3.6-A3B with and without CoT.

\paragraph{Corpus and agents.}
The games were generated in a separate LLM harness with its own engine and its own agent code. They did not use Eger et al.'s program, which produced HanabiData, or the simulator of the AI-AI corpus. The corpus comprises all 1{,}861 games in 39 conditions that were completed by 24 April 2026, with 9{,}542 LLM plays and 9{,}551 rule-agent plays. Each of 36 conditions has 50 games (seeds 0--49); three conditions whose runs stopped early have 45, 8 and 8 games (seeds 0--44, 0--7 and 0--7). The rule-based partners are ports, written for the harness, of seven production-rule agents (IGGI, Internal, Outer, Piers, Van den Bergh, Flawed, Simple) and of the two agents of Eger et al. (Intentional, Full). These ports are separate code from the agents of the AI-AI corpus. The ports of Intentional and Full were modeled on the corresponding \texttt{pyhanabi} classes (Appendix~\ref{app:agents_identity}), and their fidelity was not verified against a reference test. The port of Full is therefore not established to be equivalent to the HanabiData implementation, and the values of this appendix are not compared numerically with the HanabiData value of $+16.9$~pp. In the rest of this appendix, ``Full'' denotes the port.

\paragraph{Prompt, posterior and parse failures.}
The LLM was prompted with the rules, the game state and a per-card candidate analysis. The prompt also held a persistent convention pool seeded with ``assume your partner is a rational, cooperative agent.'' The literal failure probability $\hat{p}$ is computed exactly as in \S\ref{sec:posterior}; the per-card approximation moved no cell by more than 0.7~pp. The harness replaced LLM outputs that failed to parse with a fallback \emph{discard} (548 of 52{,}686 LLM actions, concentrated in the Qwen conditions without CoT). Because every fallback action was a discard, no play is affected.

\paragraph{Properties of the LLM harness.}
The harness differs from standard Hanabi in two rules, and some games were aborted (established from the harness code and the action logs). These games are therefore not complete standard games, and the completed-game filter of the main text does not apply to this corpus.
\begin{itemize}[leftmargin=*,nosep]
\item \emph{Round cap.} A cap of 30 rounds ended 1{,}590 of the 1{,}861 games (85.4\%) with cards left in the deck (1--13 cards, median 8). Eight games ended at round 30 with an empty deck, and 217 ended on lives lost. The plays analyzed are thus early- and mid-game plays. No game reached 25 points (maximum 19, mean 8.8), and scores from this corpus are not comparable with full-game scores.
\item \emph{Discards at eight tokens.} The engine accepts a discard when all eight information tokens are available. This occurred 1{,}131 times (742 by an LLM, 389 by a rule agent). Discards are not plays and enter no gap directly, but they change which states are reached.
\item \emph{Aborted games.} 46 games were aborted mid-game because a rule agent (Simple 26, Van den Bergh 14, Piers 6) repeatedly attempted a hint with no token available, which the engine rejected. The plays recorded before the abort are legal. These games are part of the pool of seven rule-agent partners below; none involved Full.
\end{itemize}

No hint touched zero cards (0 of 17{,}202 LLM hints and 0 of 31{,}286 rule-agent hints), there was no self-hint or logged hint at zero tokens, and no rule agent read its own cards or the deck. The LLM chose only actions from the engine's legal list. All 52{,}610 rule-agent actions, including the 6{,}347 by Full, were reproduced by re-running the agent code on a state rebuilt from public information only. The deal, the draws and the scores were reproduced from the seeds. The prompt contains no hidden card; this was established from the prompt-formatting code, not from the stored prompts. Across the 19{,}093 plays, no play with $\hat{p} = 0$ failed and no play with $\hat{p} = 1$ succeeded.

\begin{table}[!ht]
\centering
\caption{The five LLM\,$\times$\,Full cells (50 games each; seeds 0--49): counts by actor. ``At 8'' = discards made with eight tokens available, out of all discards. ``Cap'' = games ended by the 30-round cap (in parentheses: of which the deck was empty). ``Lives lost'' = games ended by the loss of all lives. Zero-card hints, self-hints, hints at zero tokens and aborted games: 0 in every row.}
\label{tab:llm_harness}
\footnotesize
\begin{tabular}{llrrrrrr}
\toprule
LLM & Actor & Actions & Plays & At 8 / discards & Fallbacks & Cap & Lives lost \\
\midrule
GPT-5.4 & LLM & 1{,}341 & 158 & 25 / 755 & 0 & 35 (1) & 15 \\
 & Full & 1{,}341 & 391 & 33 / 188 & 0 & & \\
GPT-5.4-mini & LLM & 1{,}363 & 190 & 51 / 907 & 0 & 37 (1) & 13 \\
 & Full & 1{,}354 & 273 & 43 / 125 & 0 & & \\
GPT-5.4-mini+CoT & LLM & 1{,}474 & 171 & 37 / 960 & 0 & 46 (0) & 4 \\
 & Full & 1{,}473 & 333 & 28 / 133 & 0 & & \\
Qwen3.6-A3B & LLM & 1{,}023 & 95 & 29 / 476 & 40 & 20 (2) & 30 \\
 & Full & 1{,}021 & 409 & 33 / 193 & 0 & & \\
Qwen3.6-A3B+CoT & LLM & 1{,}160 & 112 & 23 / 617 & 4 & 28 (0) & 22 \\
 & Full & 1{,}158 & 404 & 29 / 183 & 0 & & \\
\bottomrule
\end{tabular}
\end{table}

In the five LLM\,$\times$\,Full cells (250 games; Table~\ref{tab:llm_harness}), 162 games ended at the cap with cards left in the deck, 4 ended at round 30 with an empty deck and 84 ended on lives lost. Full made 166 of its 822 discards (20\%) at eight tokens, and the LLMs 165 of 3{,}715. No game was aborted.

\begin{figure}[H]
\centering
\includegraphics[width=\textwidth]{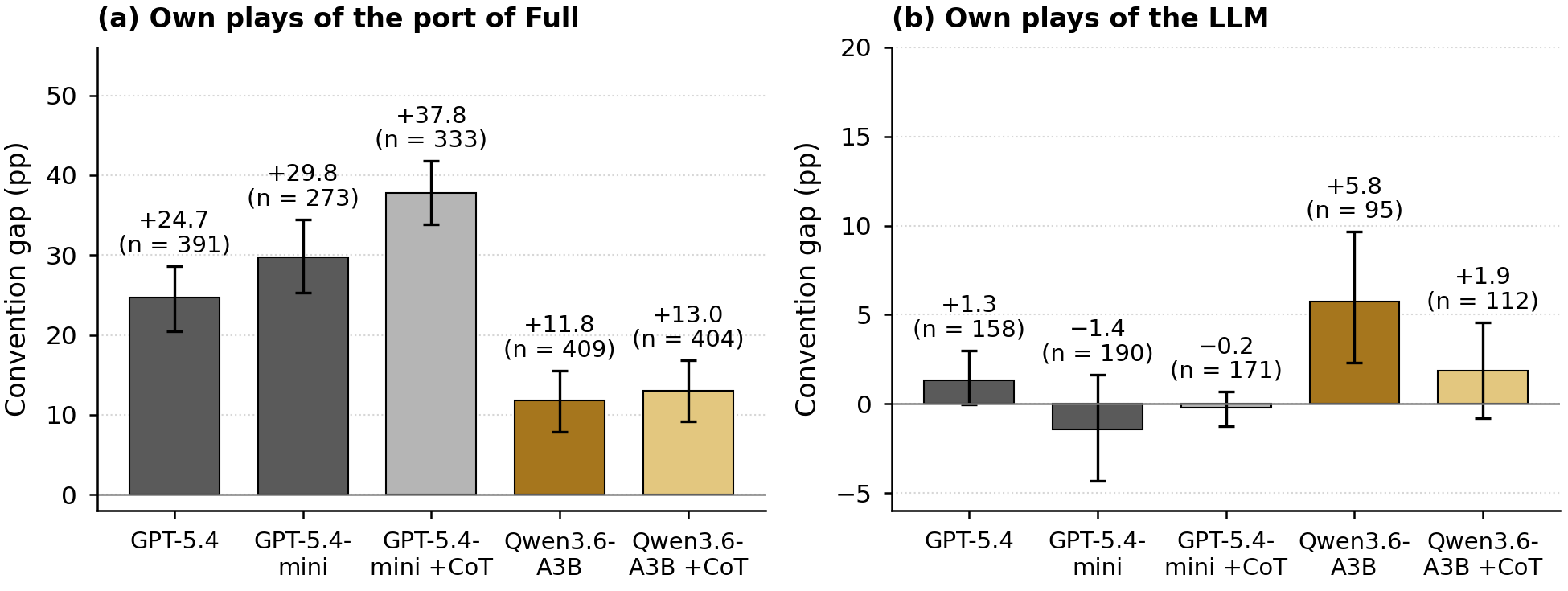}
\caption{Gap on each actor's own plays for the five pairings of an LLM with the port of Full (pooled over the plays of 50 capped games per pairing). \textbf{(a)} Plays of the port of Full ($n = 273$--$409$): all five values were at least $+11.8$~pp. \textbf{(b)} The LLM's plays ($n = 95$--$190$): only the Qwen3.6-A3B interval excluded zero. CoT variants in lighter colors. Error bars: 95\% play-level bootstrap intervals (2{,}000 resamples).}
\label{fig:llm_full_test}
\end{figure}

\paragraph{Estimation.}
The gap was computed separately on each actor's own plays (Figure~\ref{fig:llm_full_test}; Table~\ref{tab:llm_full_gap}). The intervals in the figure are play-level bootstrap intervals (2{,}000 resamples). Table~\ref{tab:llm_full_gap} adds game-level bootstrap intervals (10{,}000 draws, resampling the games with at least one play of the actor: 48--50 games for Full and 36--47 for the LLM). The cells whose play-level interval excludes zero are the same under game-level resampling. Neither interval accounts for a further dependence: the LLM carries its strategy and convention pool across the 50 games of a cell, so a cell is one adaptive run and not 50 independent games. All intervals are conditional on that run.

\begin{table}[!ht]
\centering
\caption{Gap on each actor's own plays in the LLM\,$\times$\,Full cells: plays, mean $\hat{p}$, failure rate, gap (pp), and 95\% bootstrap intervals resampling plays or games. The game-level lower bound printed as $-$0.0 lies between $-$0.03 and 0 across bootstrap seeds, so that interval includes zero.}
\label{tab:llm_full_gap}
\footnotesize
\begin{tabular}{llrrrrll}
\toprule
LLM & Actor & Plays & Mean $\hat{p}$ (\%) & Failure (\%) & Gap & CI, plays & CI, games \\
\midrule
GPT-5.4 & Full & 391 & 43.33 & 18.67 & +24.7 & [+20.5, +28.7] & [+21.1, +28.3] \\
GPT-5.4-mini & Full & 273 & 46.62 & 16.85 & +29.8 & [+25.2, +34.6] & [+24.2, +35.1] \\
GPT-5.4-mini+CoT & Full & 333 & 48.31 & 10.51 & +37.8 & [+33.7, +41.6] & [+34.1, +41.3] \\
Qwen3.6-A3B & Full & 409 & 42.17 & 30.32 & +11.8 & [+7.9, +15.7] & [+8.3, +15.3] \\
Qwen3.6-A3B+CoT & Full & 404 & 39.77 & 26.73 & +13.0 & [+9.3, +16.7] & [+9.3, +16.4] \\
\midrule
GPT-5.4 & LLM & 158 & 1.95 & 0.63 & +1.3 & [$-$0.02, +3.0] & [$-$0.0, +2.9] \\
GPT-5.4-mini & LLM & 190 & 21.19 & 22.63 & $-$1.4 & [$-$4.5, +1.7] & [$-$4.1, +1.1] \\
GPT-5.4-mini+CoT & LLM & 171 & 1.52 & 1.75 & $-$0.2 & [$-$1.2, +0.8] & [$-$1.2, +0.7] \\
Qwen3.6-A3B & LLM & 95 & 9.96 & 4.21 & +5.8 & [+2.2, +9.6] & [+2.3, +10.0] \\
Qwen3.6-A3B+CoT & LLM & 112 & 5.43 & 3.57 & +1.9 & [$-$0.8, +4.6] & [$-$0.2, +4.2] \\
\bottomrule
\end{tabular}
\end{table}

\paragraph{Plays of the port of Full (panel a).}
Full's own-play gap was positive in every LLM pairing. It ranged from $+11.8$~pp (Qwen3.6-A3B) to $+37.8$~pp (GPT-5.4-mini+CoT), and the smallest lower interval bound was $+7.9$~pp. In this harness, Full's plays thus failed less often than the literal reference predicts when its partner was one of these LLMs. The gap is over all of Full's plays; no hint-count stratum was computed for this corpus. We take conditions (i) and (iii) to hold for this harness (seeded pseudo-random shuffles, which we treat as uniform; Full's actions reproduced from public information only). Under Corollary~\ref{cor:nonzero}, the positive estimated gap is then evidence that the LLMs' choices depended on Full's hidden cards beyond their literal content; one possibility is that they hinted playable cards. This is consistent with, and does not establish, that Full's hint-reading rule is what turns this dependence into successful plays.

\paragraph{The LLMs' plays (panel b).}
Against Full, the LLM's own-play gap was close to zero in four of five conditions (Table~\ref{tab:llm_full_gap}). The only cell whose interval excluded zero was Qwen3.6-A3B ($+5.8$~pp $[+2.2, +9.6]$, $n = 95$). With CoT enabled the value was $+1.9$~pp $[-0.8, +4.6]$. Pooled across the seven production-rule partners, which panel~(b) does not show, the LLM-side gap was $-$0.05~pp for GPT-5.4 ($n = 1{,}953$) and $-$0.89~pp for GPT-5.4-mini ($n = 1{,}987$). It was $-$0.13~pp for GPT-5.4-mini+CoT ($n = 2{,}131$) and $-$1.11~pp for Qwen3.6-A3B ($n = 1{,}945$). These pools include the plays of the 46 aborted games. Qwen3.6-A3B+CoT was run with one such partner only (IGGI; 8 games, $n = 52$, $+0.45$~pp).

\paragraph{Pinned zeros and uncertain plays.}
Most of these values near zero carry no evidence about what Full's hints conveyed beyond their literal content, because the LLMs rarely played uncertain cards, that is, cards with $\hat{p} > 0$ (Table~\ref{tab:llm_pinned}). GPT-5.4 and GPT-5.4-mini+CoT played cards certain to be playable under the literal reference ($\hat{p} = 0$) on 96--97\% of their plays. Then $\hat{p} = y = 0$ on nearly every play, and the gap is zero whatever the partner's hints depend on (pinned, \S\ref{sec:method_convgap}). Only GPT-5.4-mini without CoT played many uncertain cards ($n = 56$). These failed at about the literal rate (76.8\% observed, 95\% Wilson interval [64, 86]\%, against 71.9\% predicted). This is the only condition in which the near-zero gap is informative: on its uncertain plays the observed rates were compatible with the literal reference at this sample size, which is not evidence of equivalence. The positive Qwen3.6-A3B cell rests on 17 uncertain plays and is one of five comparisons. For comparison, the share of Full's plays at $\hat{p} = 0$ in the same cells was 19--31\%.

\paragraph{Scope.}
The two panels differ. The gap of the port of Full was positive with every LLM partner. The LLMs' gaps were mostly pinned at zero because the LLMs played almost only cards certain to be playable under the literal reference. These statements concern five LLM configurations, one prompt design, one adaptive 50-game run per cell, capped games and an unverified port of Full. Full's own hints shape the states in which it later plays, so the own-play gap is a property of the pair, not of Full alone. Whether the result holds in full-length games under standard end-of-game and discard rules, or with a verified implementation of Full, was not tested. The results do not establish how these LLMs, or LLMs in general, interpret hints. Per-condition tables, game counts, fallback actions and the harness checks are in the public repository.

\begin{table}[!ht]
\centering
\caption{The LLM's own plays against Full: share of plays at $\hat{p} = 0$, and the number, mean $\hat{p}$ and failure rate of the remaining plays.}
\label{tab:llm_pinned}
\footnotesize
\begin{tabular}{lrrrrrr}
\toprule
Condition & Plays & $\hat{p} = 0$ (\%) & Plays with $\hat{p} > 0$ & Mean $\hat{p}$ (\%) & Failure (\%) & Gap (pp) \\
\midrule
GPT-5.4 & 158 & 96.2 & 6 & 51.3 & 16.7 & +1.3 \\
GPT-5.4-mini+CoT & 171 & 97.1 & 5 & 51.9 & 60.0 & $-$0.2 \\
GPT-5.4-mini & 190 & 70.5 & 56 & 71.9 & 76.8 & $-$1.4 \\
Qwen3.6-A3B+CoT & 112 & 85.7 & 16 & 38.0 & 25.0 & +1.9 \\
Qwen3.6-A3B & 95 & 82.1 & 17 & 55.7 & 23.5 & +5.8 \\
\bottomrule
\end{tabular}

\smallskip
{\footnotesize Mean $\hat{p}$ and failure rate are over the plays with $\hat{p} > 0$; the gap is over all plays.}
\end{table}

\section{Further Discussion}
\label{app:further}

These notes place the results in context and add no claim.

\subsection{The Literal Failure Probability as a Reference Model}
\label{app:rational}

The literal failure probability $\hat{p}$ of \S\ref{sec:posterior} is the failure probability that an observer would assign if hints carried only their literal content. It enumerates the card identities consistent with the hints and the visible cards, weights them by the remaining copy counts, and returns an exact finite sum. It plays the role of a fixed reference, in the way the literal listener $L_0$ serves as the reference in Rational Speech Acts models. It is not a model of what a player knows, and it is not an optimal decision rule: it uses no information from the timing of hints, from the partner's policy or from the choice of which card to play.

\S\ref{sec:discussion} discusses what a positive or negative convention gap does and does not show.

\subsection{Relation to Subjective Cooperativity Ratings}
\label{app:convergence}

\citet{attig2024perceived} had participants play Hanabi with a rule-based agent (Piers) and a reinforcement-learning agent ($N = 8$). The reinforcement-learning agent was rated lower on Perceived Cooperativity (2.43 vs.\ 3.58) and on Playing Style Similarity (1.81 vs.\ 3.13). The lower-rated agent had been designed for zero-shot coordination. Both studies found differences between agents: theirs in questionnaire ratings, ours in the convention gap. The two studies measure different things. Theirs measured perceived cooperation by questionnaire; the convention gap is a calibration diagnostic relative to a literal reference and does not measure cooperation quality or subjective cooperativity. We did not collect ratings, so the relation between the gap and perceived cooperativity is untested.
\section{Statement on the Use of AI-Assisted Tools}
\label{app:ai_use}

The authors used AI assistants (Claude Code with Claude Fable 5.1 and Claude Opus 5.5; Anthropic, 2026) for coding, data analysis, and drafting and editing text. The assistants also contributed to the methodology: they proposed several of the robustness analyses and controls reported here, among them the exact null control and its prespecified replication (Appendix~\ref{app:nullcontrol}) and the split of the gap into plays with zero and positive literal failure probability, and they drafted the formal statement and proof of Proposition~\ref{prop:zero} from the authors' argument. These contributions arose in extended multi-turn sessions. The transcripts of the Claude Code sessions are retained. Earlier sessions used Claude models in Anthropic's Cowork application; their transcripts and model versions were not retained. The authors checked the proofs, the code and every reported number, and are responsible for all content.

\end{document}